%% file: TwoHands.tex
\documentclass[acmtog]{acmart}

\usepackage{booktabs} 

\setcitestyle{square}

\usepackage[ruled]{algorithm2e} 

\SetAlFnt{\small}
\SetAlCapFnt{\small}
\SetAlCapNameFnt{\small}
\SetAlCapHSkip{0pt}

\input{macros.tex}

\usepackage[prologue]{xcolor}
\IncMargin{-\parindent}
\usepackage{rotating}
\usepackage{subcaption}
\usepackage{forloop}

\begin{document}
\title{Real-time Pose and Shape Reconstruction of Two Interacting Hands With a Single Depth Camera}

\author{Franziska Mueller}
\affiliation{%
  \institution{MPI Informatics, Saarland Informatics Campus}}
\email{frmueller@mpi-inf.mpg.de}
\author{Micah Davis}
\affiliation{%
  \institution{Universidad Rey Juan Carlos}}
\email{davis.micah@urjc.es}
\author{Florian Bernard}
\email{fbernard@mpi-inf.mpg.de}
\author{Oleksandr Sotnychenko}
\affiliation{%
  \institution{MPI Informatics, Saarland Informatics Campus}}
\email{osotnych@mpi-inf.mpg.de}
\author{Mickeal Verschoor}
\email{mickeal.verschoor@urjc.es}
\author{Miguel A. Otaduy}
\email{miguel.otaduy@urjc.es}
\author{Dan Casas}
\affiliation{%
  \institution{Universidad Rey Juan Carlos}}
\email{dan.casas@urjc.es}
\author{Christian Theobalt}
\affiliation{%
  \institution{MPI Informatics, Saarland Informatics Campus}}
\email{theobalt@mpi-inf.mpg.de}

\renewcommand{\shortauthors}{Mueller, Davis, Bernard, Sotnychenko, Verschoor, Otaduy, Casas, and Theobalt}

\begin{teaserfigure}
  \frame{\includegraphics[width=\linewidth]{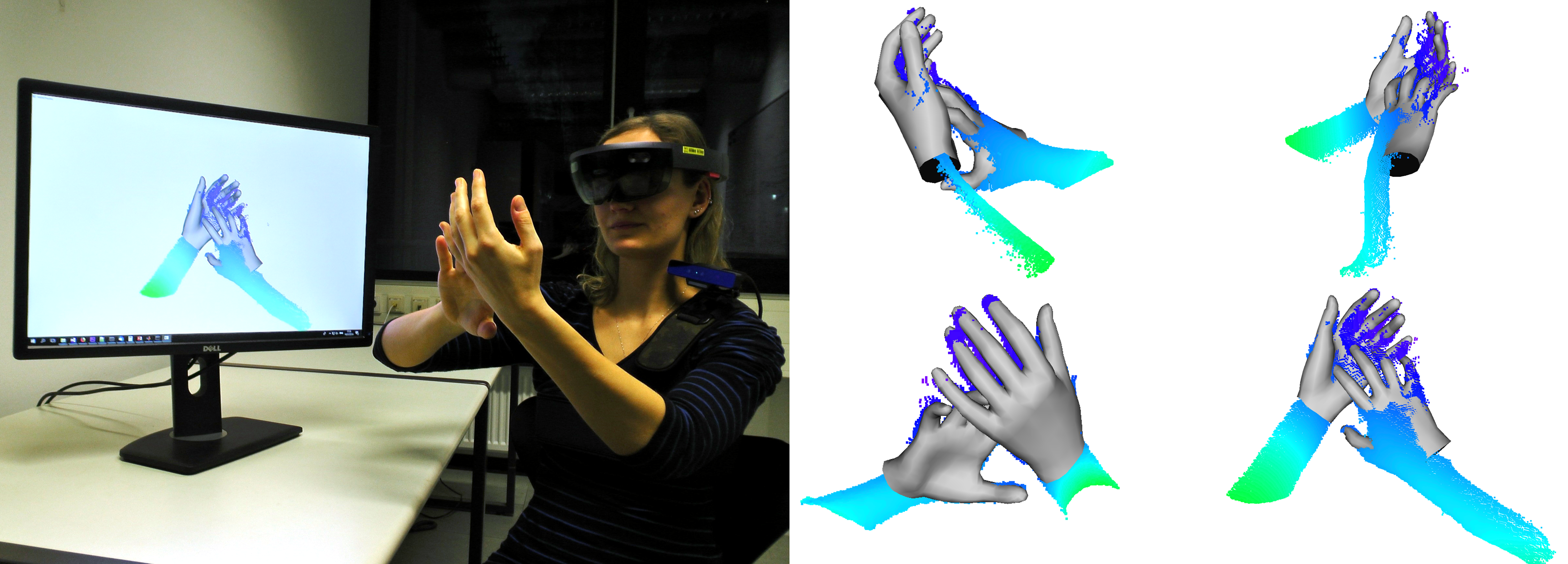}}
  \caption{We present a method that estimates the pose and shape of two interacting hands in real time from a single depth camera. On the left we show an AR setup with a shoulder-mounted depth camera. On the right we show the depth data and the estimated 3D hand pose and shape from four different views.}
  \label{fig:teaser}
\end{teaserfigure}

\begin{abstract}
We present a novel method for real-time pose and shape reconstruction of two strongly interacting hands. Our approach is the first two-hand tracking solution that combines an extensive list of favorable properties, namely it is marker-less, uses a single consumer-level depth camera, runs in real time, handles inter- and intra-hand collisions, and automatically adjusts to the user's hand shape. In order to achieve this, we embed a recent parametric hand pose and shape model and a dense correspondence predictor based on a deep neural network into a suitable energy minimization framework. For training the correspondence prediction network, we synthesize a two-hand dataset based on physical simulations that includes both hand pose and shape annotations while at the same time avoiding inter-hand penetrations. To achieve real-time rates, we phrase the model fitting in terms of a nonlinear least-squares problem so that the energy can be optimized based on a highly efficient GPU-based Gauss-Newton optimizer. We show state-of-the-art results in scenes that exceed the complexity level demonstrated by previous work, including tight two-hand grasps, significant inter-hand occlusions, and gesture interaction.\footnote{project website: https://handtracker.mpi-inf.mpg.de/projects/TwoHands/}
\end{abstract}

%
%


%
%


\setcopyright{acmlicensed}
\acmJournal{TOG}
\acmYear{2019}\acmVolume{38}\acmNumber{4}\acmArticle{49}\acmMonth{7} \acmDOI{10.1145/3306346.3322958}

\maketitle

\input{sections/intro.tex}
\input{sections/relwork.tex}
\input{sections/method.tex}
\input{sections/evaluation.tex}
\input{sections/discussion.tex}
\input{sections/conclusion.tex}

\begin{acks}
The authors would like to thank all participants of the live recordings.
The work was supported by the ERC Consolidator Grants \emph{4DRepLy} (770784) and \emph{TouchDesign} (772738).
Dan Casas was supported by a Marie Curie Individual Fellowship (707326).
\end{acks}

\bibliographystyle{ACM-Reference-Format}
\bibliography{TwoHands}

\input{sections/appendix.tex}
\end{document}

%% file: macros.tex
\DeclareMathOperator*{\argmin}{argmin}

\usepackage{pifont}

\newcommand{\cmark}{\ding{51}}%
\newcommand{\xmark}{\ding{55}}%
\usepackage{adjustbox}
\usepackage{array}
\newcolumntype{R}[2]{%
    >{\adjustbox{angle=#1,lap=\width-(#2)}\bgroup}%
    l%
    <{\egroup}%
}
\newcommand*\rot{\multicolumn{1}{R{45}{1em}}}

\newcommand{\lleft}{\text{left}}
\newcommand{\rright}{\text{right}}
\newcommand{\total}{\text{total}}
\newcommand{\data}{\text{data}}
\newcommand{\reg}{\text{reg}}
\newcommand{\point}{\text{point}}
\newcommand{\plane}{\text{plane}}
\newcommand{\shape}{\text{shape}}
\newcommand{\pose}{\text{pose}}
\newcommand{\temp}{\text{temp}}
\newcommand{\coll}{\text{coll}}

\newcommand{\filluptopage}[1]{%
  \clearpage
  \loop\ifnum\value{page}<#1\relax
    \null\clearpage
  \repeat
  \loop\ifnum\value{page}=#1\relax
    \null\clearpage
  \repeat
}

\newcommand{\cref}[1]{Section \ref{#1}}

\makeatletter
\@tfor\next:=abcdefghijklmnopqrstuvwxyz\do{%
  \def\command@factory#1{%
    \expandafter\def\csname vec#1\endcsname{\mathbf{#1}}
  }
 \expandafter\command@factory\next
}

\@tfor\next:=ABCDEFGHIJKLMNOPQRSTUVWXYZ\do{%
  \def\command@factory#1{%
    \expandafter\def\csname mat#1\endcsname{\mathbf{#1}}
  }
 \expandafter\command@factory\next
}

\@tfor\next:=ABCDEFGHIJKLMNOPQRSTUVWXYZ\do{%
  \def\command@factory#1{%
    \expandafter\def\csname set#1\endcsname{\mathcal{#1}}
  }
 \expandafter\command@factory\next
}

\def\greekmatrices#1{%
 \@for\next:=#1\do{%
    \def\X##1;{%
     \expandafter\def\csname mat##1\endcsname{\boldsymbol{\csname##1\endcsname}}
     }
   \expandafter\X\next;
  }
}
\def\greekvectors#1{%
 \@for\next:=#1\do{%
    \def\X##1;{%
     \expandafter\def\csname vec##1\endcsname{\boldsymbol{\csname##1\endcsname}}
     }
   \expandafter\X\next;
  }
}
\greekvectors{alpha,beta,iota,gamma,lambda,nu,eta,Gamma,varsigma,Phi,Theta,delta,mu}
\greekmatrices{alpha,beta,iota,gamma,lambda,nu,eta,Gamma,varsigma,Phi,Theta,mu}

\makeatother

\newcommand{\R}{\mathbb{R}}

%% file: sections/intro.tex
\section{Introduction}
The marker-less estimation of hand poses is a challenging problem that has received a lot of attention in the vision and graphics communities. The relevance of the problem is owed to the fact that hand pose recognition plays an important role in many application areas such as human-computer interaction \cite{kim2012digits}, augmented and virtual reality (AR/VR) \cite{holl2018efficient}, sign language recognition \cite{koller2016deep}, as well as body language recognition relevant for psychology.
Depending on the particular application, additional requirements are frequently imposed on the method, such as performing hand tracking in real time, or dynamically adapting the tracking to person-specific hand shapes for increased accuracy. 
Ideally, reconstruction should be possible with a simple hardware setup and therefore methods with a single color or depth camera are widely researched.
Existing marker-less methods for hand pose estimation typically rely on either RGB \cite{Zimmermann:2017um,mueller_cvpr2018,cai2018weakly}, depth images \cite{sridhar_cvpr2015,Taylor_siggraphasia2017,Supan2018,Yuan_2018_CVPR}, or  a combination of both \cite{rogez_eccv2014workshop,oikonomidis2011efficient}. 
The major part of existing methods considers the problem of processing a single hand only \cite{oberweger_iccv2015,Ye_2018_ECCV,qian_cvpr2014}. 
Some of them are even able to handle object interactions~\cite{tzionas_ijcv2016,sridhar_eccv2016,mueller_iccv2017}, which is especially challenging due to potential occlusions.

As humans naturally use both their hands during daily routine tasks, many applications require to track both hands simultaneously (see Fig.~\ref{fig:teaser}), rather than tracking a single hand in isolation. 
While there are a few existing works that consider the problem of tracking two hands at the same time, they are limited in at least one of the following points: 
(i) they only work for rather simple interaction scenarios (e.g.\:no tight two-hand grasps, significant inter-hand occlusions, or gesture interaction),
(ii) they are computationally expensive and not real-time capable, 
(iii) they do not handle collisions between the hands,
(iv) they use a person-specific hand model that does not automatically adapt to unseen hand shapes, or
(v) they heavily rely on custom-built dedicated hardware. 
In contrast to existing methods, our approach can handle two hands in interaction while not having any of the limitations (i)-(v), see \autoref{table:PropertiesSOTA}.

We present for the first time a marker-less method that can track two hands with complex interactions in real time with a single depth camera, while at the same time being able to estimate the person's hand shape. From a technical point of view, this is achieved thanks to a novel learned dense surface correspondence predictor that is combined with a recent parametric hand model~\cite{Romero_siggraphasia2017}.
These two components are combined in an energy minimization framework to find the pose and shape parameters of both hands in a given depth image. 
Inspired by the recent success of deep learning approaches, especially for image-based prediction tasks \cite{badrinarayanan2015segnet,zhang2017beyond, Guler_2017_CVPR, Guler_2018_CVPR}, we employ a correspondence regressor based on deep neural networks. Compared to ICP-like local optimization approaches, using such a global correspondence predictor is advantageous, as it is less prone to the failures caused by wrong initialization and can easily recover even from severe tracking errors (see our supplementary video).
Since it is not feasible to obtain reliable dense correspondence annotations in real data, we create a synthetic dataset of interacting hands to train the correspondence regressor. 
Here, it is crucial to obtain \emph{natural}  interactions between the hands, which implies that simply rendering a model of the left and of the right hand (in different poses) into the same view is not sufficient. 
Instead, we make use of an extension of the motion capture-driven physical simulation \cite{verschoor2018soft} that leads to faithful, collision-free, and physically plausible simulated hand-hand interactions.

The main contributions of our approach are summarized as follows:
\begin{itemize}
    \item For the first time we present a method that can track two interacting hands in real time with a single depth camera, while at the same time being able to estimate the hand shape and taking collisions into account.
    \item Moreover, our approach is the first one that leverages physical simulations for creating a two-hand tracking dataset that includes both pose and dense shape annotations while at the same time avoiding inter-hand penetrations. 
    \item Contrary to existing methods, our approach is more robust and reliable in involved hand-hand interaction settings.
\end{itemize}

\begin{table}[]
\caption{
Our method is the first to combine several desirable properties.
}
\setlength{\tabcolsep}{9pt}
\begin{tabular}{lccccc}
& \rot{[Oikon. 2012]} & \rot{[Tzionas 2016]} & \rot{[Tkach 2017]} & \rot{[Taylor 2017]} & \rot{Ours} \\
\midrule
Interacting Hands & \cmark & \cmark & \xmark & \cmark & \cmark \\
Shape Estimation & \xmark & \xmark & \cmark & \xmark & \cmark \\ 
Real Time & \xmark & \xmark & \cmark & \cmark & \cmark \\ 
Commodity Sensor & \cmark & \cmark & \cmark & \xmark  & \cmark \\ 
Collision Avoidance & \cmark & \cmark & \cmark & \xmark & \cmark \\
\bottomrule
\end{tabular}
\setlength{\tabcolsep}{6pt}
\label{table:PropertiesSOTA}
\end{table}

%% file: sections/relwork.tex
\begin{figure*}[h!t!]
  \includegraphics[width=\linewidth,page=1,trim={0 7.5cm 0 8.3cm},clip]{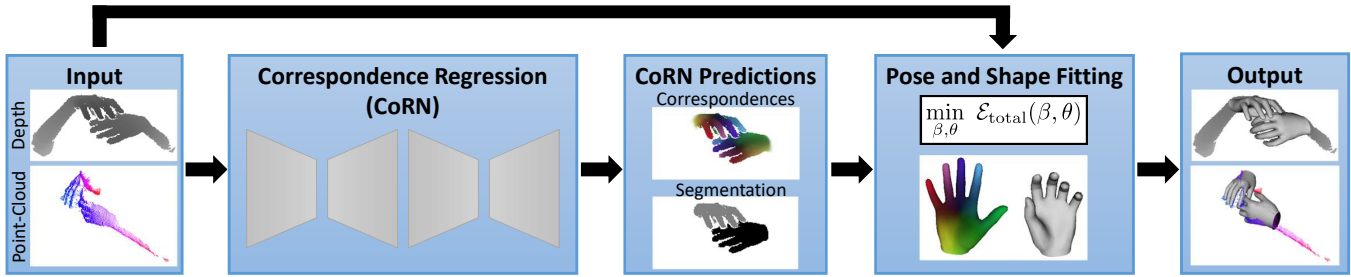}
  \caption
  {Overview of our two-hand pose and shape estimation pipeline.
  Given only a depth image as input, our dense correspondence regression network (CoRN) computes a left/right segmentation and a vertex-to-pixel map. To obtain the hand shape estimation and pose tracking we use this data in an energy minimization framework, where a parametric hand pose and shape model is fit so that it best explains the input data.}
  \label{fig:pipeline}
\end{figure*}

\section{Related Work}

Hand pose estimation is an actively researched topic that has a wide range of applications, for example in human--computer interaction or activity recognition.
While many methods reconstruct the motion of a single hand in isolation, comparably few existing approaches can  work with multiple interacting hands, or estimate the hand shape.
In the following, we discuss related works that tackle one of these problems.

\paragraph{Capturing a Single Hand:}
Although multi-camera setups \cite{sridhar_iccv2013,oikonomidis2011full} are advantageous in terms of tracking quality, e.g.\:less ambiguity under occlusions, they are infeasible for many applications due to their inflexibility and cumbersome setup.
Hence, the majority of recent hand pose estimation methods either uses a single RGB or depth camera.
These methods can generally be split into three categories: generative, discriminative, and hybrid algorithms.
Generative methods fit a model to image evidence by optimizing an objective function \cite{melax2013dynamics, tagliasacchi_sgp2015,tkach2016sphere}.
While they have the advantage that they generalize well to unseen poses, a downside is that generally they are sensitive to the initialization and thus may not recover from tracking failures.
At the other end of the spectrum are discriminative methods that use machine learning techniques to estimate the hand pose with (usually) a single prediction~\cite{wan2016hand, tang2014latent, rogez_eccv2014workshop,choi2015collaborative}.
Despite being dependent on their training corpus, they do not require initialization at test time and hence recover quickly from failures.
The recent success of deep neural networks has led to many works that regress hand pose from depth images \cite{tompson_tog2014,oberweger_iccv2015,wan2017crossing,Baek_2018_CVPR,Ge_2018_CVPR} or even from more underconstrained monocular RGB input \cite{simon2017hand,Zimmermann:2017um,Spurr_2018_CVPR,cai2018weakly,mueller_cvpr2018}.
Hybrid methods \cite{sridhar_cvpr2015,tang_iccv2015} combine generative and discriminative techniques, for example to get robust initialization for model fitting.
A more detailed overview of depth-based approaches for single-hand pose estimation is provided by Yuan et~al.~\shortcite{Yuan_2018_CVPR} and Supan\v{c}i\v{c} et~al.~\shortcite{Supan2018}.

\paragraph{Hand Shape Models:} 
There exist various hand models, i.e.\:models of hand geometry and pose, that are used for pose estimation by generative and hybrid methods, ranging from a set of geometric primitives \cite{oikonomidis2011efficient,qian_cvpr2014} to surface models like meshes \cite{sharp2015accurate}.
Such models are usually personalized to individual users and are obtained manually, e.g.\:by laser scans or simple scaling of a base model.
Only few methods estimate a detailed hand shape from depth images automatically.
Khamis et~al.~\shortcite{khamis2015learning} build a shape model of a hand mesh from sets of depth images acquired from different users.
A method for efficiently fitting this model to a sequence of a new actor was subsequently presented by Tan et~al.~\shortcite{tan_cvpr2016}.
Tkach et~al.~\shortcite{Tkach_siggraphasia2017} jointly optimize pose and shape of a sphere mesh online, and accumulate shape information over time to minimize uncertainty.
In contrast, Remelli et~al.~\shortcite{Remelli_2017_ICCV} fit a sphere mesh directly to the whole image set by multi-stage calibration with local anisotropic scalings. 
Recently, Romero~et~al.~\shortcite{Romero_siggraphasia2017} proposed a low-dimensional parametric model for hand pose and shape which was obtained from 1000 high-resolution 3D scans of 31 subjects in a wide variety of hand poses.

\paragraph{Capturing Two Hands:}
Reconstructing two hands jointly introduces profound new challenges, such as 
the inherent segmentation problem and more severe occlusions.
Some methods try to overcome these challenges by using marker gloves \cite{han2018online} or multi-view setups \cite{ballan_eccv2012}.
Other approaches tackle the problem from a single depth camera to achieve more flexibility and practical usability. 
An analysis-by-synthesis approach is employed by Oikonomidis et~al.~\shortcite{oikonomidis2012tracking} who minimize the discrepancy of a rendered depth image and the input using particle swarm optimization. 
Kyriazis and Argyros~\shortcite{kyriazis_cvpr2014} apply an ensemble of independent trackers, where the per-object trackers broadcast their state  to resolve collisions.
Tzionas et~al.~\shortcite{tzionas_ijcv2016} use discriminatively detected salient points and a collision term based on distance fields
to obtain an intersection-free model fit.
Nevertheless, the aforementioned single-camera methods do not achieve real-time rates, and operate at 0.2 to 4 frames per second. There exist some methods that track two hands in real time, albeit without being able to deal with close hand-hand interactions.
Taylor et~al.~\shortcite{taylor_siggraph2016} jointly optimize pose and correspondences of a subdivision surface model but the method fails when the hands come close together, making it unusable for capturing any hand-hand interaction. 
Taylor et~al.~\shortcite{Taylor_siggraphasia2017} employ machine learning techniques for hand segmentation and palm orientation initialization, and subsequently fit an articulated distance function.
They use a custom-built high frame-rate depth camera to minimize the motion between frames, thus being able to fit the model with very few optimizer steps.
However, they do not resolve collisions and they do not estimate hand shape, so that they require a given model for every user.
While they show some examples of hand-hand interactions, they do not show very close and elaborate interactions, e.g.\:with tight grasps.

In contrast to previous two-hand tracking solutions, our approach (i) runs in real time with a commodity camera, (ii) is marker-less, (iii) uses a single (depth) camera only,  (iv) handles hand collisions, and (v) automatically adjusts to the user's hand shape.

%% file: sections/method.tex
\section{Overview}
In Fig.~\ref{fig:pipeline} we provide an overview of the pipeline for performing real-time hand pose and shape reconstruction of two interacting hands from a single depth sensor. 
First, we train a neural network that regresses dense  correspondences between the hand model and a depth image that depicts two (possibly interacting) hands. 
In order to disambiguate between pixels that belong to the left hand, and pixels that belong to the right hand, our dense correspondence map also encodes the segmentation of the left and right hand. 
For obtaining realistic training data of hand interactions, we make use of motion capture-driven physical simulation to generate (synthetic) depth images along with ground-truth correspondence maps. This data is additionally augmented with real depth data that is used for training the segmentation channel of the correspondence map.
The so-obtained correspondence maps are then used to initialize an energy minimization framework, where we fit a  parametric hand model to the given depth data.
During fitting we make use of statistical pose and shape regularizers to avoid implausible configurations, a temporal smoothness regularizer, as well as a collision regularizer in order to avoid interpenetration between both hands and within each hand.

In order to achieve real-time performance, we phrase the energy minimization step in terms of a nonlinear least-squares formulation, and make use of a highly efficient ad-hoc data-parallel GPU implementation based on a Gauss-Newton optimizer. 

In the remainder of this section we describe the hand model that we use for the tracking (Sec.~\ref{sec:hand-model}).
Subsequently, we provide a detailed explanation of the dense correspondence regression including the data generation (Sec.~\ref{sec:dense-correspondeces}),  followed by a description of the pose and shape estimation (Sec.~\ref{sec:energy}).

\subsection{Hand Model} \label{sec:hand-model}
\begin{figure}
  \includegraphics[width=\linewidth,page=3,trim={0.5cm 3cm 3cm 3cm},clip]{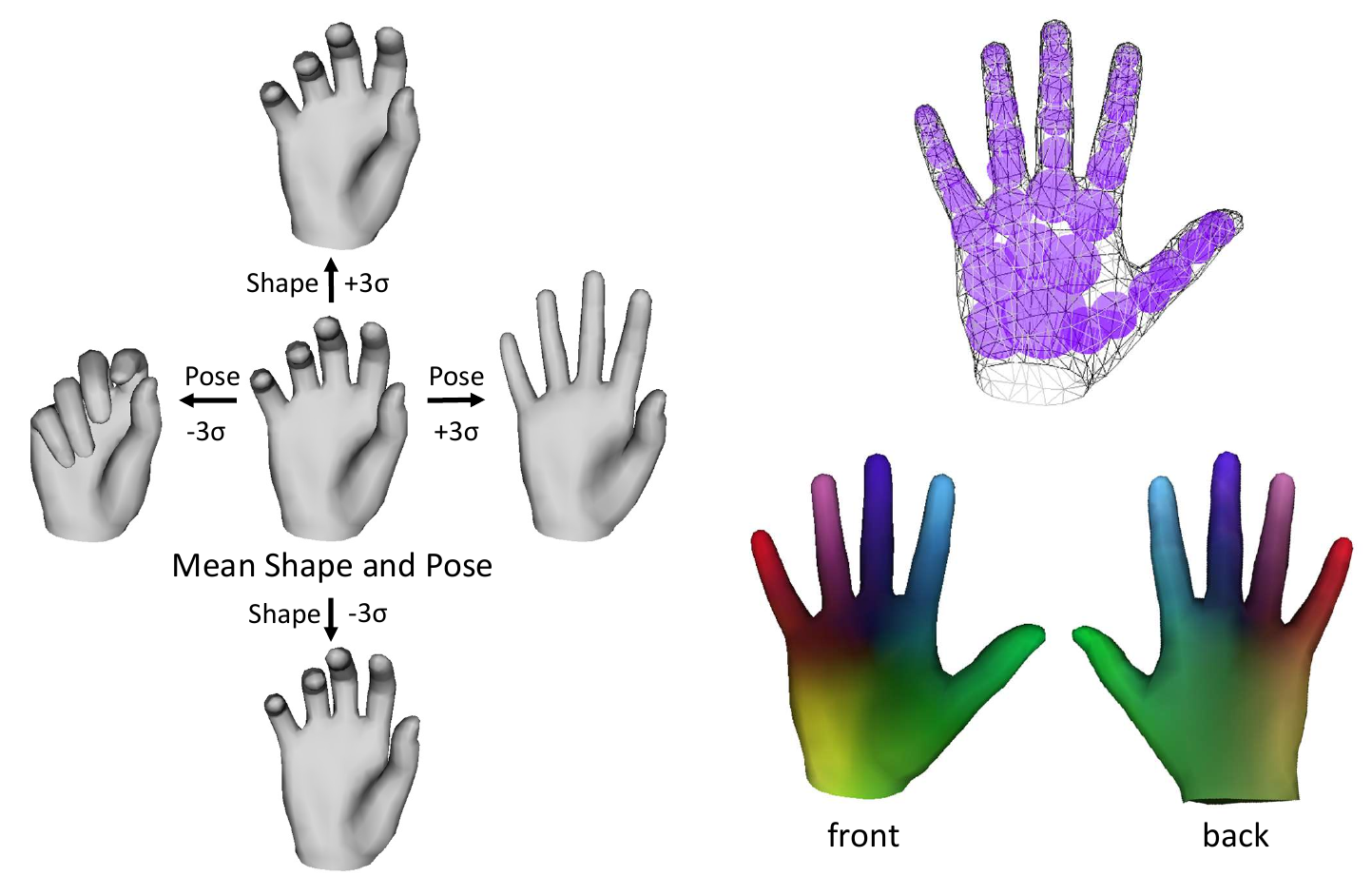}
  \caption
  {Illustration of MANO hand model (left) that is augmented with our collision proxies (top right), as well as the correspondence color-encoding (bottom right). Notice that front and back color assignments differ in saturation, especially in the palm area.}
  \label{fig:model}
\end{figure}

As 3D hand representation, we employ the recently introduced MANO model \cite{Romero_siggraphasia2017}, which is a low-dimensional parametric hand surface model that captures hand shape variation as well as hand pose variation, see Fig.~\ref{fig:model} (left). 
It was built from about 1000 3D hand scans of 31 persons in wide range of different hand poses. The hand surface is represented by a 3D mesh with vertices $\setV$, where $N_V := \vert \setV \vert = 778$.
The MANO model defines a function $\vecv: \R^{N_S} \times \R^{N_P} \rightarrow \R^{3N_V}$, that computes the 3D positions of all of the mesh's  $N_V$ vertices, given a shape parameter vector $\beta \in \R^{N_S}$ and pose parameter vector $\theta \in \R^{N_P}$, with $N_S = 10$ and $N_P = 51$. 
We use the notation  $\vecv_i(\beta,\theta) \in \R^3$ to denote the 3D position of the $i$-th vertex. Parameters $\beta$ and $\theta$ are coefficients of a low-dimensional pose and shape subspace that was obtained by PCA on the training data. 
As such, the MANO model naturally allows for a statistical regularization by simply imposing that the parameters are close to zero, which corresponds to a Tikhonov regularizer.

Note that since we are tracking two hands that can move independently, we use independent hand models of the left and right hand, which are denoted by $\setV_{\lleft}$ and $\setV_{\rright}$ with vertices
$\vecv_{\lleft}(\beta_\lleft,\theta_\lleft)$ and $\vecv_{\rright}(\beta_\rright,\theta_\rright)$, respectively. 
For notational convenience, we stack the parameters of the left and right hand so that we have $\beta = (\beta_\lleft, \beta_\rright)$ and $\theta = (\theta_\lleft, \theta_\rright)$, and we use $\setV$ with $N_{v} := \vert \setV \vert = 2{\cdot}778$ to denote the combined vertices of the left and the right hand.

To resolve interpenetrations at high computational efficiency, we add collision proxies to the hand model.
We follow the approach of Sridhar~et~al.~\shortcite{sridhar_cvpr2015}, who approximate the volumetric extent of the hand with a set of spheres that are modeled with 3D Gaussians.
Using this formulation, interpenetrations can then be avoided by penalizing the overlap of the Gaussians during pose optimization.
Note that the overlap between Gaussians is differentiable and can be computed in closed form---in contrast to na{\"i}ve binary collision checking.
We combine the Gaussians with the existing MANO model by rigging their positions to the hand joints and coupling their standard deviations to pairs of manually selected vertices. 
By doing this, we ensure that the position and size of the Gaussians vary in accordance with the pose and shape parameters $\beta$ and $\theta$.
For each hand we add $35$ 3D Gaussians, which leads to a total number of $N_C = 70$ for the combined two-hands model.
A visualization of the isosurface at 1 standard deviation of the Gaussians is shown in Fig.~\ref{fig:model} (top right).
Next, we describe our correspondence regressor that is eventually coupled with the two-hands model in order to perform pose and shape reconstruction.

\section{Dense Correspondence Regression} \label{sec:dense-correspondeces}
Let $\setI$ be the input depth image of pixel-dimension $h$ by $w$ defined over the image domain $\Omega$. 
Our aim is to learn a vertex-to-pixel correspondence map $c: {\setV} \rightarrow \bar{\Omega}$ that assigns to each vertex of the model $\setV$ a corresponding pixel of $\setI$ in the image domain $\Omega$.
In order to allow the possibility to not assign an image pixel to a vertex (i.e. a vertex currently not visible), we extend the set $\Omega$ to also include $\emptyset$, which is denoted by $\bar{\Omega}$. 

\subsection{Dense Correspondence Encoding}
To obtain the vertex-to-pixel correspondence map $c$ we make use of a pixel-to-color mapping $\setN : \Omega \rightarrow [0,1]^{4}$
that assigns to each image pixel 
a 4-channel color value that encodes the correspondence. 
Here, the first 3 channels correspond to the dense correspondence on the hand surface (with the colors as shown in Fig.~\ref{fig:model} bottom right) and the last channel encodes the segmentation label (0: left hand, 0.5: right hand, 1: non-hand).
Due to this correspondence encoding in image space, it is more convenient to learn the pixel-to-color mapping $\setN$, compared to directly learning the vertex-to-pixel mapping $c$.
We emphasize that the function $\setN$ is defined over the entire input image domain and thus is able to predict color-encoded correspondence values for any pixel. As such, it contains correspondence information for \emph{both hands simultaneously} and thus implicitly learns to handle interactions between the left and the right hand. 
Please refer to \cref{sec:data-generation} on how we generate the training data and \cref{sec:neural-network-regressor} for details on how we learn $\setN$.

In order to associate the hand \emph{model vertices} in $\setV$ to image pixels in $\Omega$ based on the function $\setN$, we also define a vertex-to-color mapping $\setM: \setV \rightarrow [0,1]^4$, similar to recent dense regression works \cite{taylor2012vitruvian,wei2016dense,huang2016volumetric}.
Note that the output of $\setM$ for symmetric vertices in the left and right hand model only differs in the last component (0: left hand, 0.5: right hand), whereas the first three components encode the position on the hand surface as visualized in Fig.~\ref{fig:model} (bottom right).
Hence, the correspondences between vertices and pixels can be determined based on the similarity of the colors obtained by the mappings $\setN$ and $\setM$. 
In order to assign a color value $\setM(i)$ to the $i$-th model vertex, we first use multi-dimensional scaling~\cite{bronstein2006multigrid} for embedding the hand model into a three-dimensional space that (approximately) preserves geodesic distances. 
Subsequently, we map an HSV color space cylinder onto the embedded hand mesh such that different hues are mapped onto different fingers, cf.~Fig.~\ref{fig:model} (bottom right). 
As we later demonstrate (Fig.~\ref{fig:ablationPCK}), the proposed geodesic HSV embedding leads to improved results compared to a na\"{i}ve coloring (by mapping the RGB cube onto the original mesh, which is equivalent to regressing 3D vertex positions in the canonical pose).

\paragraph{Obtaining vertex-to-pixel mappings from the color encodings:}
To obtain a vertex-to-pixel correspondence map $c$ for a given depth image $\setI$, we first map the image pixels to colors using $\setN$ (a function over the image domain).
Subsequently, we compare the per-pixel colors obtained through $\setN$ with the fixed per-vertex colors $\setM$, from which the vertex-to-pixel maps are constructed by a thresholded nearest-neighbor strategy.
For all $i \in \setV$ we define
\begin{align}
    \hat{c}(i) &= \argmin\limits_{j \in \Omega} ~~\| \setN(j) - \setM(i) \|_2 \,,\text{ and}\\
    c(i) &= \begin{cases}
    \hat{c}(i)  \quad & \text{if } \| \setN(\hat{c}(i)) - \setM(i) \|_2 < \eta,
    \\
    \emptyset & \text{otherwise}\,.
    \end{cases}
\end{align} 
If the closest predicted color for some vertex $i$ is larger than the empirically chosen threshold $\eta{=}0.04$, this vertex is likely to be invisible in the input depth image.

\input{sections/data-generation.tex}

\paragraph{Real Data with Segmentation Annotation:}
When only trained with synthetic data, neural networks tend to overfit and hence may not generalize well to real test data.
To overcome this, we integrate real depth camera footage of hands into our so far synthetically generated training set.
Since it is infeasible to obtain dense correspondence annotations on real data, we restrict the annotation on real data to the left/right hand segmentation task.
As body paint \cite{tompson_tog2014,soliman2018fingerinput} has less influence on the observed hand shape, in contrast to colored gloves \cite{Taylor_siggraphasia2017}, we use body paint to obtain reliable annotations by color segmentation in the RGB image provided by the depth camera.
In total, we captured 3 users (1 female, 2 male) with varying hand shapes (width: 8--10cm, length: 17--20.5cm). 
We recorded $\approx 3,000$ images per subject and viewpoint (shoulder-mounted camera and frontal camera), resulting in a total number of 19,926 images. 

\subsection{Neural Network Regressor}
\label{sec:neural-network-regressor}
\begin{figure}
  \includegraphics[width=\linewidth,page=2,trim={0 3.5cm 0.5cm 5cm},clip]{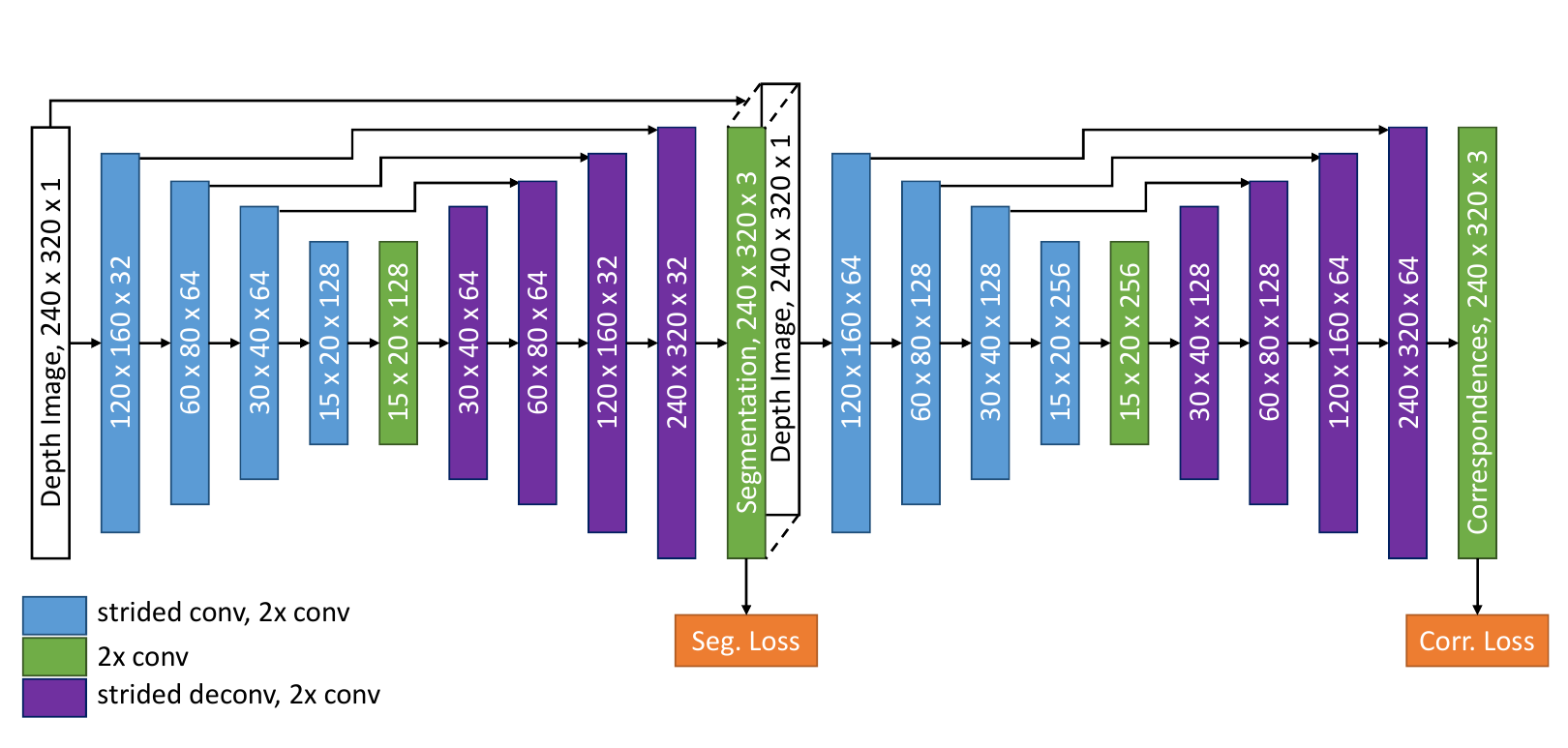}
  \caption
  {Our correspondence regression network (CoRN) consists of two stacked encoder-decoder networks. The output sizes of the layer blocks are specified as height $\times$ width $\times$ number of feature channels. In addition, the colors of the layer blocks indicate which operations are performed (best viewed in color). }
  \label{fig:cnn}
\end{figure}

Based on the mixed real and synthetic training data described in \cref{sec:data-generation}, we train a neural network that learns the pixel-to-color mapping $\setN$, as depicted in Fig.~\ref{fig:cnn}. 
Inspired by recent architectures used for per-pixel predictions \cite{unet2015,newell2016stacked}, our network comprises two stacked encoder-decoder processing blocks.
The first block is trained to learn the segmentation task, i.e.\:it outputs per-class probability maps in the original input resolution for the three possible classes $\{\lleft, \rright, \text{non-hand}\}$.   
These class probability maps are concatenated with the input depth image $\setI$ and fed into the second encoder-decoder to regress the 3-channel per-pixel hand surface correspondence information.
The final mapping $\setN: \Omega \rightarrow [0,1]^4$ is then obtained by concatenating the correspondence output with the label of the most likely class for each pixel.
Note that we scale the class labels to also match the range $[0,1]$ by setting $\lleft = 0$, $\rright = 0.5$, and $\text{non-hand} = 1$.
Both our encoder-decoder subnetworks share the same architecture.
We downsample the resolution using convolutions with stride 2 and upsample with the symmetric operation, deconvolutions with stride 2. Note that every convolution and deconvolution is followed by batch normalization and rectified linear unit (ReLU) layers.
In addition, we use skip connections to preserve spatially localized information and enhance gradient backpropagation.
Since the second subnetwork needs to learn a harder task, we double the number of features in all layers.
The segmentation loss is formulated as the softmax cross entropy, a standard classification loss.
For the correspondence loss, we use the squared Euclidean distance as commonly used in regression tasks.
We train the complete network end-to-end, with mixed data batches containing both synthetic and real samples in one training iteration.
For the latter, only the segmentation loss is active.

\section{Pose and Shape Estimation}\label{sec:energy}
The pose and shape of the hands present in image $\setI$ are estimated by fitting the hand surface model (Sec.~\ref{sec:hand-model}) to the depth image data. To this end, we first extract the foreground  point-cloud $\{ \vecd_j \in \R^3 \}_{j=1}^{N_I}$ in the depth image $\setI$, along with the respective point-cloud normals $\{ \vecn_j \in \R^3 \}_{j=1}^{N_I}$ obtained by Sobel filtering. Based on the assumption that the hands and arms are the objects closest to the camera, the foreground is extracted using a simple depth-based thresholding strategy, where $N_I$ denotes the total number of foreground pixels (of both hands together).
Subsequently, this point-cloud data is used in conjunction with the learned vertex-to-pixel correspondence map $c$ within an optimization framework. By minimizing a suitable nonlinear least-squares energy, which we will define next, the hand model parameters that best explain the point-cloud data are determined. 

The total energy function for both the left and the right hand is defined as
\begin{equation}
    \setE_{\total}(\beta, \theta) = \setE_{\data}(\beta, \theta) + \setE_{\reg}(\beta, \theta) \,,
\end{equation}
where $\beta$ are the shape parameters and $\theta$ are the hand pose parameters, as described in Sec.~\ref{sec:hand-model}. 

\subsection{Data Term} 
The data term $\setE_{\data}$ measures for a given parameter tuple $(\beta,\theta)$ how well the hand model explains the depth image $\setI$, and the term $\setE_{\reg}$ is a regularizer that accounts for temporal smoothness, plausible hand shapes and poses, as well as avoiding interpenetrations within and between the hands. We define the data term based on a combination of a point-to-point and a point-to-plane term as 
\begin{equation}
    \setE_{\data}(\beta, \theta) = \omega_{\point} E_{\point}(\beta, \theta) + \omega_{\plane} E_{\plane}(\beta, \theta)\,,
\end{equation}
where we use $\omega_{\odot}$ to denote the relative weights of the terms.

\paragraph{Point-to-point:} Let $\gamma_i$ be the visibility indicator  for the $i$-th vertex, which is defined to be $1$ if $c(i) \neq \emptyset$, and $0$ otherwise.
The point-to-point energy measures the distances between all visible model vertices $\vecv_i(\beta, \theta)$ and the \emph{corresponding} 3D point at pixel $c(i)$, and  is defined as
\begin{equation}
    E_{\point}(\beta, \theta) = \sum_{i=1}^{N_\setV} \gamma_i ||  \vecv_i(\beta, \theta) - \vecd_{c(i)} ||_2^2\,.
\end{equation}
\paragraph{Point-to-plane:}
The point-to-plane energy is used to penalize the deviation from the model vertices $\vecv_i(\beta, \theta)$ and the point-cloud surface tangent, and is defined as
\begin{equation}
    E_{\plane}(\beta, \theta) = \sum_{i=1}^{N_\setV} \gamma_i \: \langle \vecv_i(\beta, \theta) - \vecd_{c(i)}, \vecn_{c(i)} \rangle^2 \,.
\end{equation}

\subsection{Regularizer} 
Our regularizer $\setE_{\reg}$ comprises statistical pose and shape regularization terms, a temporal smoothness term, as well as a collision term. We define it as
\begin{align}
    \setE_{\reg}(\beta, \theta) = &\omega_{\shape} E_{\shape}(\beta) + \omega_{\pose} E_{\pose}(\theta) \\
    &\omega_{\temp} E_{\temp}(\beta, \theta) + \omega_{\coll} E_{\coll}(\beta, \theta) \,.
\end{align}

\paragraph{Statistical Regularizers:} As explained in Sec.~\ref{sec:hand-model}, the MANO model is parameterized in terms of a low-dimensional linear subspace obtained via PCA. Hence, in order to impose a plausible pose and shape at each captured frame, we use the Tikhonov regularizers
\begin{equation}
    E_{\shape}(\beta) = || \beta ||_2^2 \quad\text{and}\quad E_{\pose}(\theta) = || \theta ||_2^2 \,.
\end{equation}

\paragraph{Temporal Regularizer:}
In order to achieve temporal smoothness, we use a zero velocity prior on the shape parameters $\beta$ and the pose parameters $\theta$, i.e. we define
\begin{equation}
    E_{\temp}(\beta, \theta) = || \beta^{(t)} - \beta^{(t{-}1)} ||_2^2 + || \theta^{(t)} - \theta^{(t{-}1)} ||_2^2 \,.
\end{equation}

\paragraph{Collision Regularizer:} In order to avoid interpenetration between individual hands, as well as interpenetrations between the left and the right hand, we use a collision energy term. As described in Sec.~\ref{sec:hand-model}, we place spherical collision proxies inside each hand mesh, and then penalize overlaps between these collision proxies. Mathematically, we phrase this based on the overlap of (isotropic) Gaussians \cite{sridhar_cvpr2015}, which results in soft collision proxies defined as smooth occupancy functions.
The energy reads
\begin{equation}
    E_{\coll}(\beta,\theta) = \sum_{p=1}^{N_C}\sum_{q=p+1}^{N_C} \int_{\R^3} G_p(x; \beta,\theta) \cdot G_q(x; \beta,\theta) \, dx \,.
\end{equation}
Here, $G_p, G_q$ denote the Gaussian collision proxies whose mean and standard deviation depend on both the shape parameters $\beta$ as well as on the pose parameters $\theta$. 

\subsection{Optimization}
We have phrased the energy $\setE_\total$ in terms of a nonlinear least-squares formulation, so that it is amenable to be optimized based on the Gauss-Newton algorithm. All derivatives of the residuals can be computed analytically, so that we can efficiently compute all entries of the Jacobian on the GPU with high accuracy. More details on the GPU implementation can be found in the Appendix.

Note that although in principal it would be sufficient 
to optimize  for  the  shape parameter once per actor and then keep it fixed throughout the sequence, 
we perform the shape optimization in each frame of the sequence.
This has the advantage that a poorly chosen frame for estimating the shape parameter does not have a negative impact on the tracking of subsequent frames. 
We have empirically found that the hand shape is robust and does not significantly change throughout a given sequence. 

%% file: sections/data-generation.tex
\begin{figure}
\includegraphics[width=\linewidth,height=4cm]{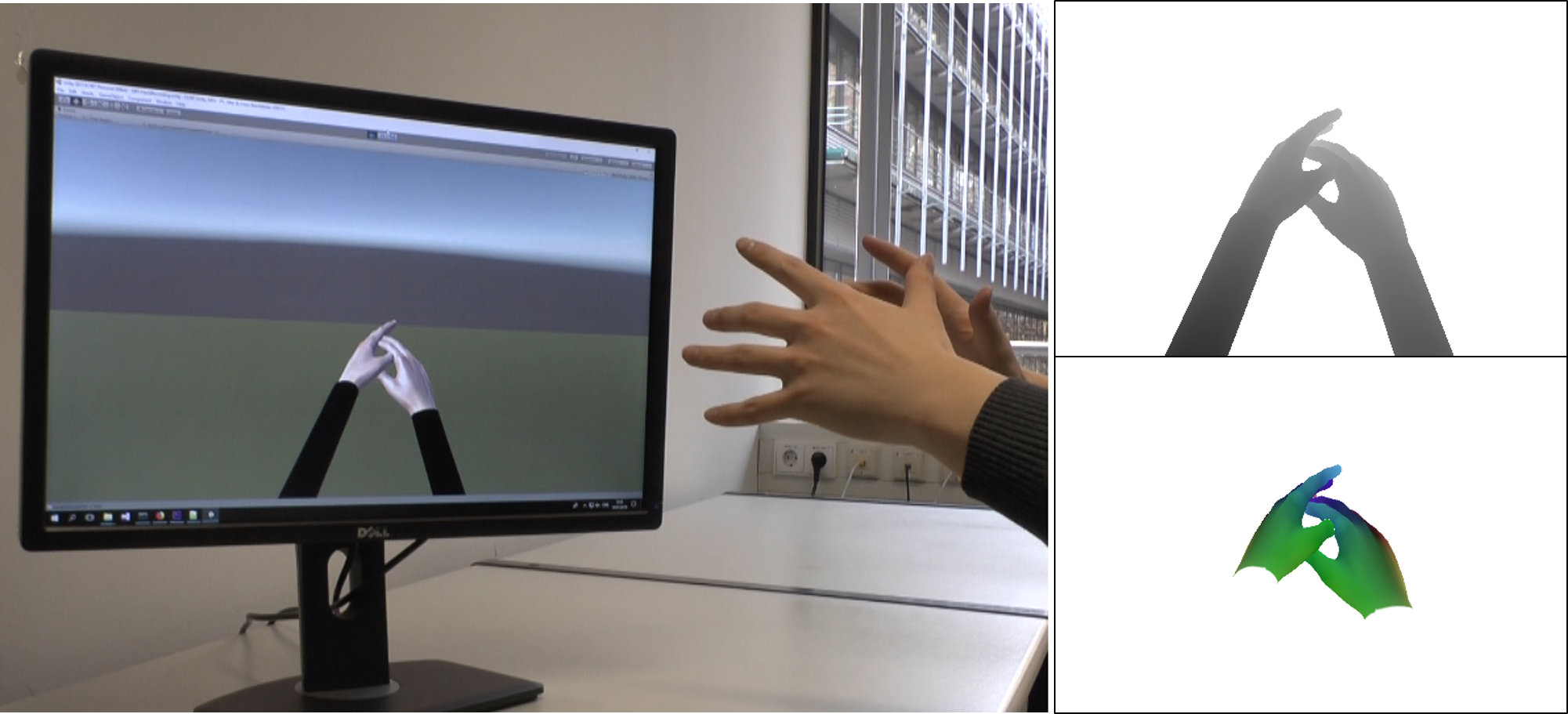}
	\caption
	{We generate our synthetic dataset by tracking two hands, separated by a safety distance, which are used to control in real-time a physically-based simulation of two interacting hands in the virtual scenario (left). We output the depth map (top right) and dense surface annotations (bottom right) of the resulting simulation.}
	\label{fig:data-generation}
\end{figure}

\subsection{Data Generation} \label{sec:data-generation}
In the following, we describe how we obtain suitable data to train the correspondence regression network.

\paragraph{Synthetic Data from Mocap-Driven Hand Simulation}
To overcome the challenge of generating training data with dense correspondences under complex hand-hand interactions we leverage a physics-based hand simulator, similar in spirit to \cite{Zhao_TOG2013}. To this end, we drive the simulation using skeletal hand motion capture (mocap) data \cite{leap_motion} to maximize natural hand motion. We tackle the issue that existing hand mocap solutions cannot robustly deal with close and complex hand-hand interactions by letting the actor move both hands at a safety distance from each other. This safety distance is subtracted in the simulation to produce closely interacting hand motions. By running the hand simulation in real time, the actor receives immediate visual feedback and is thus able to simulate natural interactions. 
Fig. \ref{fig:data-generation} depicts a live session of this data generation step. 

We extended the work of Verschoor et~al.~\shortcite{verschoor2018soft} by enabling simultaneous two hand simulation as well as inter-hand collision detection.
The hand simulator consists of an articulated skeleton surrounded by a low-resolution finite-element soft tissue model. The hands of the actor are tracked using Leap Motion~\shortcite{leap_motion}, and the mocap skeletal configuration is linked through viscoelastic springs (a.k.a. PD controller) to the articulated skeleton of the hand simulator. In this way, the hand closely follows the mocap input during free motion, yet it reacts to contact. The hand simulator resolves inter-hand collisions using a penalty-based frictional contact model, which provides smooth soft tissue interactions at minimal computational cost. We have observed that the soft tissue layer is particularly helpful at allowing smooth and natural motions in highly constrained situations such as interlocking fingers. 
As the hands are commanded by the mocap input, their motion is inherently free of intra-hand collisions. While inter-hand interaction may produce finger motions that lead to intra-hand collisions, we found those to be negligible for the training purposes of this step. We thus avoided self-collision handling to maintain real-time interaction at all times.

In practice, in this data generation step, we output a depth image for each simulated frame as well as the corresponding rendered image of the hand meshes colored with the  mapping $\setM$. 
Additionally, we postprocess the generated depth images to mimic typical structured-light sensor noise at depth discontinuities.
Using the above procedure, we recorded 5 users and synthesized $80{,}000$ images in total.

%% file: sections/evaluation.tex
\section{Evaluation}
In this section we thoroughly evaluate our proposed two-hand tracking approach. In Sec.~\ref{sec:implementation} we present additional implementation details. Subsequently, in Sec.~\ref{sec:ablation} we perform an ablation study, followed by a comparison to state-of-the-art tracking methods in Sec.~\ref{sec:sota}. Eventually, in Sec.~\ref{sec:moreresults} we provide additional results that demonstrate the ability of our method to adapt to user-specific hand shapes.

\subsection{Implementation}\label{sec:implementation}
Our implementation runs on two GPUs of type NVIDIA GTX 1080 Ti. One GPU runs the correspondence regression network CoRN, as well as the per-vertex correspondence matching for frame $t+1$, while the other GPU runs the model optimization for frame $t$. Overall, we achieve 30 fps using an implementation based on C++, CUDA, and the Tensorflow library.
We have used a depth camera Intel RealSense SR300 for our real-time results and evaluation. In Sec. \ref{sec:sota} we also demonstrate results when using a publicly available dataset that was captured with a different sensor.

Unless stated otherwise, for training CoRN we always use synthetic and real images (cf.~Sec.~\ref{sec:data-generation}) rendered and recorded from a \emph{frontal} view-point. We emphasize that it is reasonable to use view-specific correspondence regressors as for a given application it is usually known from which view-point the hands are to be tracked.

\subsection{Ablation Study}\label{sec:ablation}

\begin{figure}
	\centering
	\begin{subfigure}[b]{0.49\columnwidth}
		 \includegraphics[width=\linewidth,trim={2.5cm 7.5cm 3.5cm 8cm},clip]{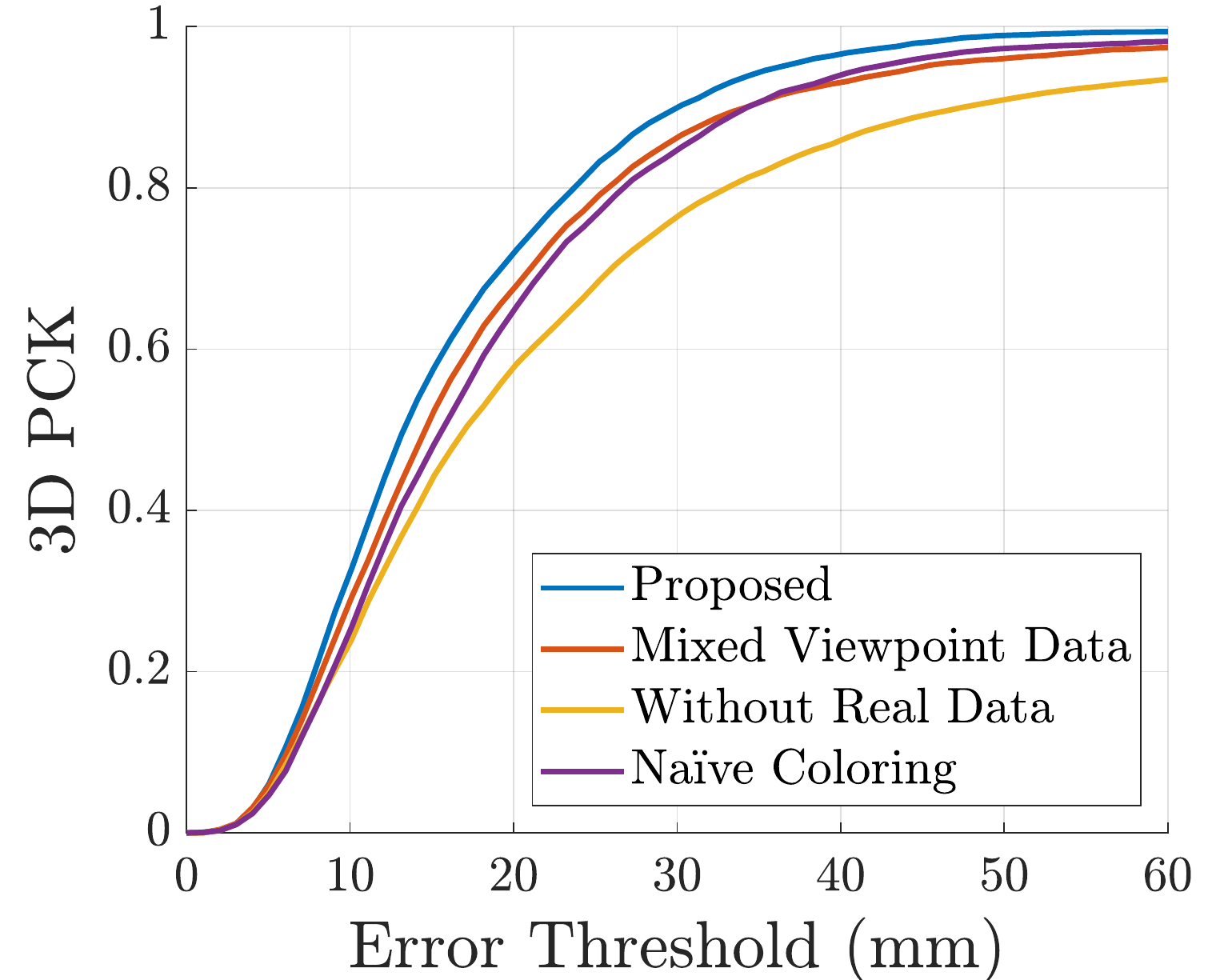}
		 \caption{}
		 \label{fig:ablationPCKa}
	\end{subfigure}
 	\begin{subfigure}[b]{0.49\columnwidth}
  		\includegraphics[width=\linewidth,trim={2.5cm 7.5cm 3.5cm 8cm}, clip]{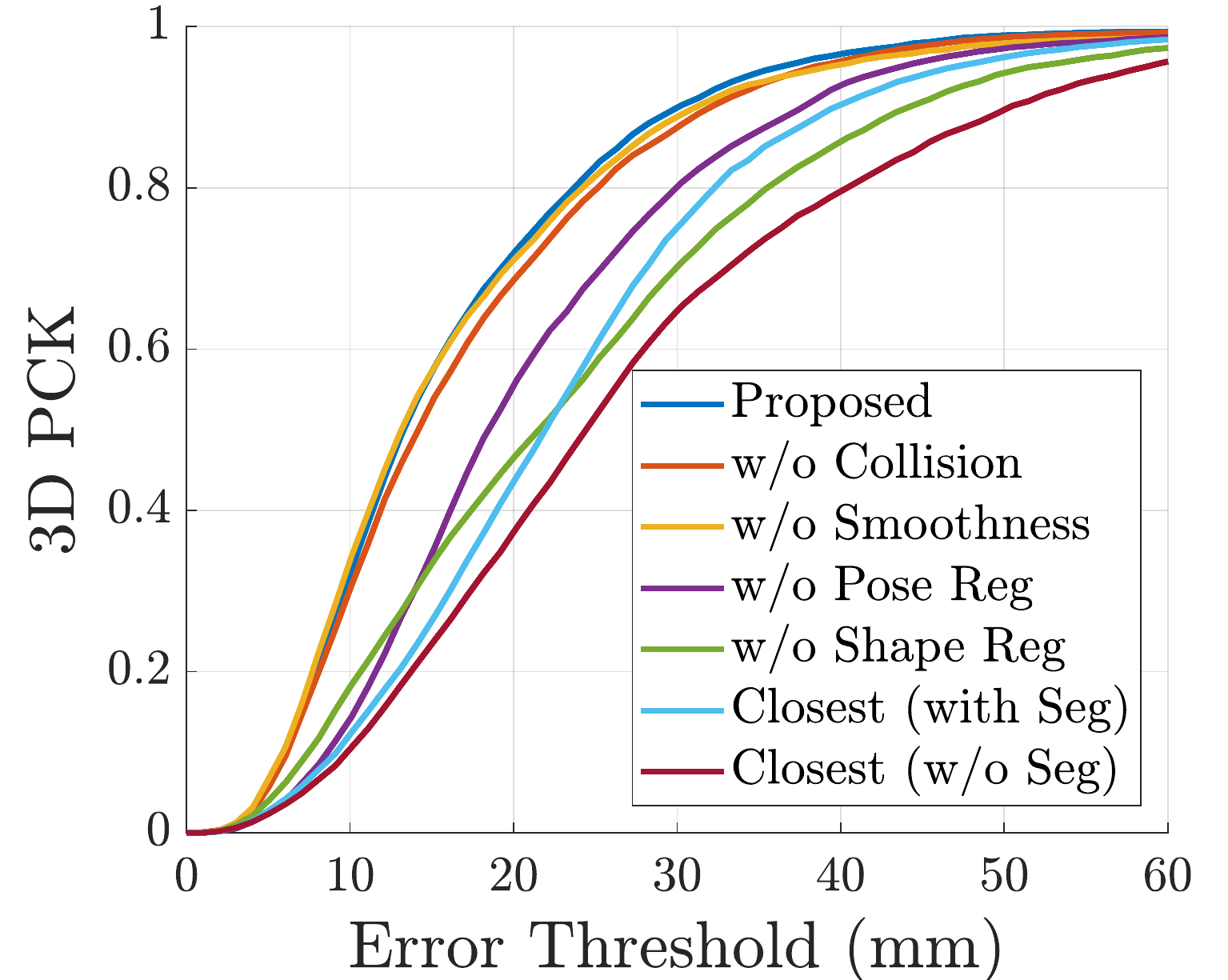}
		\caption{}
		\label{fig:ablationPCKb}	
\end{subfigure}
  \caption{Results of our ablation study. (a) shows different configurations regarding the correspondence regressor (CoRN). (b) shows configurations regarding the optimizer.}
  \label{fig:ablationPCK}
\end{figure}

\begin{figure}
  \includegraphics[width=\linewidth,page=5,trim={0 3cm 0.5cm 3cm},clip]{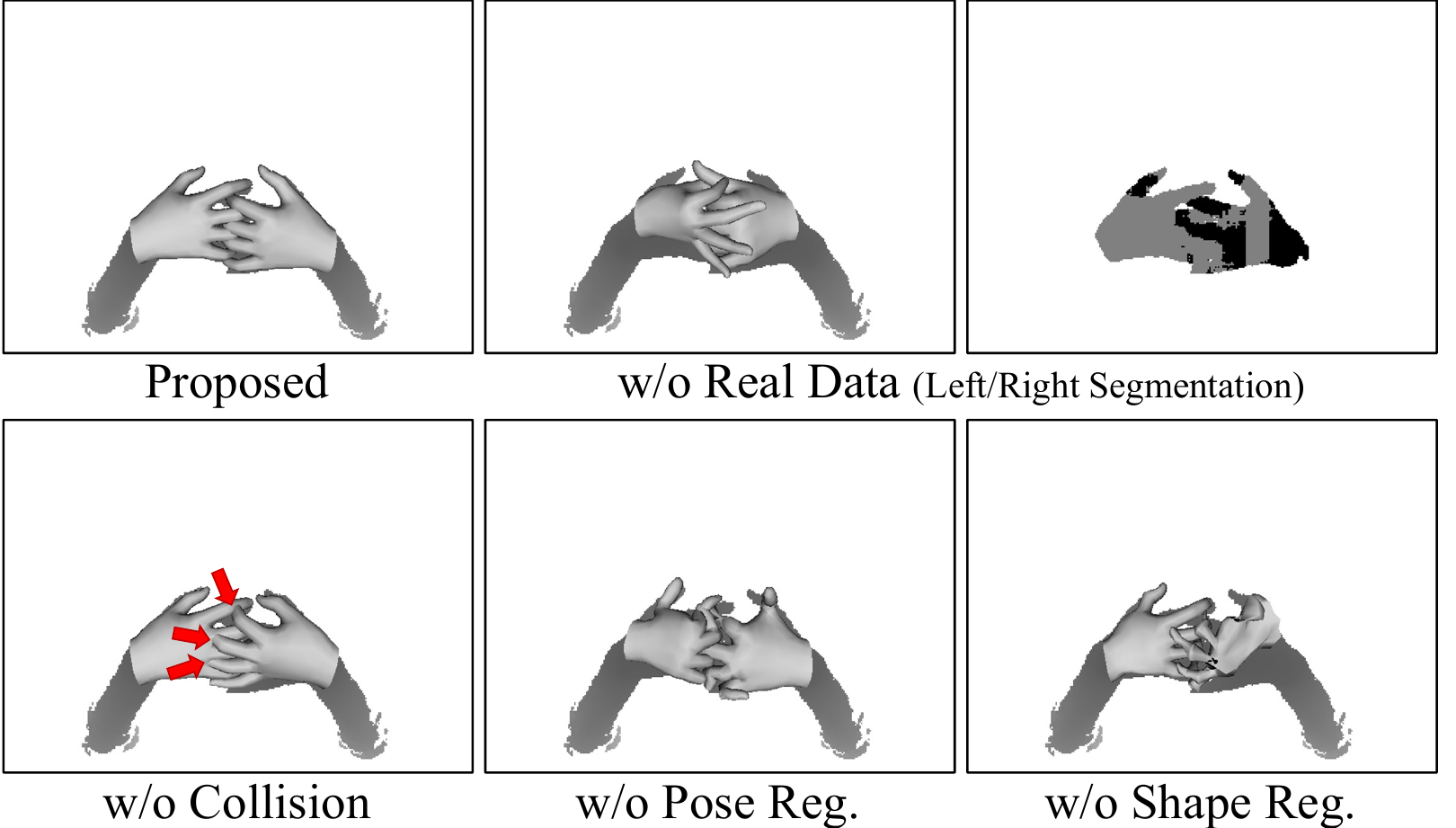}
  \caption{
  Qualitative examples from our ablation study. 
  }
  \label{fig:ablation}
\end{figure}

We have conducted a detailed ablation study, where we analyze the effects of the individual components of the proposed approach. For these evaluations we use the dataset provided by Tzionas et al.~\shortcite{tzionas_ijcv2016}, which comes with annotations of the joint positions on the depth images.

In Fig.~\ref{fig:ablationPCK} we show quantitative results of our analysis for a range of different configurations. To this end, we use the \emph{percentage of correct keypoints} (PCK) as measure, where the horizontal axis shows the error, and the vertical axis indicates the percentage of points that fall within this error. To compute the PCK, we consider the same set of keypoints as Tzionas et al.~\shortcite{tzionas_ijcv2016}.
Notice that despite using Tzionas et al.'s dataset, Fig.~\ref{fig:ablationPCK} does not show their results because they do not provide 3D PCK values. Qualitative results of our ablation study are shown in Fig.~\ref{fig:ablation}.

\paragraph{Correspondence Regression Network} In the Fig.~\ref{fig:ablationPCKa} we show four settings of different configurations for training the correspondence regressor (CoRN): 
\begin{enumerate}
    \item The proposed CoRN network as explained in Sec.~\ref{sec:dense-correspondeces} (blue line, ``Proposed'').
    \item The CoRN network but trained based on data from two viewpoints, egocentric as well as frontal (orange line, ``Mixed Viewpoint Data''). 
    \item The CoRN network that is trained only with synthetic data, i.e.\:we do not use real data as described in Sec.~\ref{sec:data-generation} in order to train the segmentation sub-network (yellow line, ``Without Real Data'').
    \item Instead of using our proposed geodesic HSV embedding as color-encoding for the correspondences (cf.~Fig.~\ref{fig:model}), we use a na\"{i}ve color-encoding by mapping the original mesh onto the RGB cube (purple line, ``Na\"{i}ve Coloring'').
\end{enumerate}
It can be seen that the proposed training setting outperforms all other settings.

\paragraph{Pose and Shape Estimation} 
In  Fig.~\ref{fig:ablationPCKb} we show different optimizer configurations.
We evaluate five versions of the energy:
\begin{enumerate}
    \item The complete energy $\setE_{\total}$ that includes all terms (blue line, ``Proposed'').
    \item The energy without the collision term $E_\coll$ (orange line, ``w/o Collision'').
    \item The energy without the temporal smoothness term $E_{\temp}$ (yellow line, ``w/o Smoothness'').
    \item The energy without the pose regularizer $E_{\pose}$ (purple line, ``w/o Pose Reg'').
    \item The energy without the pose regularizer $E_{\shape}$ (green line, ``w/o Shape Reg'').
\end{enumerate}
In addition, to demonstrate the importance of CoRN, we compare to two configurations using closest point correspondences instead:
\begin{enumerate}
    \item Finding the vertex correspondence as the closest input point that was classified with the same handedness (light blue line, ``Closest (with Seg)'').
    \item Finding the vertex correspondence as the closest input point in the whole point cloud (dark red line, ``Closest (w/o Seg)'').
\end{enumerate}
Note that we initialized the hand models manually as close as possible in the first frame to enable a fair comparison. We emphasize that this is not necessary with CoRN.

We can observe that the complete energy performs best, compared to leaving individual terms out. Moreover, we have found that removing the pose regularizer or the shape regularizer worsens the outcome significantly more compared to dropping the collision or the smoothness terms when looking at the PCK. We point out that the smoothness term removes temporal jitter that is only marginally reflected by the numbers. Similarly, while removing the collision term does not affect the PCK significantly, in the supplementary video we demonstrate that this severely worsens the results.
Using na{\"i}ve closest points instead of predicted CoRN correspondences results in significantly higher errors, this holds for both versions, with and without segmentation information.
Additionally, in Figure \ref{fig:ablation} we show qualitative examples from our ablation study that further validate that each term of the complete energy formulation is essential to obtain high quality tracking of hand-hand interaction.

\paragraph{Independence of Initialization:} In the supplemental video we also show results where our hand tracker is able to recover from severe errors that occur when the hand motion is extremely fast, so that the depth image becomes blurry. In this scenario, as soon as the hand moves with a normal speed again, the tracker is able to recover and provide an accurate tracking. Note that this is in contrast to local optimization approaches (e.g.\:based on an ICP-like procedure for pose and shape fitting) that cannot recover from bad results due to severe non-convexity of the energy landscape.

\subsection{Comparison to the State of the Art}\label{sec:sota}
Next, we compare our method with state-of-the-art methods.

\begin{figure}
  \includegraphics[width=\linewidth,page=4,trim={0cm 5cm 0 2.7cm},clip]{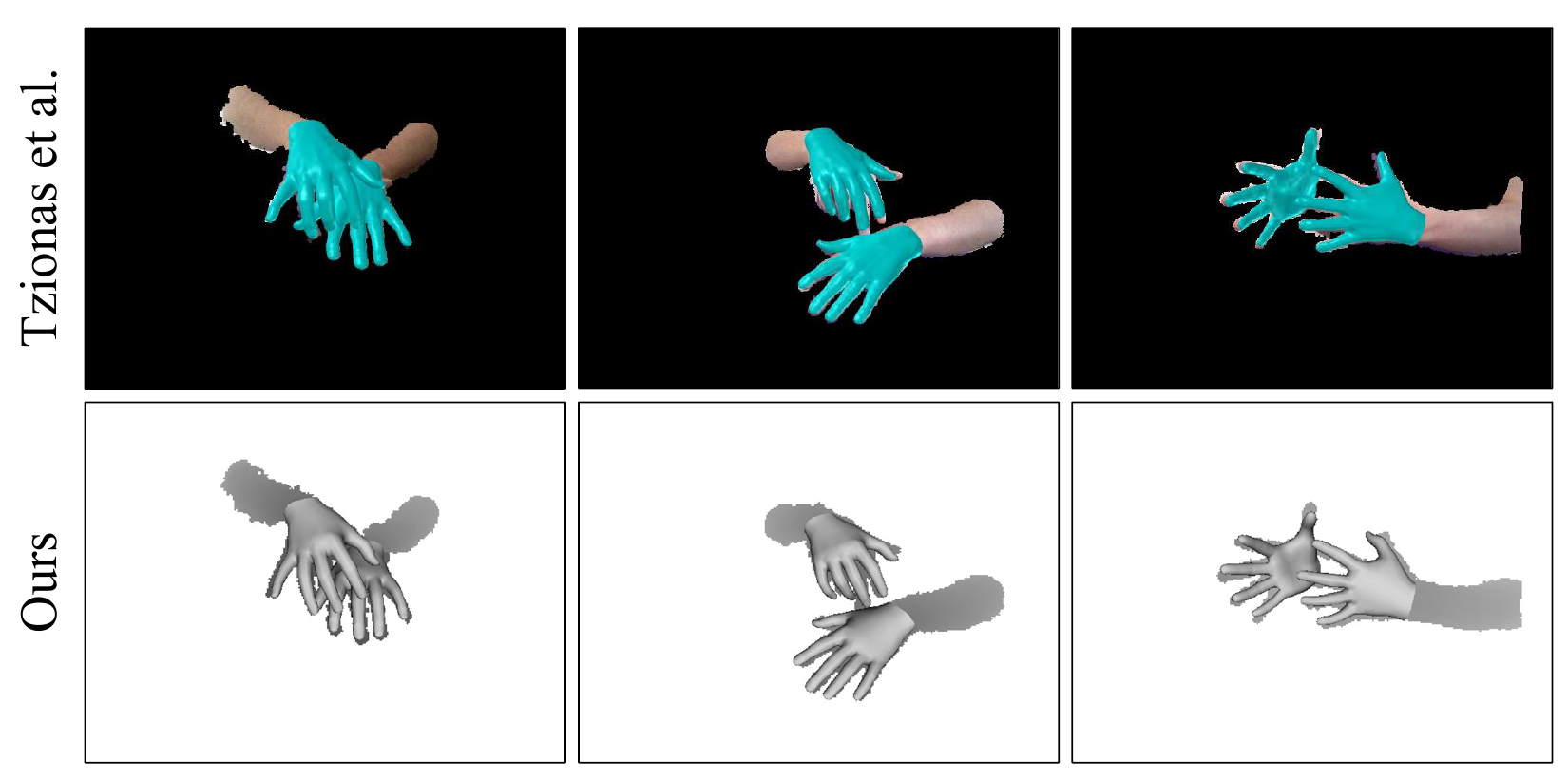}
  \caption{Qualitative comparison with \cite{tzionas_ijcv2016}. Our method achieves results with comparable visual quality while running multiple orders of magnitude faster.}
  \label{fig:comp_tzionas}
\end{figure}

\begin{figure}
  \includegraphics[width=\linewidth,page=1,trim={0cm 8.2cm 0.5cm 3cm},clip]{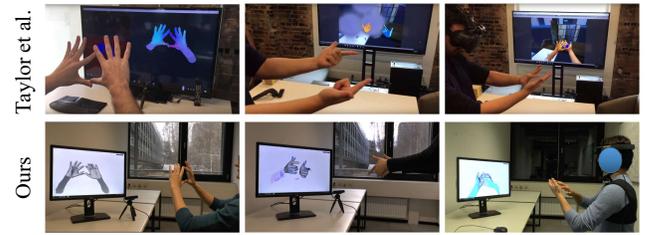}
  \caption{Qualitative comparison with \cite{Taylor_siggraphasia2017}. Our method is able to track two hands in similar poses while at the same time reconstructing shape automatically and avoiding collisions. }
  \label{fig:comp_taylor}
\end{figure}

\paragraph{Comparison to Tzionas~et~al.~\shortcite{tzionas_ijcv2016}} 
In Table \ref{table:tsionas} we present results of our quantitative comparison to the work of Tzionas~et~al.~\shortcite{tzionas_ijcv2016}. The evaluation is based on their two-hand dataset that comes with  joint annotations.
As shown, the relative 2D pixel error is very small in both methods. While it is slightly higher with our approach, we emphasize that we achieve a 150$\times$ speed-up and do not require a user-specific hand model.
Furthermore, in Fig. \ref{fig:comp_tzionas} we qualitatively show that the precision error difference does not result in any noticeable visual quality gap. Moreover, we point out that the finger tip detection method of Tzionas~et~al.~\shortcite{tzionas_ijcv2016}  is ad-hoc trained for their specific camera, whereas our correspondence regressor has never seen data from the depth sensor used in this comparison.

\paragraph{Comparison to Leap Motion~\shortcite{leap_motion}} In the supplementary video we also  compare our method qualitatively with the skeletal tracking results using the commercial solution \cite{leap_motion}.
As shown, while Leap Motion successfully tracks two hands when they are spatially separated by a significant offset, it struggles and fails for complex hand-hand interactions. In contrast, our approach is able to not only successfully track challenging hand-hand interactions, but also estimate the 3D hand shape.

\paragraph{Other Methods} Since the authors of \cite{Taylor_siggraphasia2017} did not release their dataset, we were not able to directly compare with their results. Nevertheless, in Fig.~\ref{fig:comp_taylor} and in the supplementary video we show tracking results on similar scenes, as well as some settings that are arguably more challenging than theirs.

\definecolor{ourgray}{gray}{0.4}
\begin{table}[]
\caption{
We compare our method to the method by Tzionas~et~al.~\shortcite{tzionas_ijcv2016} on their provided dataset. We show the average and standard deviation of the 2D pixel error (relative to the diagonal image dimension), as well as the per-frame runtime. Note that the pixel errors of both methods are very small, and that our method is $150\times$  faster. Moreover, our approach automatically adjusts to the user-specific hand shape, whereas  Tzionas et al. require a 3D scanned hand model.
}
\begin{tabular}{lccc}
\toprule
 & \textbf{2D Error} & \textbf{Runtime} & \textbf{Shape Estimation} \\ \midrule
\textbf{Ours} & $1.35{\pm}0.28\:\%$ & $33$ms & \cmark \\ 
Tzionas et al. & $0.63{\pm}0.12\:\%$ & $4960$ms &  \xmark \\ \bottomrule
\end{tabular}
\label{table:tsionas}
\end{table}

\subsection{More Results} \label{sec:moreresults}
In this section we present additional results on hand shape adaption as well as additional qualitative results.

\begin{figure}
  \includegraphics[width=\linewidth]{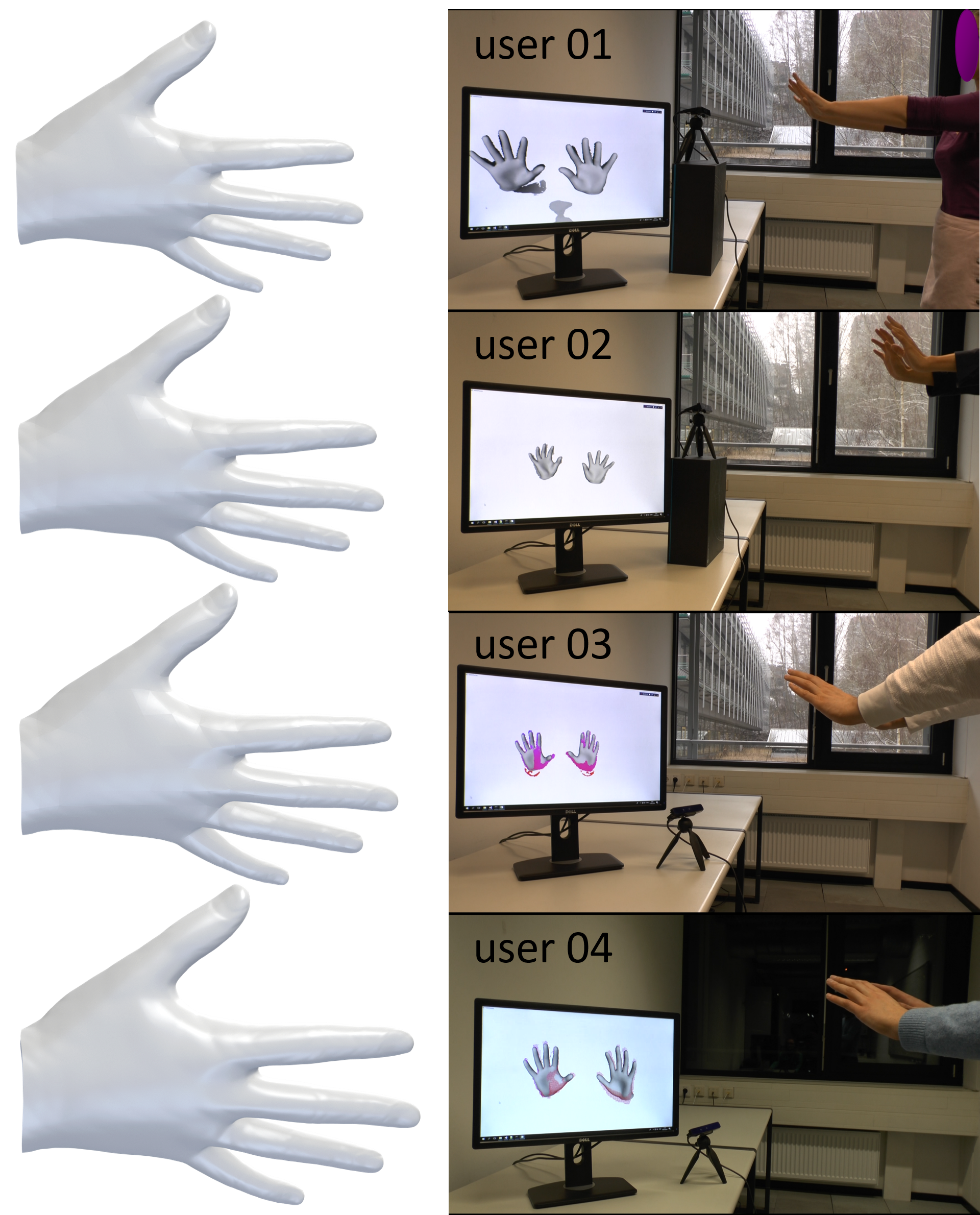}
  \caption{We present the 3D hand models (left) that we obtained from fitting our model to different users with varying hand shape. From top to bottom we show small to large hand shapes. Note that we show all four hand shapes on the left in the same pose in order to allow for a direct comparison.}
  \label{fig:shape_adaptation}
\end{figure}

\paragraph{Hand Shape Adaptation}
Here, we investigate the adaptation to user-specific hand shapes.
In Fig.~\ref{fig:shape_adaptation} we show the obtained hand shape when running our method for four different persons with varying hand shapes. It can be seen that our method is able to adjust the geometry of the hand model to the users' hand shapes. 

Due to the severe difficulty in obtaining reliable 3D ground truth data and disentangling shape and pose parameters, we cannot quantitatively evaluate shape directly. Instead, we additionally evaluate the consistency of the estimated bone lengths on the sequences of Tzionas et al.~\shortcite{tzionas_ijcv2016}. The average standard deviation is 0.6 mm, which indicates that our shape estimation is stable over time.

\begin{figure*}[ht!]
\centerline{
  \frame{\includegraphics[width=0.19\linewidth]{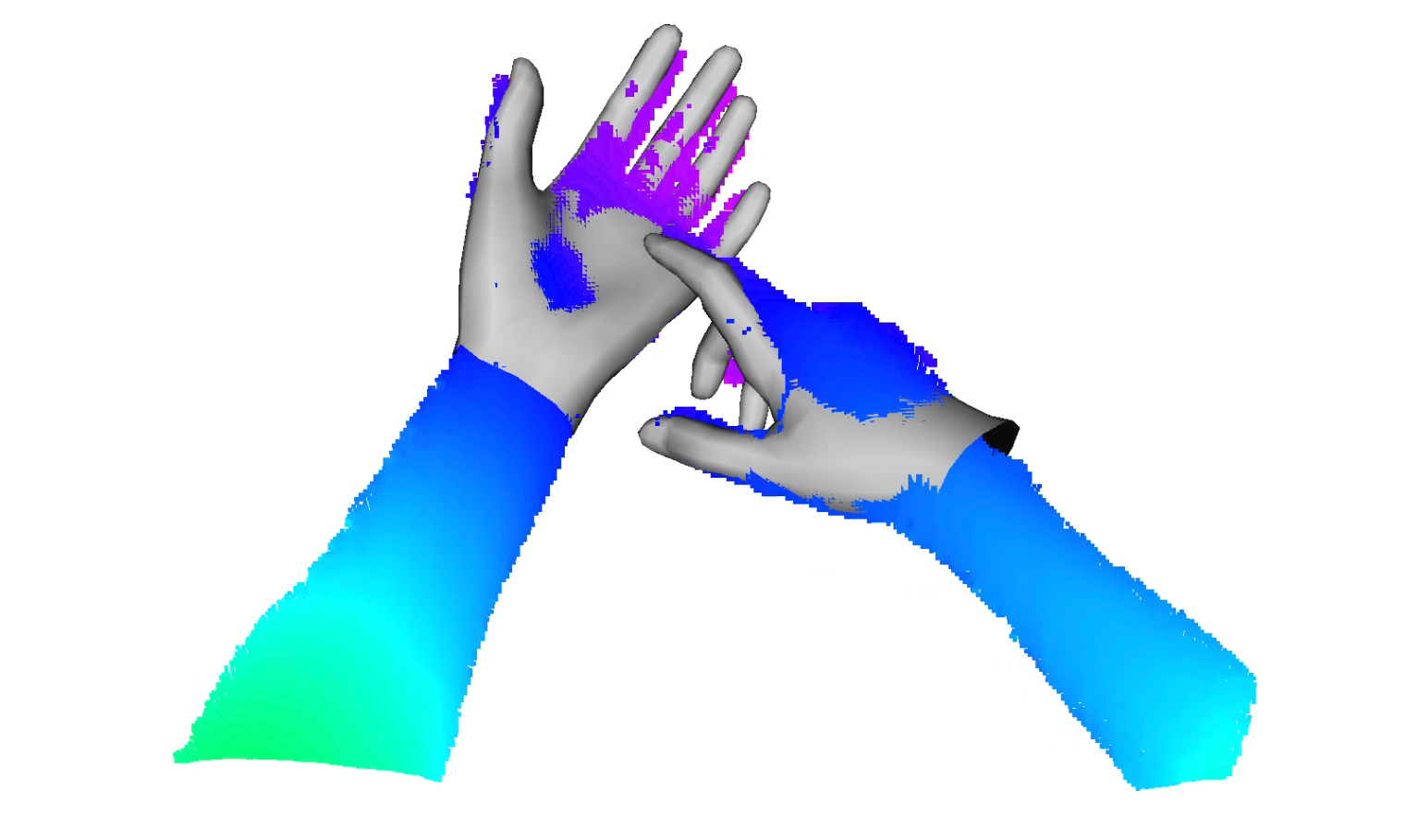}}
  \hfill
  \frame{\includegraphics[width=0.19\linewidth]{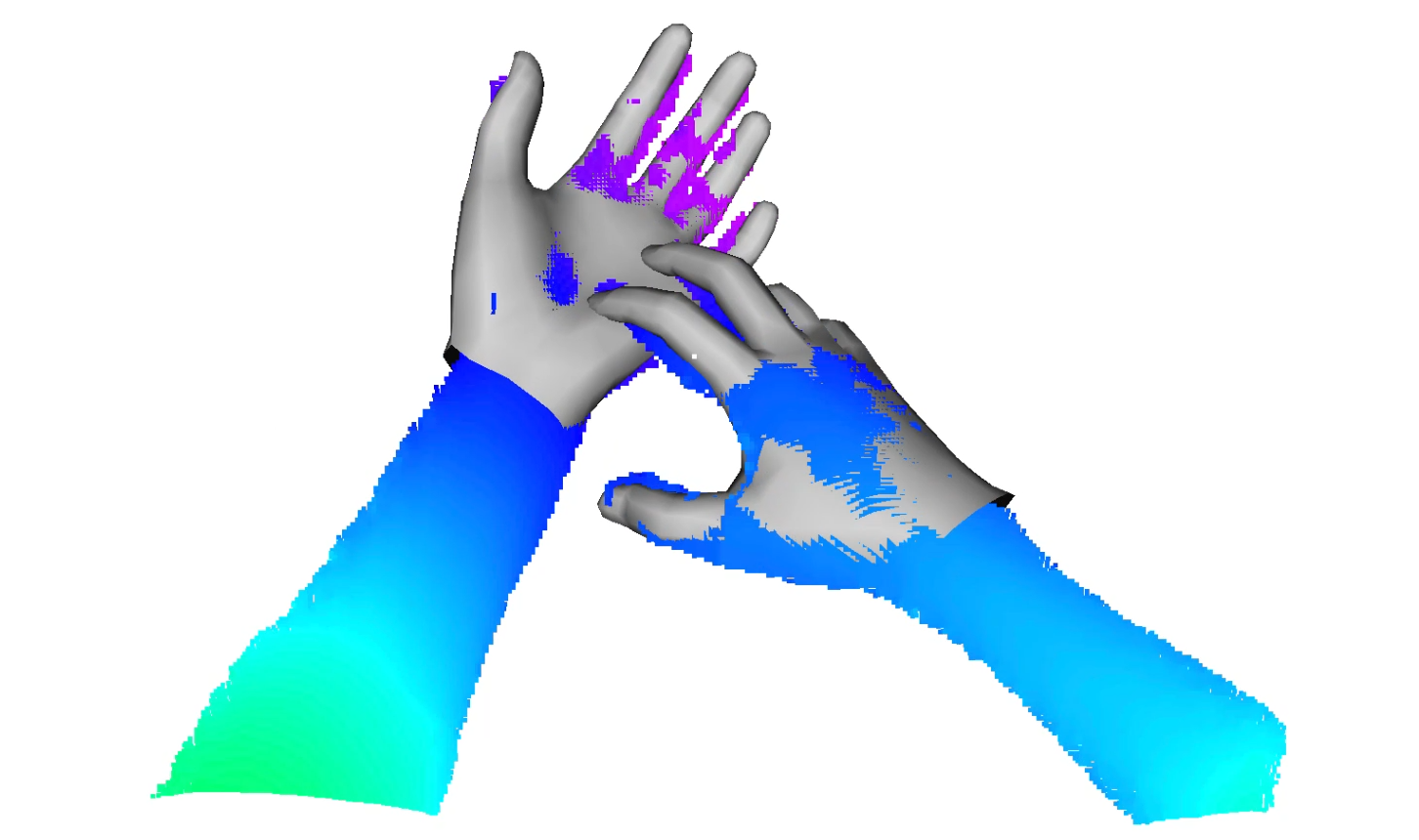}}
  \hfill
  \frame{\includegraphics[width=0.19\linewidth]{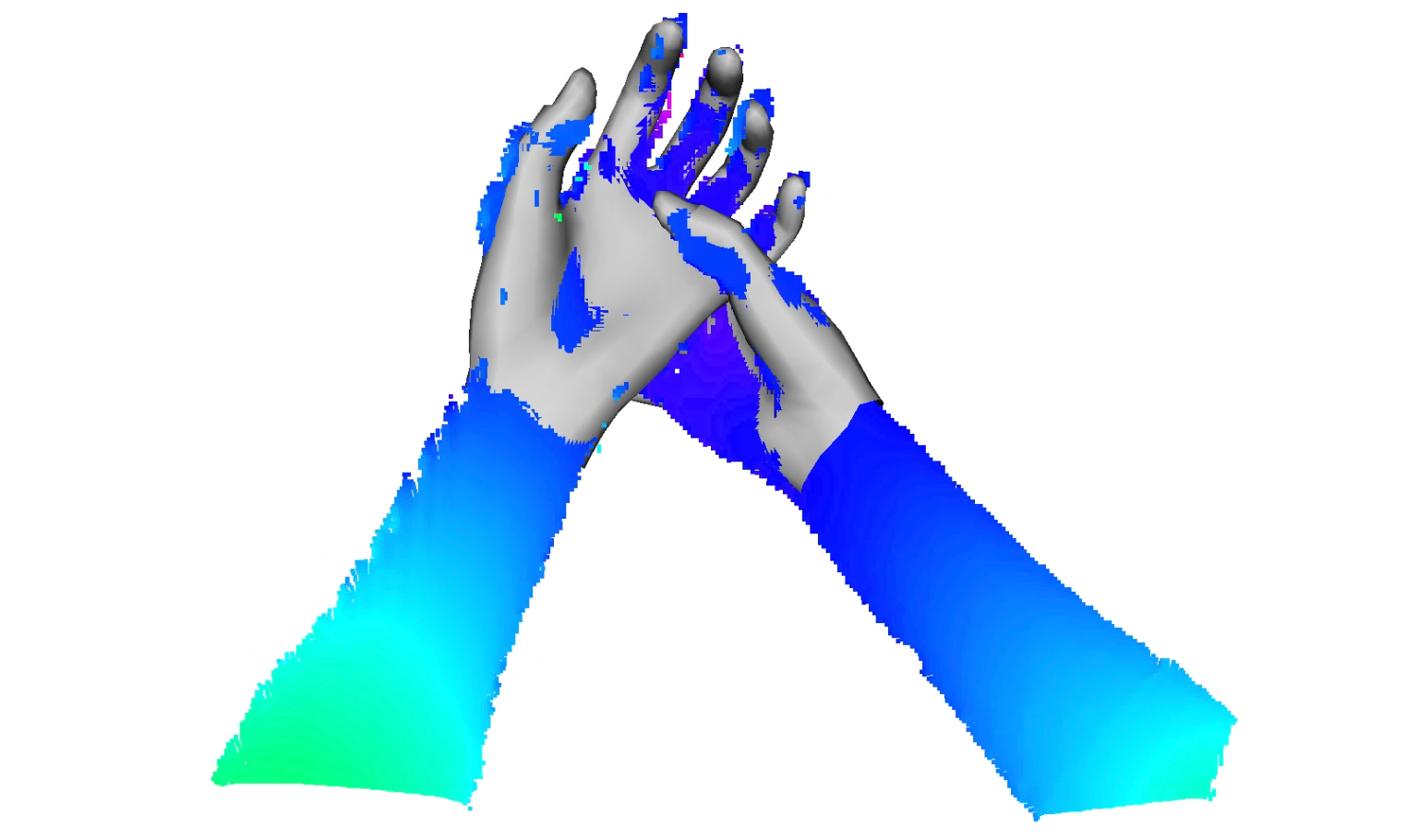}}
  \hfill
  \frame{\includegraphics[width=0.19\linewidth]{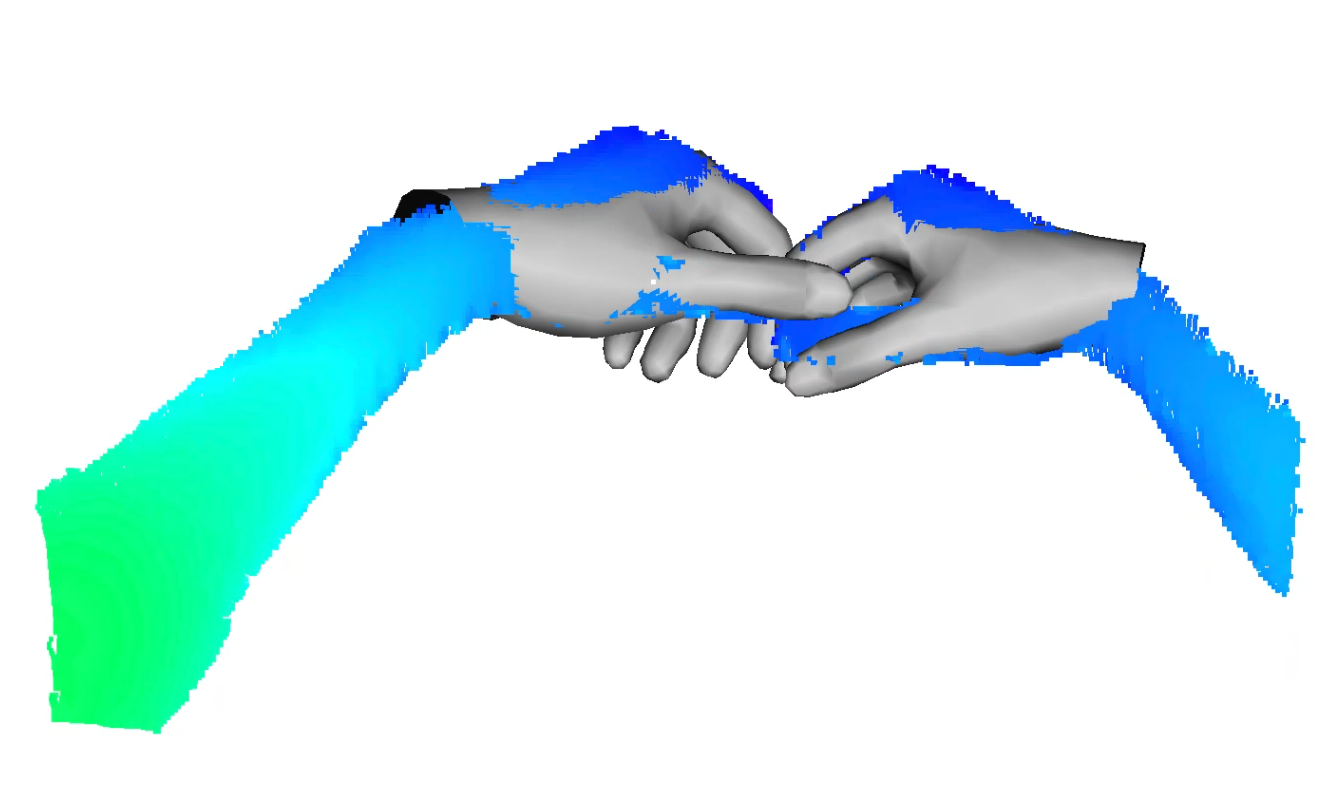}}
  \hfill
  \frame{\includegraphics[width=0.19\linewidth]{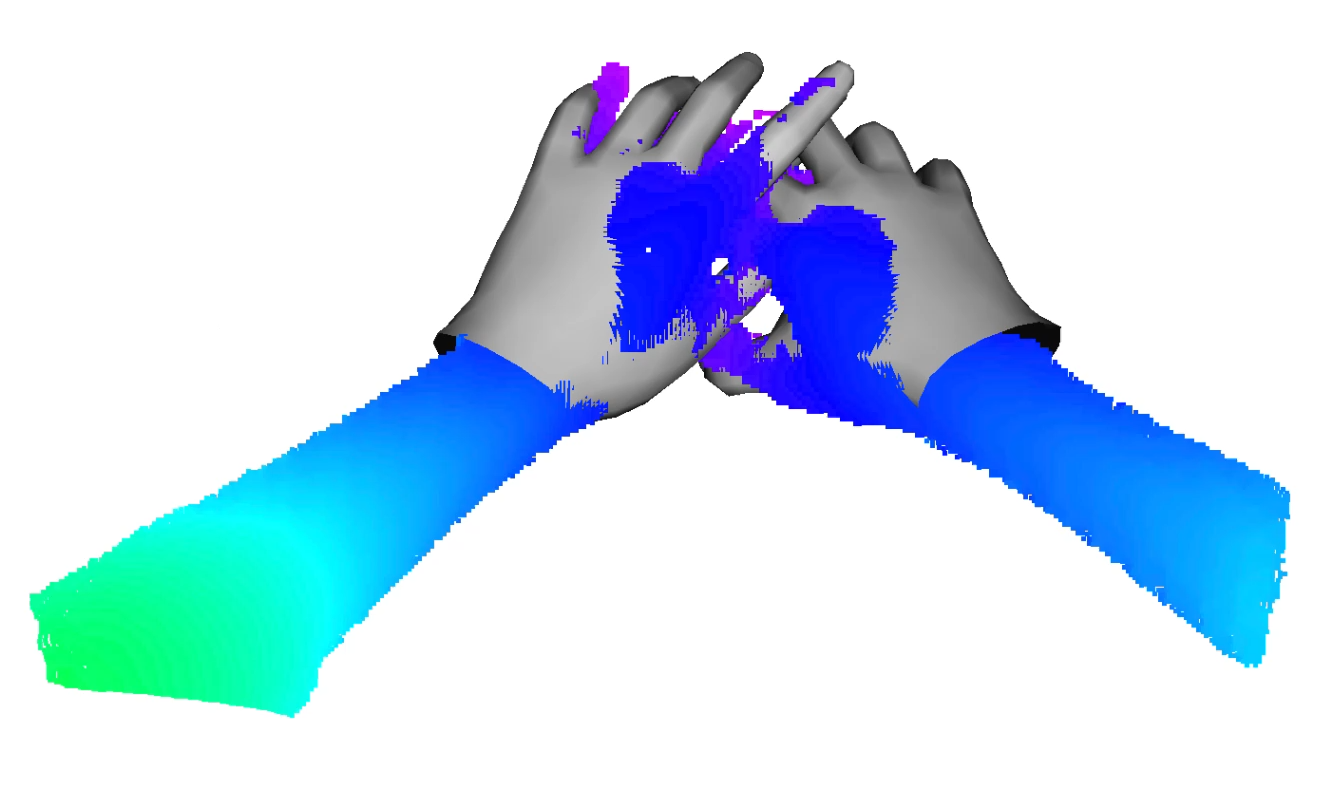}}
}
\vspace{0.1mm}
\centerline{
  \frame{\includegraphics[width=0.19\linewidth]{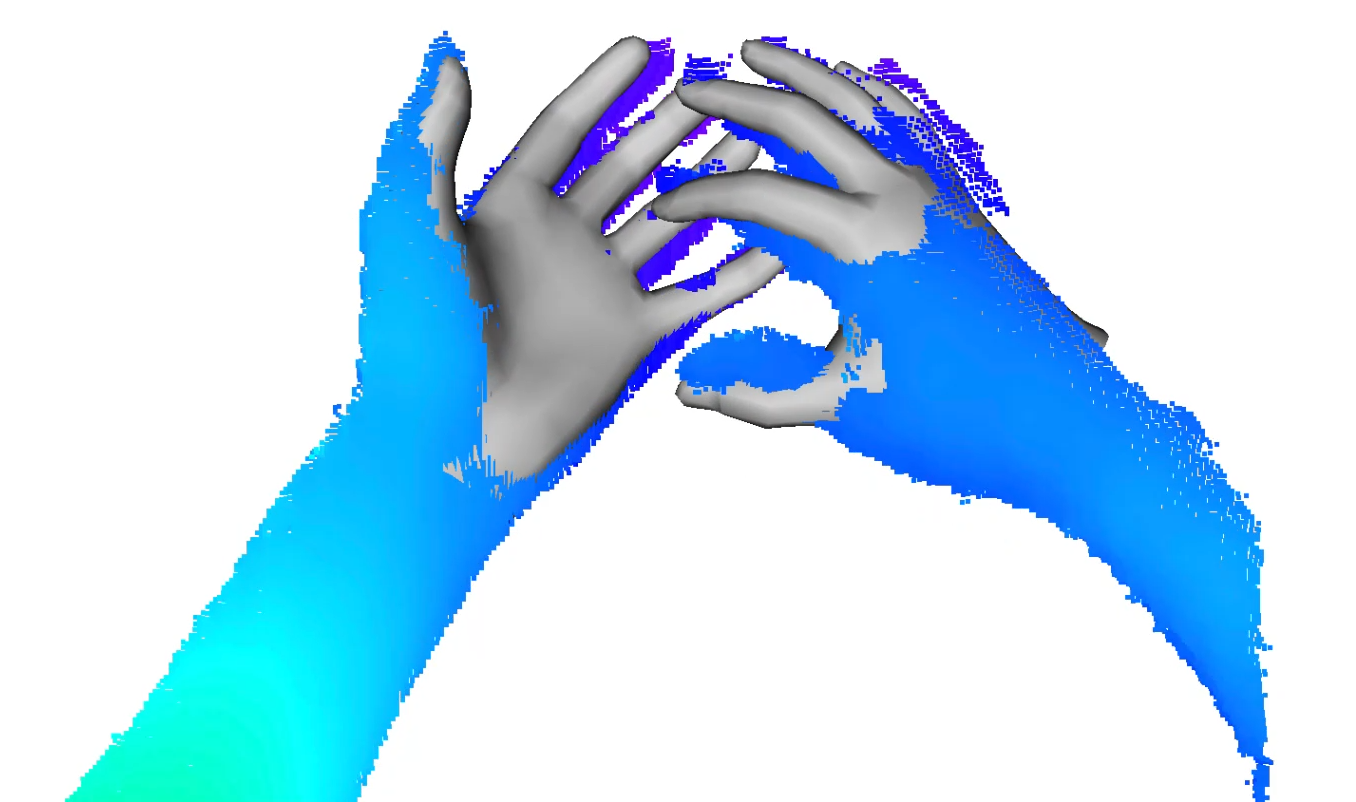}}
  \hfill
  \frame{\includegraphics[width=0.19\linewidth]{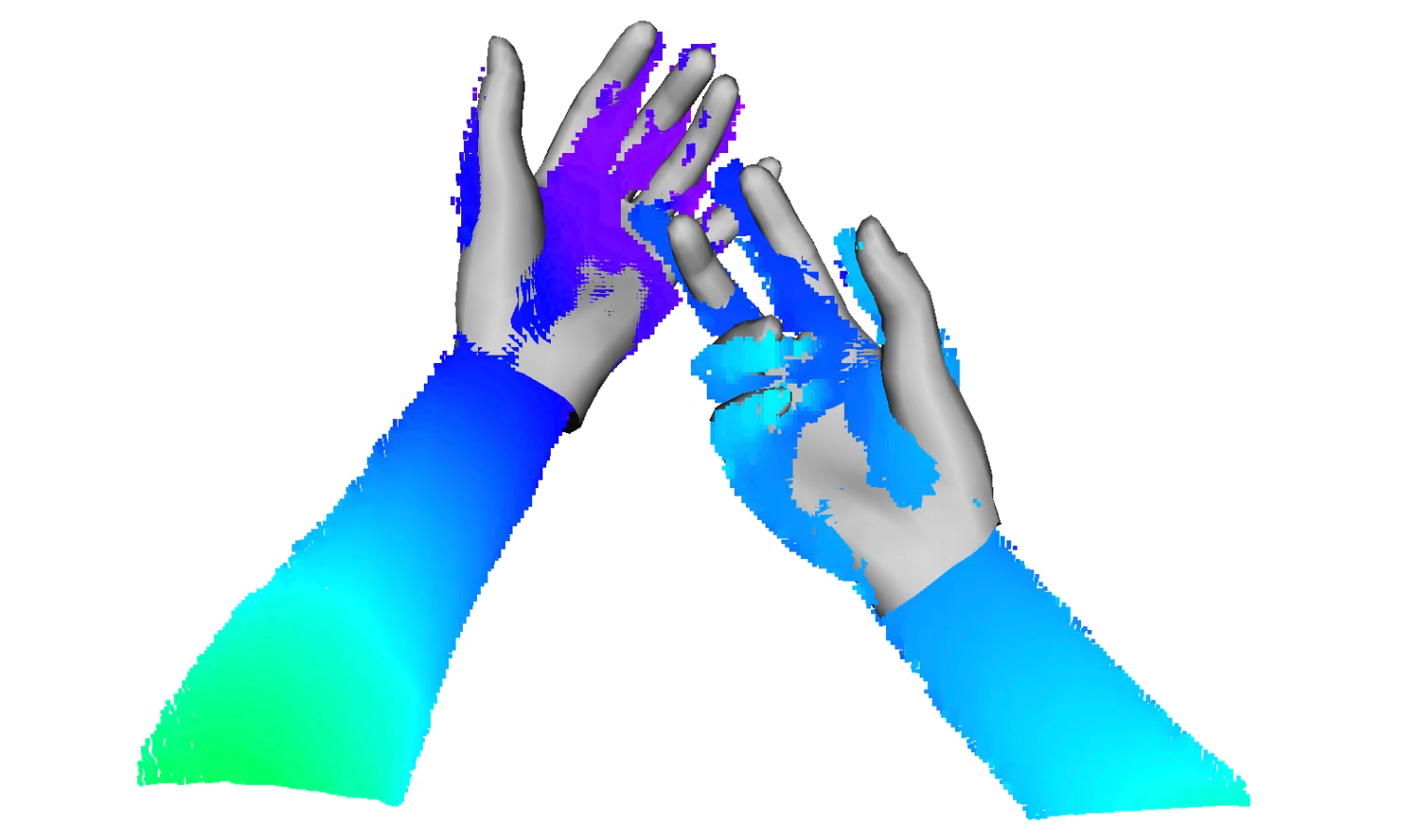}}
  \hfill
  \frame{\includegraphics[width=0.19\linewidth]{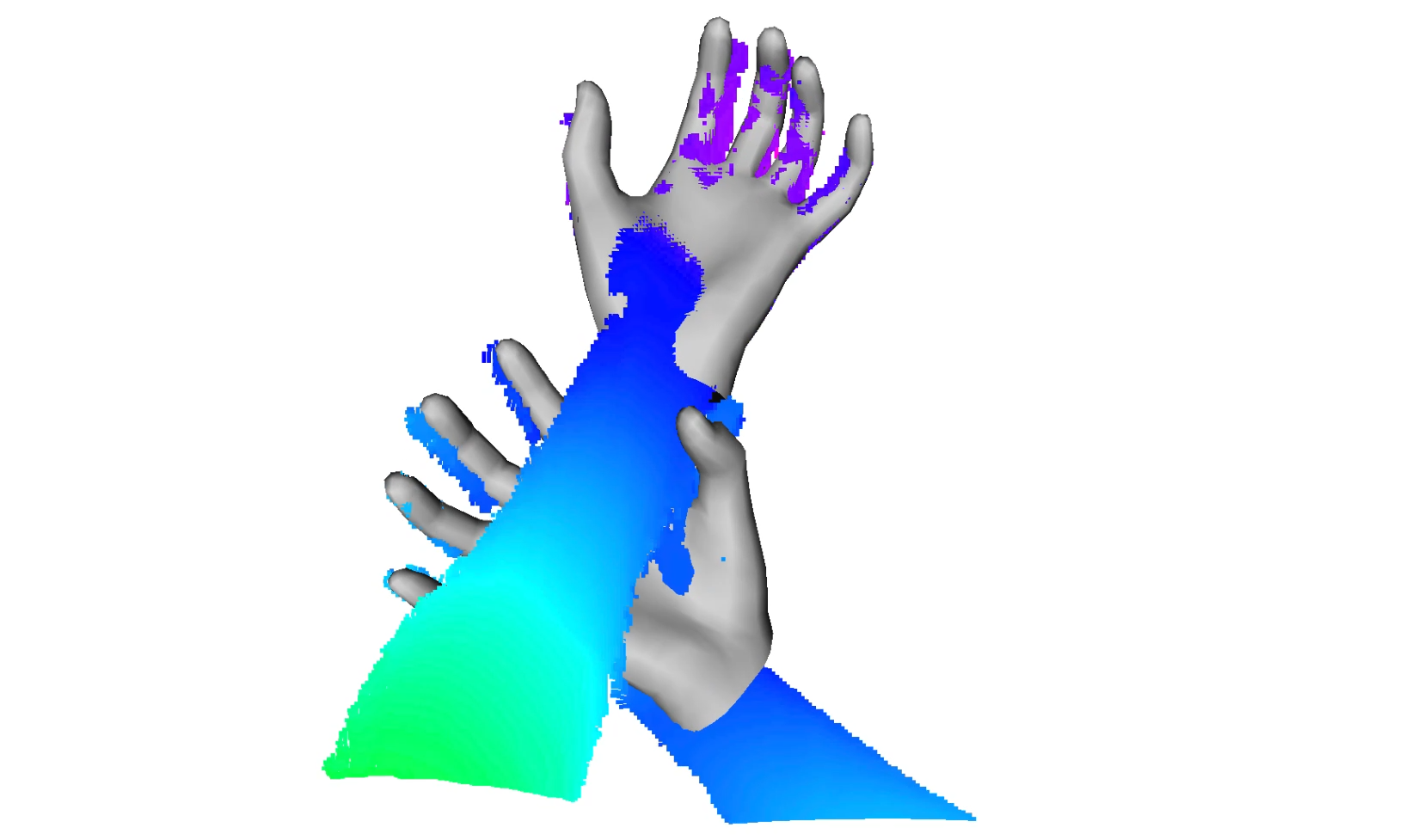}}
  \hfill
  \frame{\includegraphics[width=0.19\linewidth]{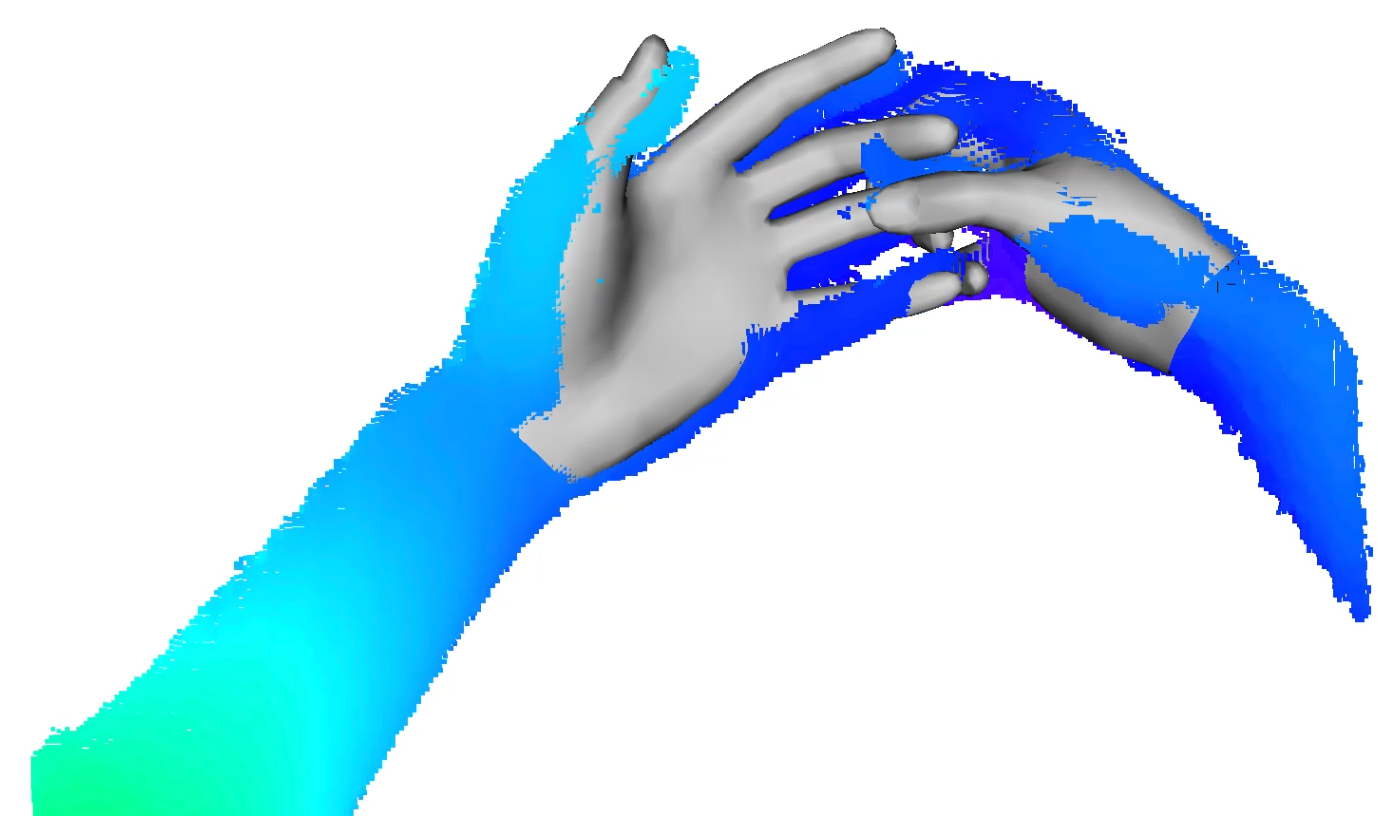}}
  \hfill
  \frame{\includegraphics[width=0.19\linewidth]{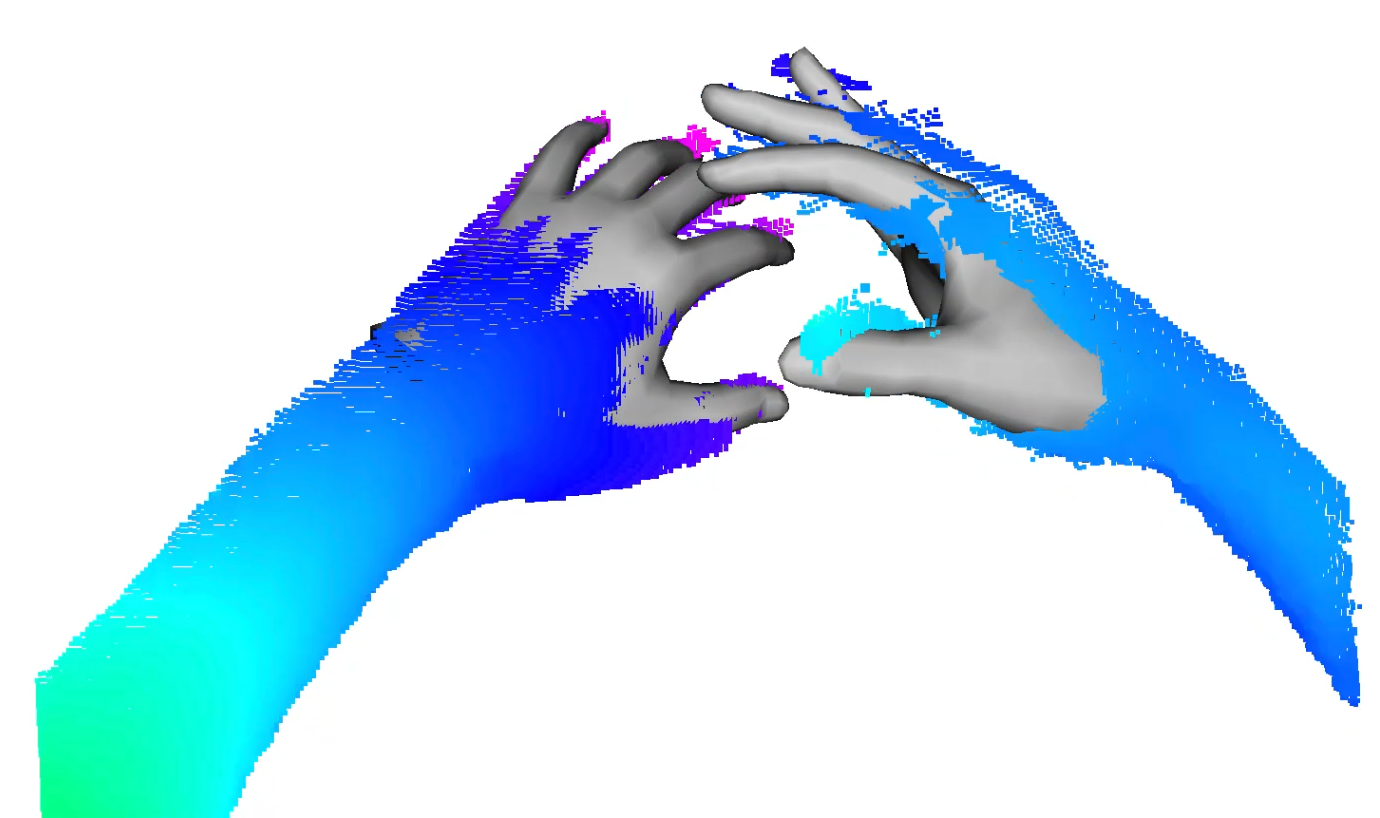}}
}
\vspace{0.1mm}
\centerline{
  \frame{\includegraphics[width=0.19\linewidth]{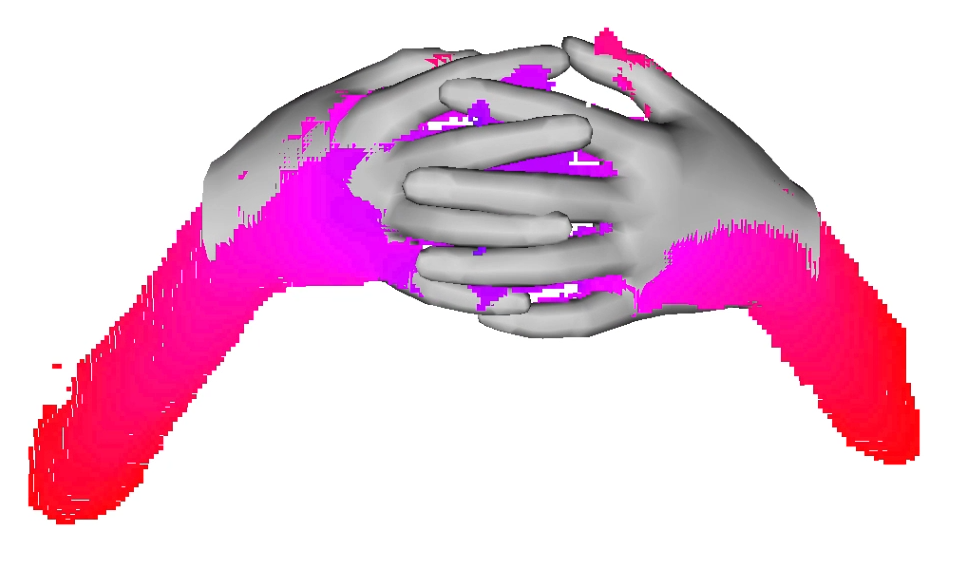}}
  \hfill
  \frame{\includegraphics[width=0.19\linewidth]{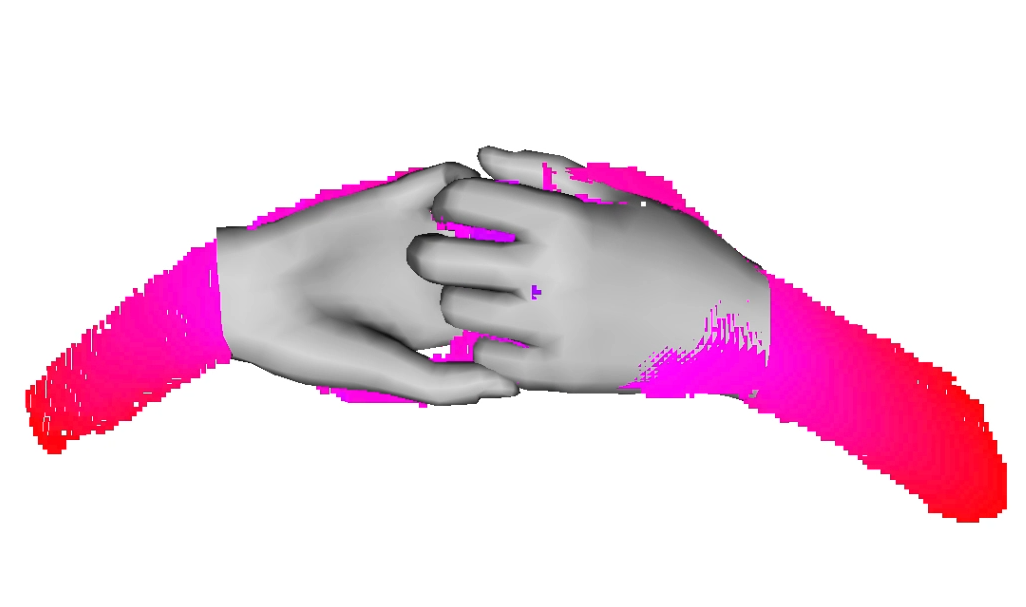}}
  \hfill
  \frame{\includegraphics[width=0.19\linewidth]{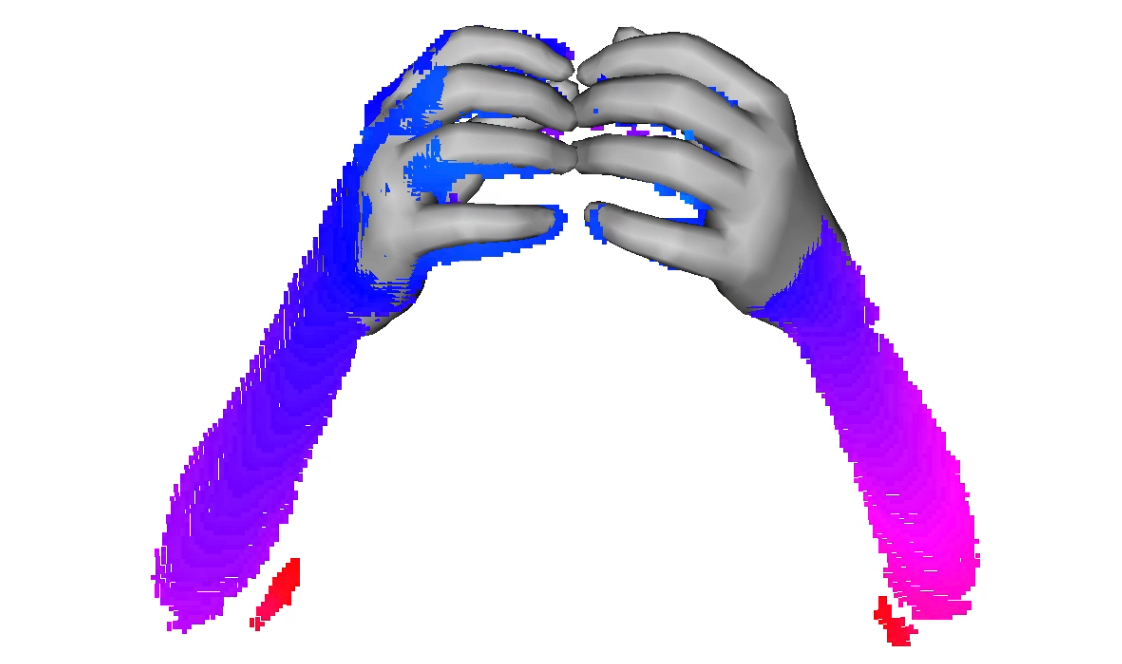}}
  \hfill
  \frame{\includegraphics[width=0.19\linewidth]{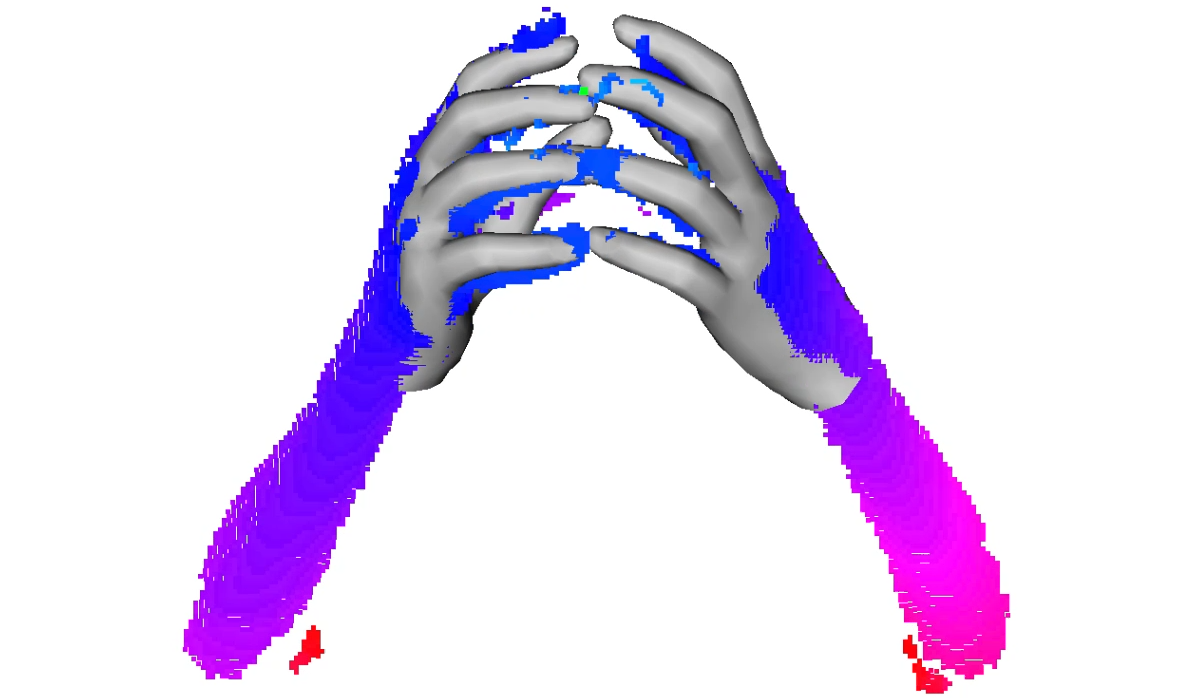}}
  \hfill
  \frame{\includegraphics[width=0.19\linewidth]{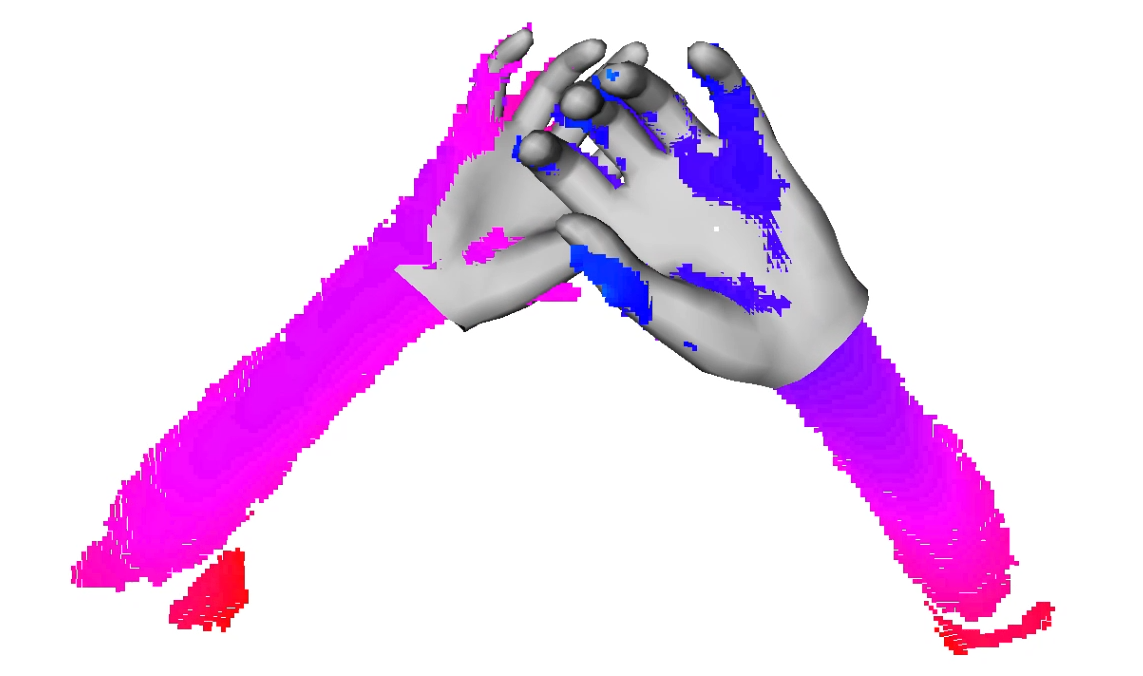}}
}
\vspace{0.1mm}
\centerline{
  \frame{\includegraphics[width=0.19\linewidth]{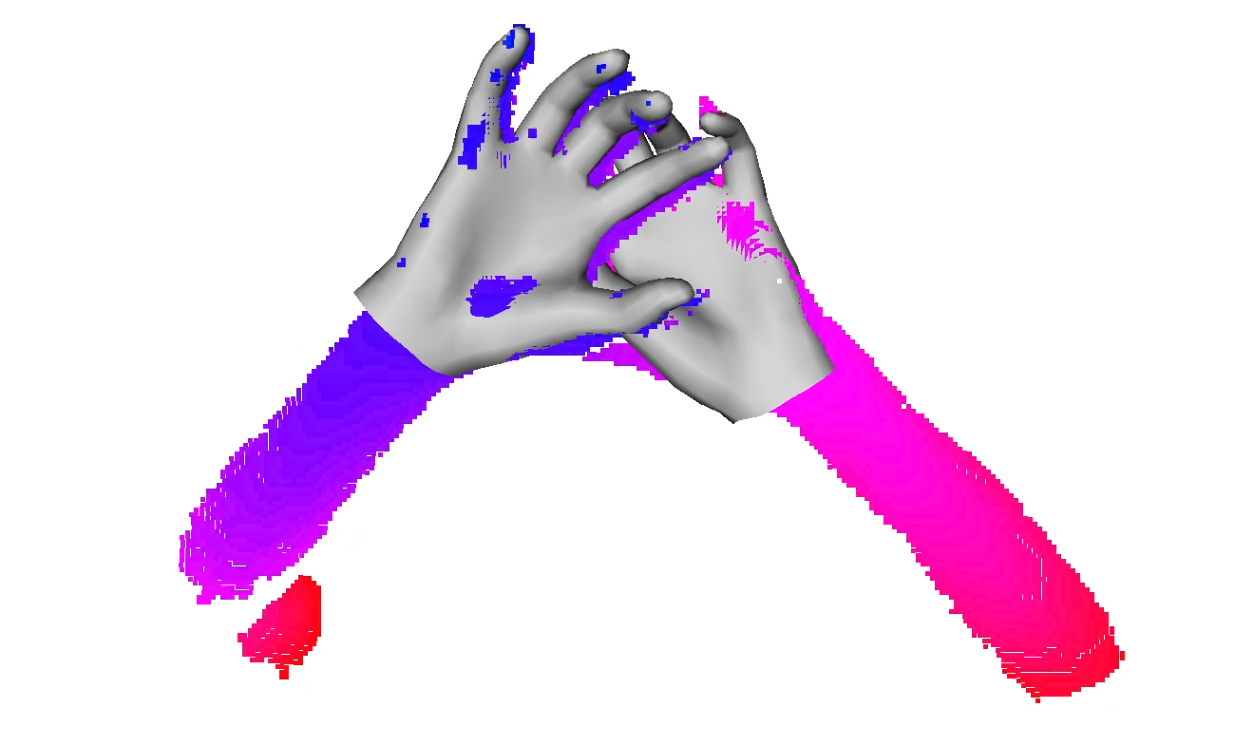}}
  \hfill
  \frame{\includegraphics[width=0.19\linewidth]{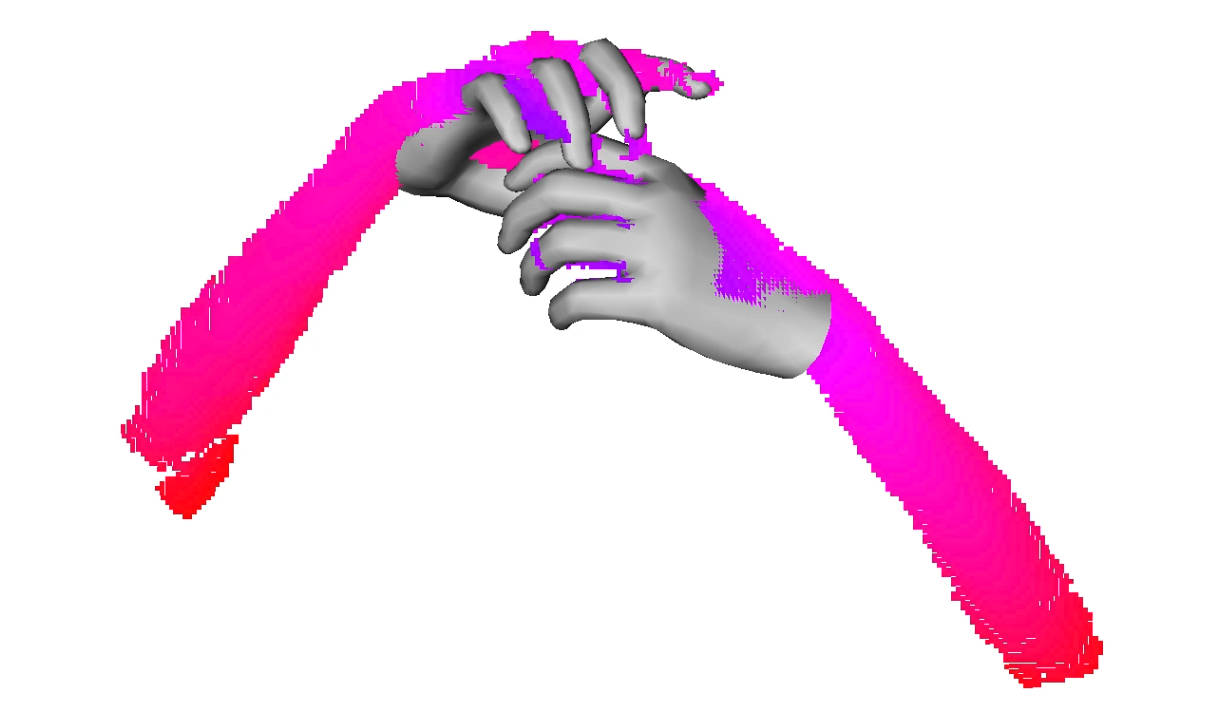}}
  \hfill
  \frame{\includegraphics[width=0.19\linewidth]{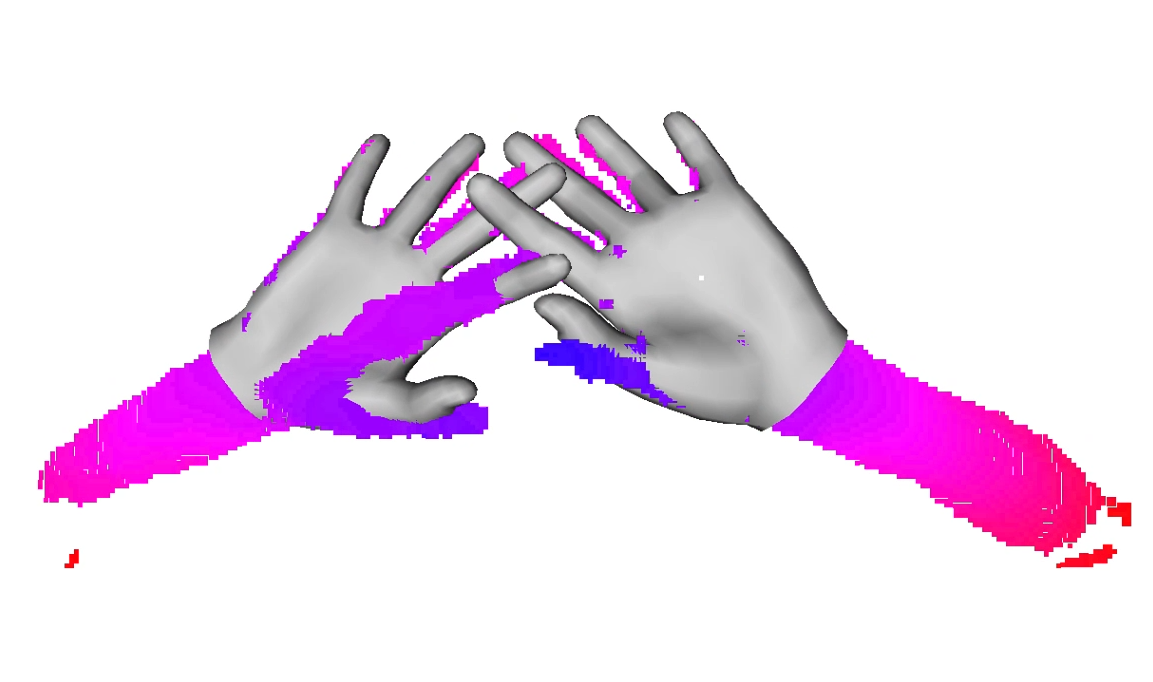}}
  \hfill
  \frame{\includegraphics[width=0.19\linewidth]{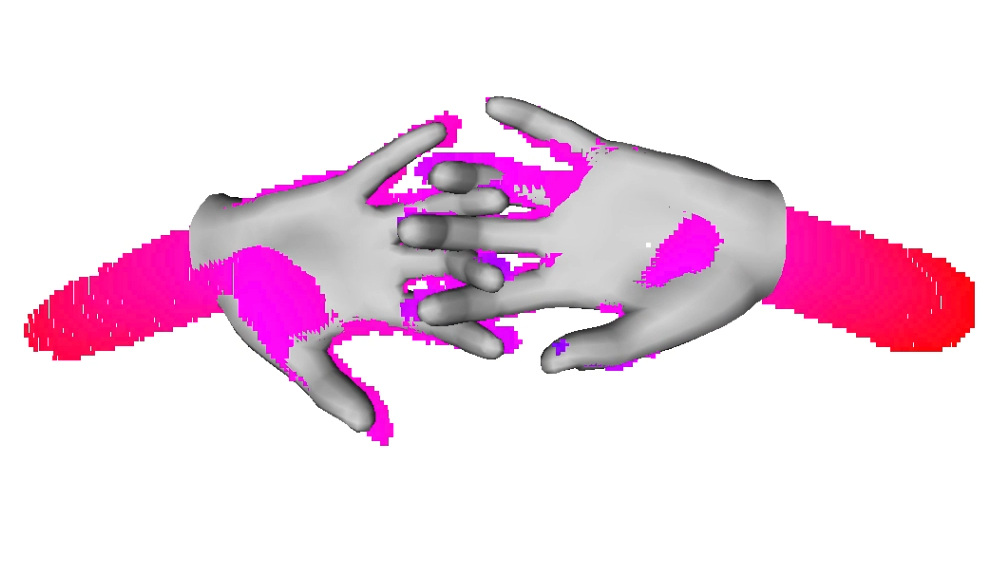}}
  \hfill
  \frame{\includegraphics[width=0.19\linewidth]{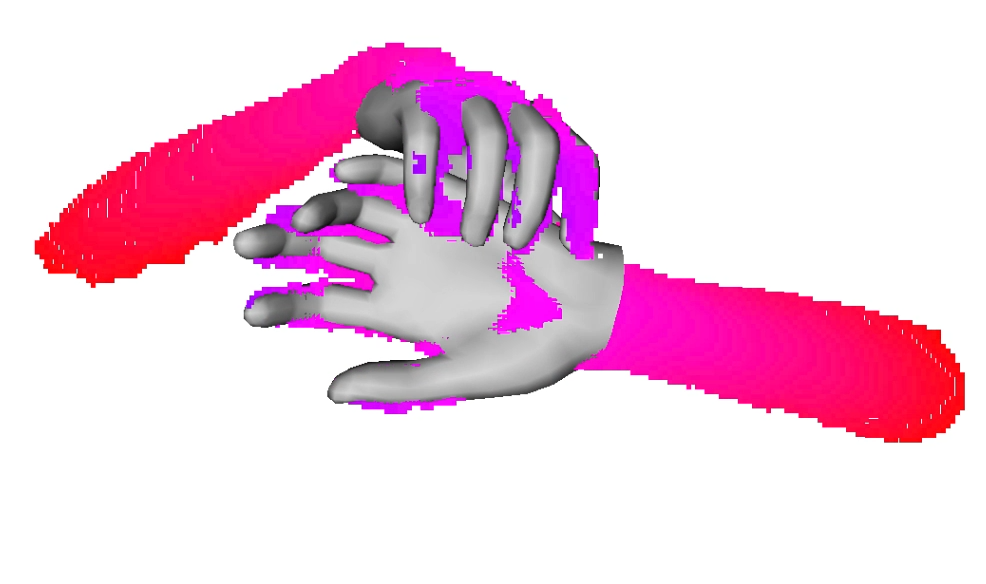}}
}
\vspace{0.1mm}
\centerline{
  \frame{\includegraphics[width=0.19\linewidth]{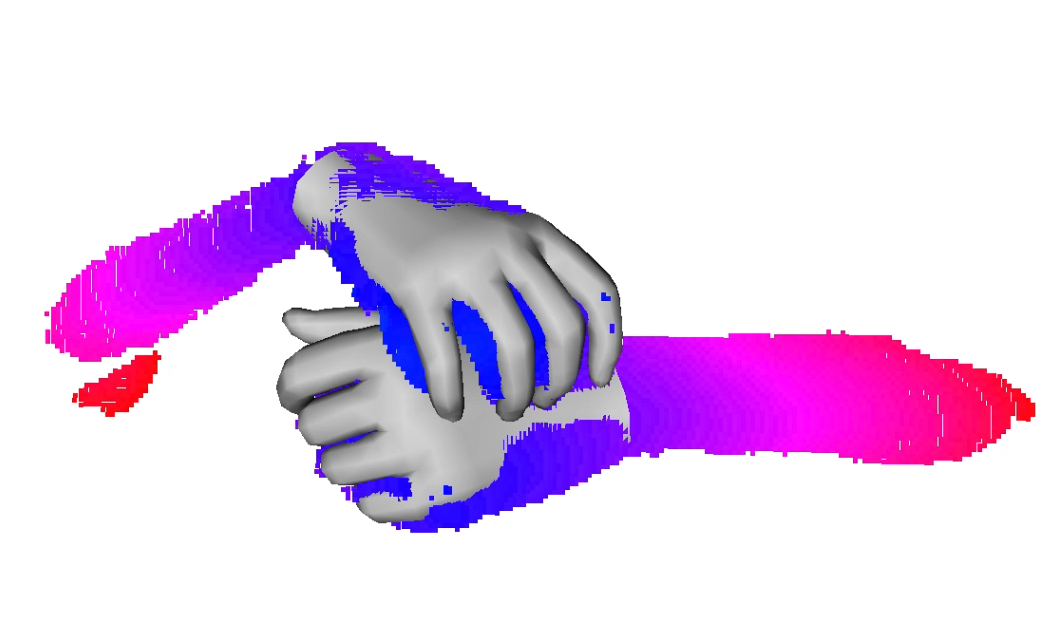}}
  \hfill
  \frame{\includegraphics[width=0.19\linewidth]{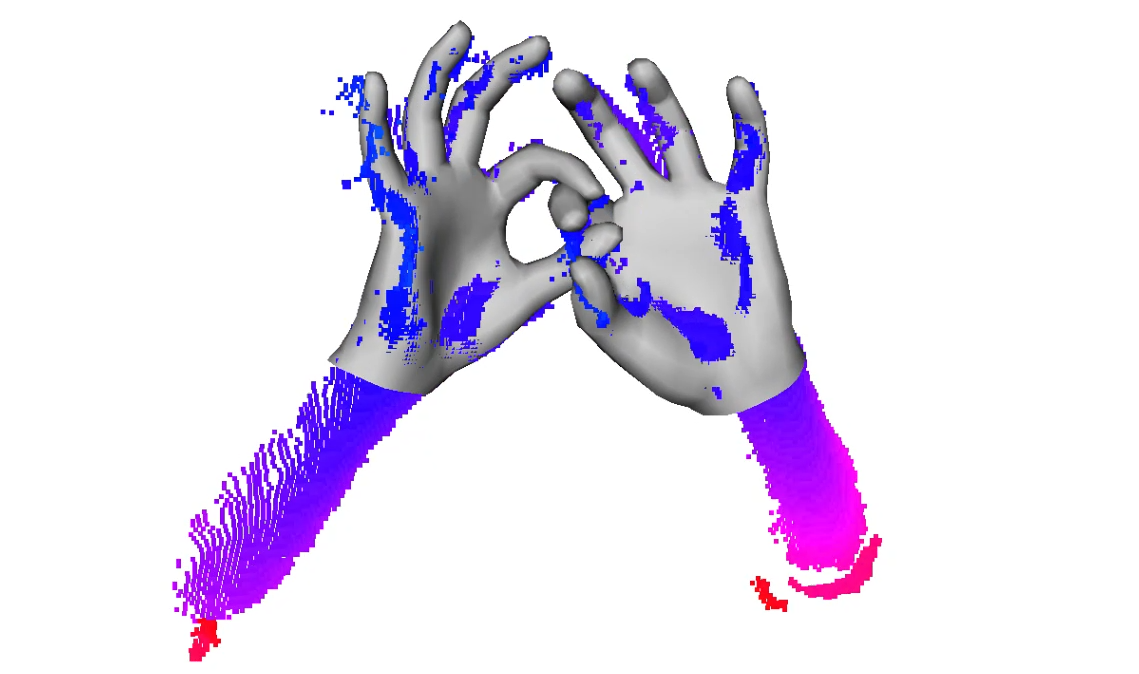}}
  \hfill
  \frame{\includegraphics[width=0.19\linewidth]{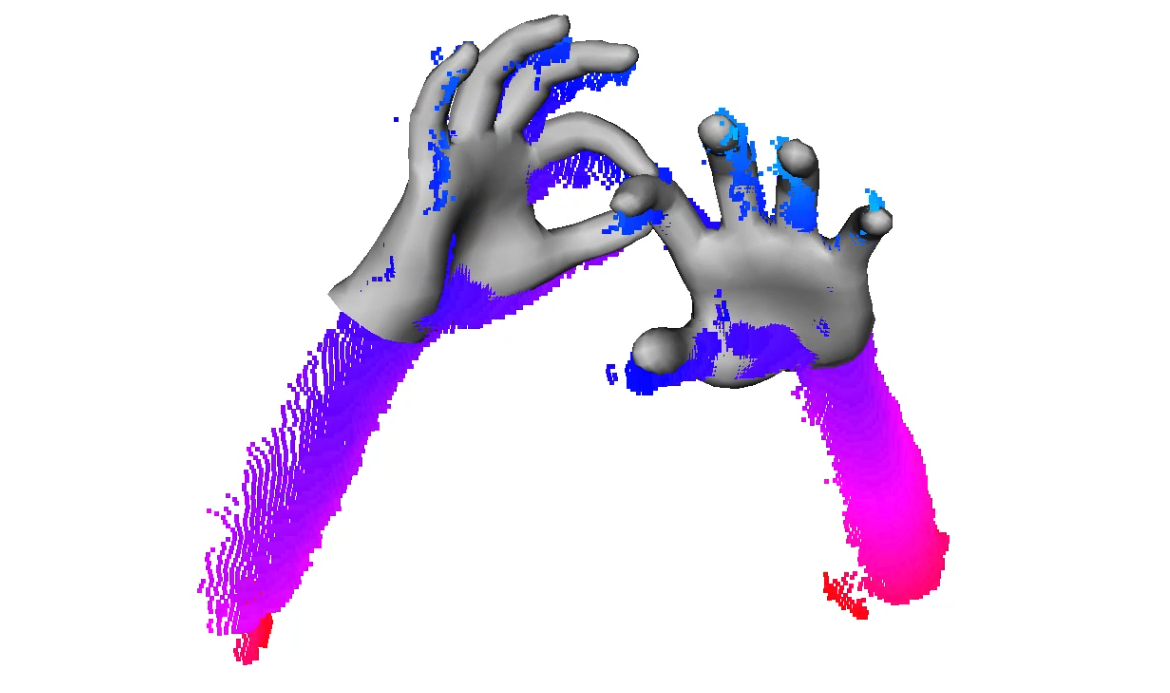}}
  \hfill
  \frame{\includegraphics[width=0.19\linewidth]{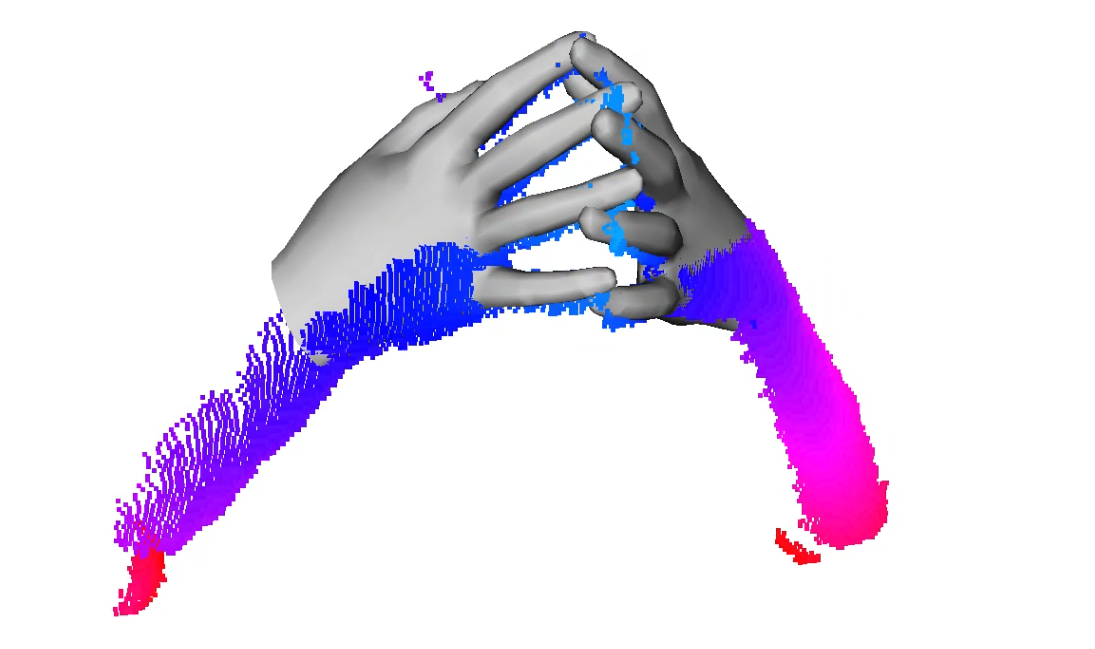}}
  \hfill
  \frame{\includegraphics[width=0.19\linewidth]{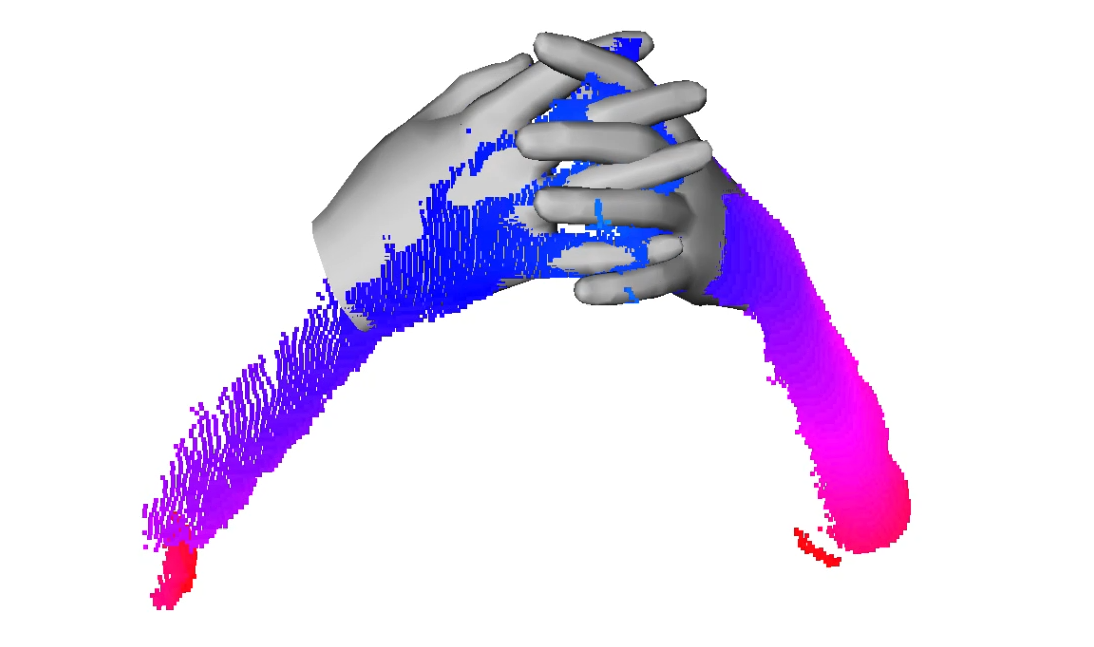}}
}
\vspace{0.1mm}
\centerline{
  \frame{\includegraphics[width=0.19\linewidth]{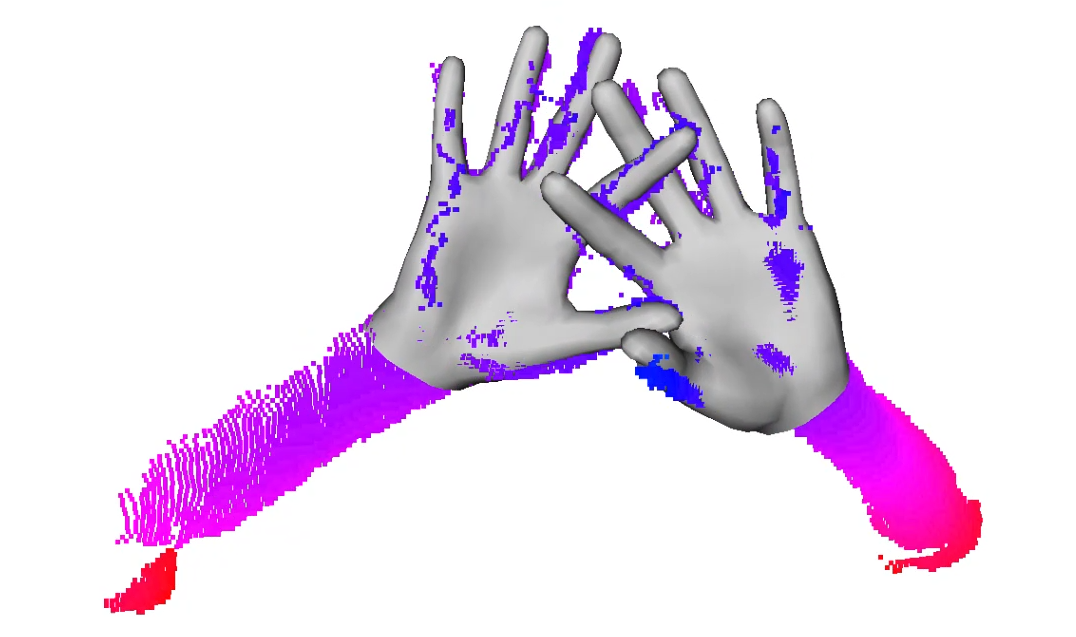}}
  \hfill
  \frame{\includegraphics[width=0.19\linewidth]{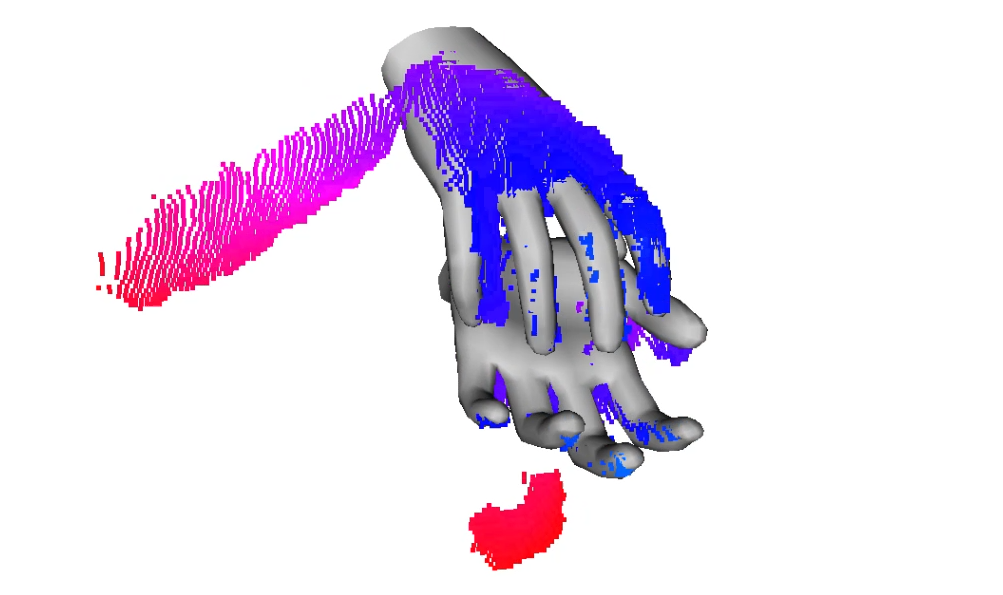}}
  \hfill
  \frame{\includegraphics[width=0.19\linewidth]{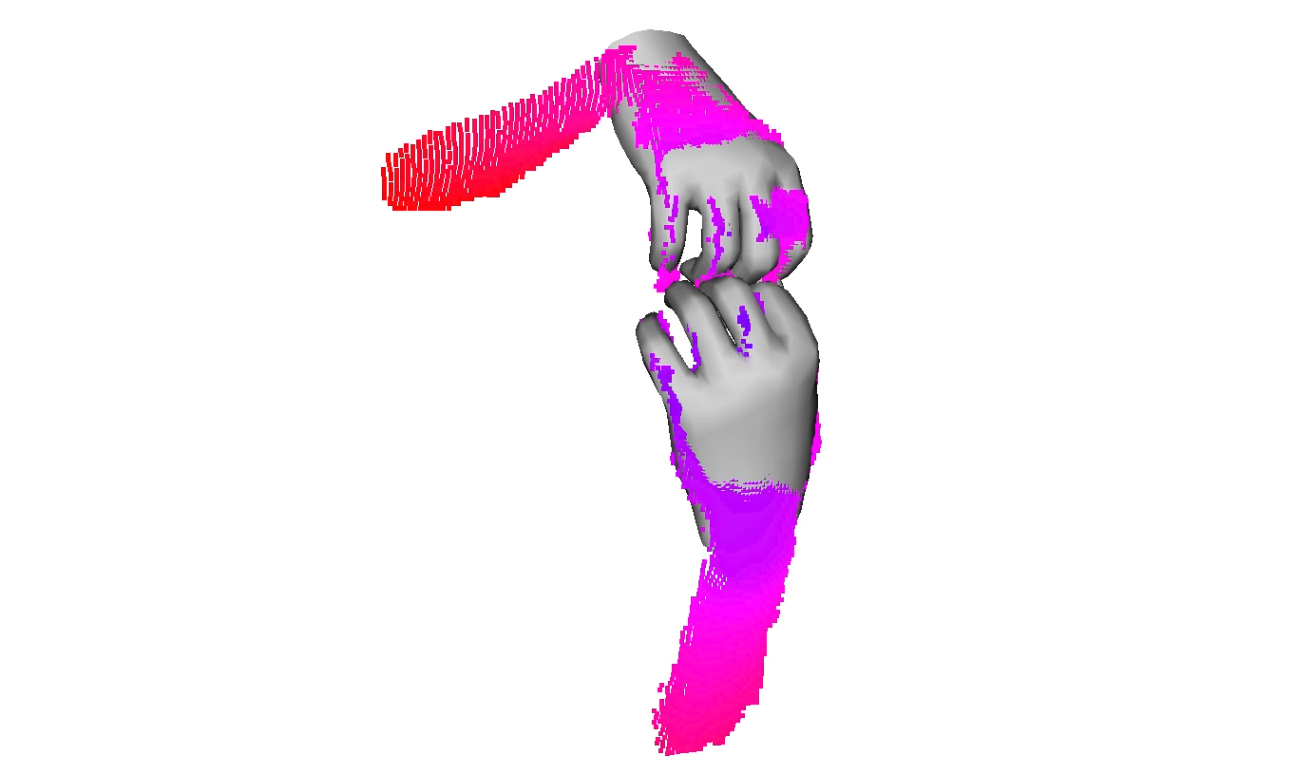}}
  \hfill
  \frame{\includegraphics[width=0.19\linewidth]{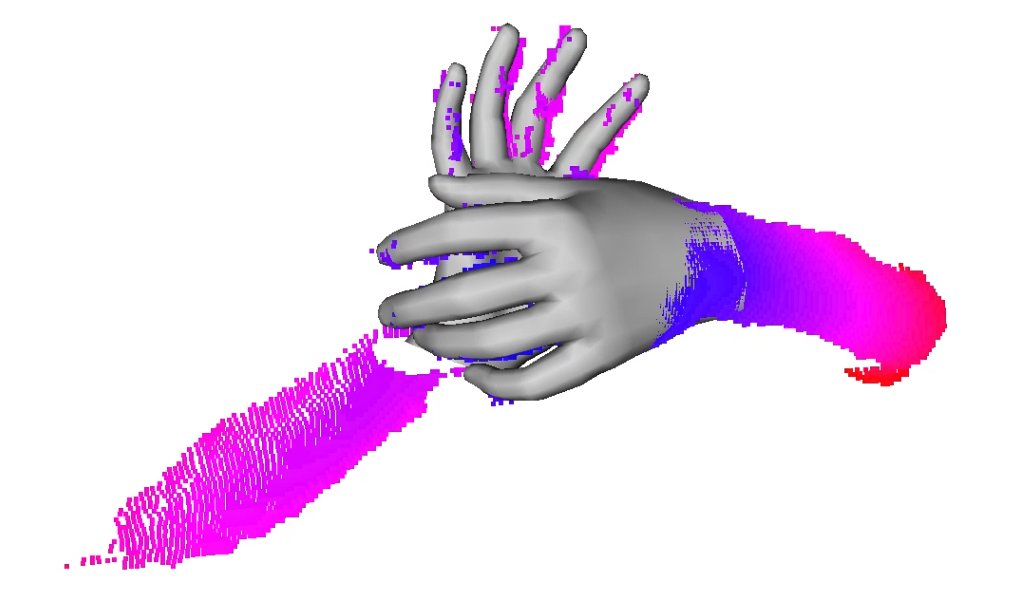}}
  \hfill
  \frame{\includegraphics[width=0.19\linewidth]{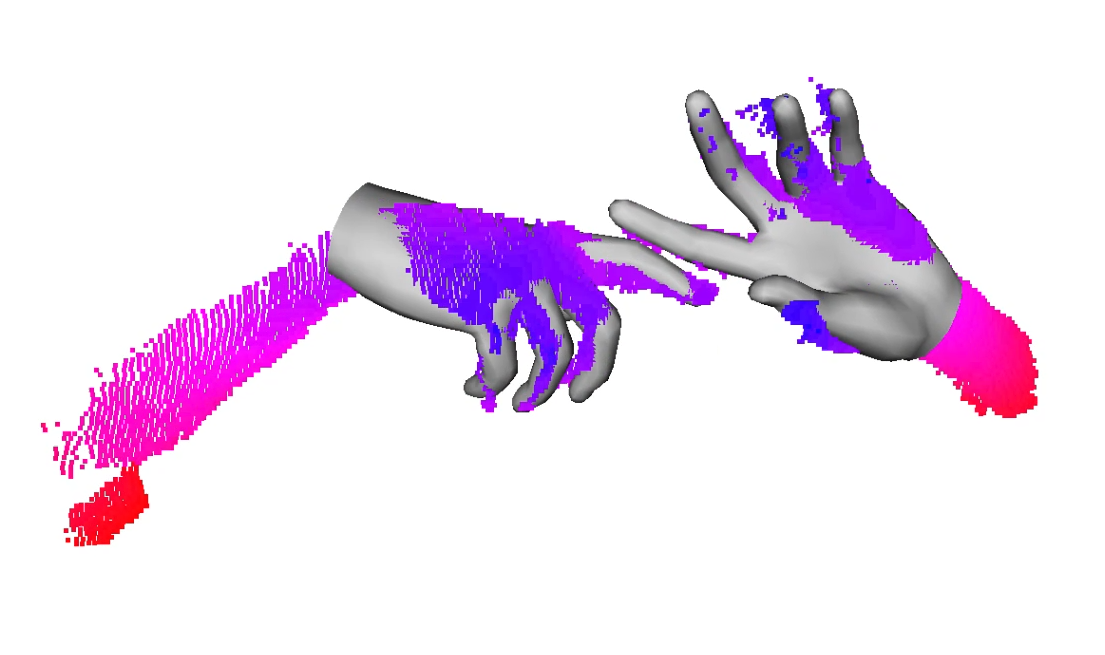}}
}
  \caption{We show qualitative results for the proposed method.  Note that the different colors of the depth image are due to different absolute depth values.}
  \label{fig:qualitative_results}
\end{figure*}

\begin{figure*}
    \centerline{
    \rotatebox{90}{\hspace{0.6cm}Depth}
    \hfill
  \frame{\includegraphics[width=0.19\linewidth, trim={1.5cm 3cm 1.5cm 2cm}, clip]{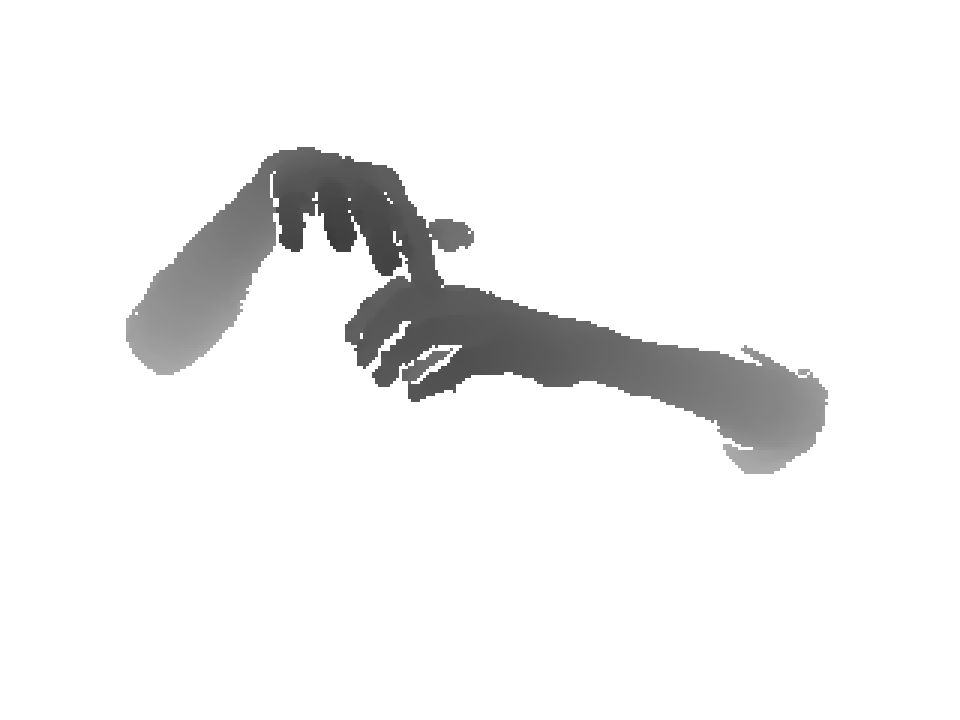}}
  \hfill
  \frame{\includegraphics[width=0.19\linewidth, trim={1.5cm 0cm 1.5cm 5cm}, clip]{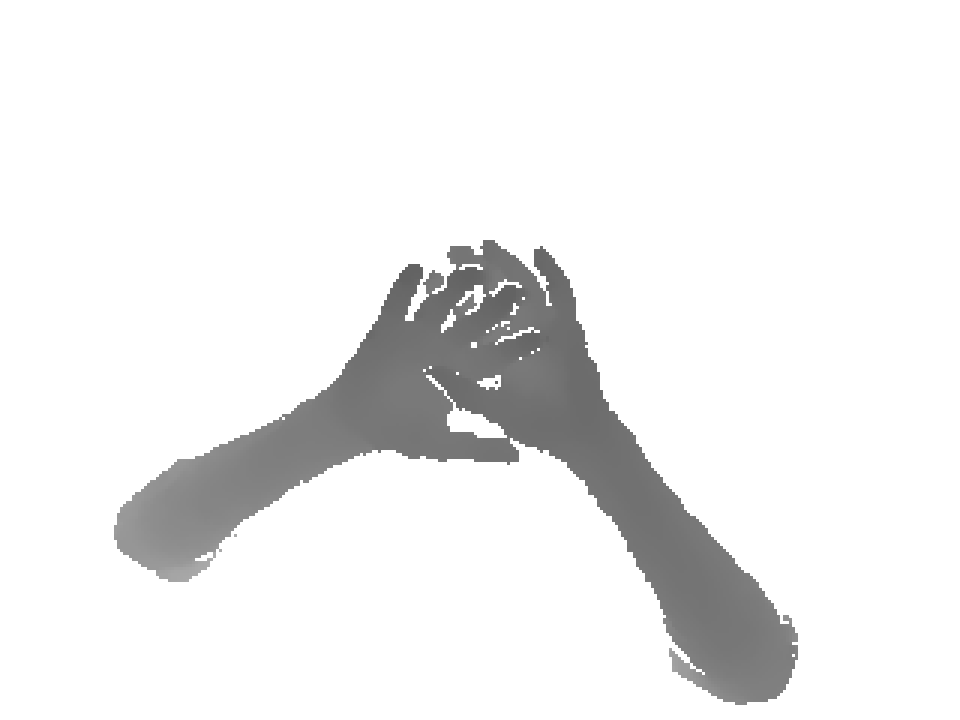}}
  \hfill
  \frame{\includegraphics[width=0.19\linewidth, trim={1.5cm 0.5cm 1.5cm 4.5cm}, clip]{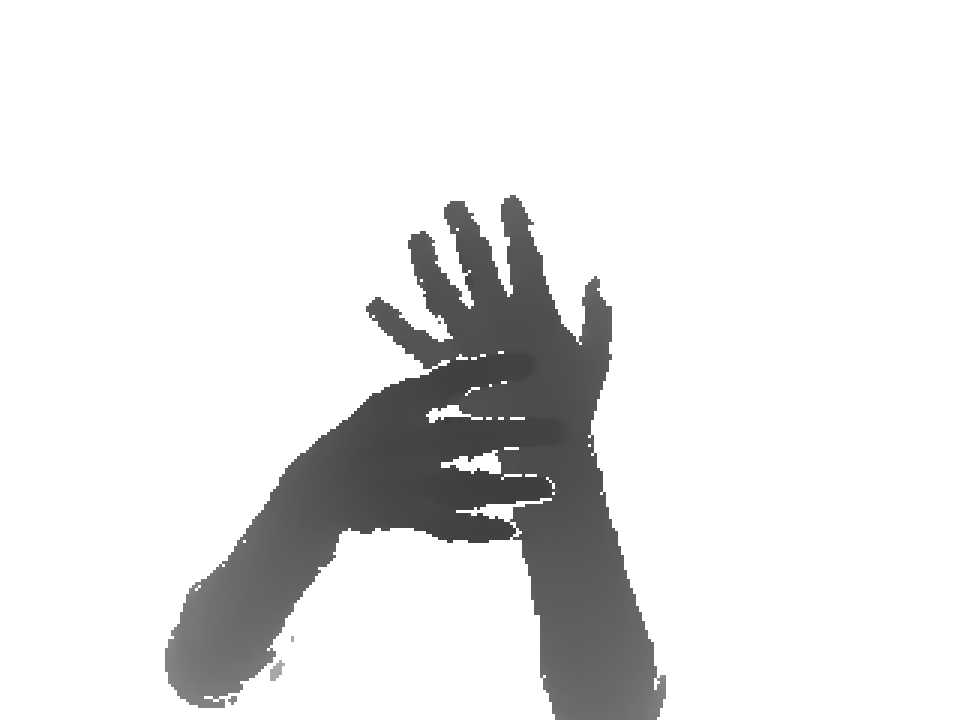}}
  \hfill
  \frame{\includegraphics[width=0.19\linewidth, trim={1.5cm 0cm 1.5cm 5cm}, clip]{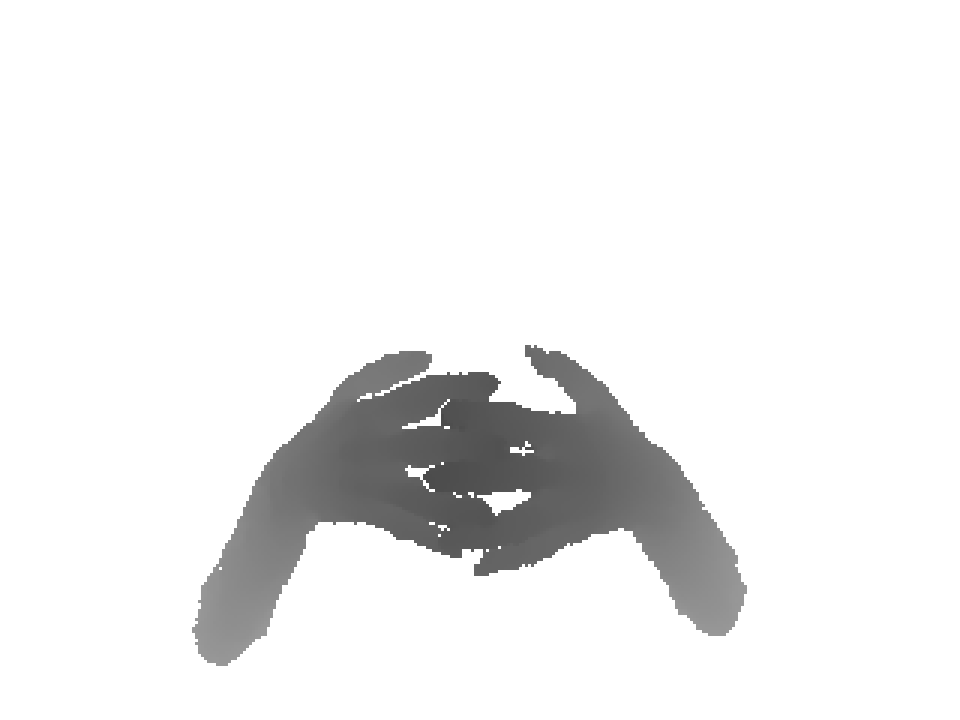}}
  \hfill
  \frame{\includegraphics[width=0.19\linewidth, trim={1.5cm 2.5cm 1.5cm 2.5cm}, clip]{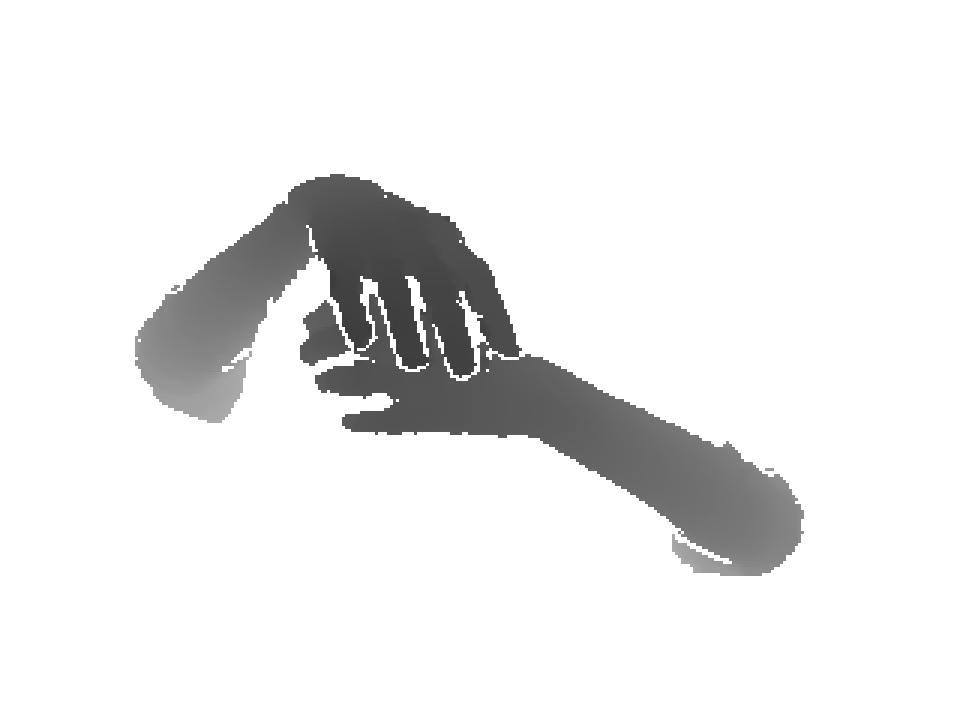}}
}
    \centerline{
    \rotatebox{90}{\hspace{0.15cm}Segmentation}
    \hfill
  \frame{\includegraphics[width=0.19\linewidth, trim={1.5cm 3cm 1.5cm 2cm}, clip]{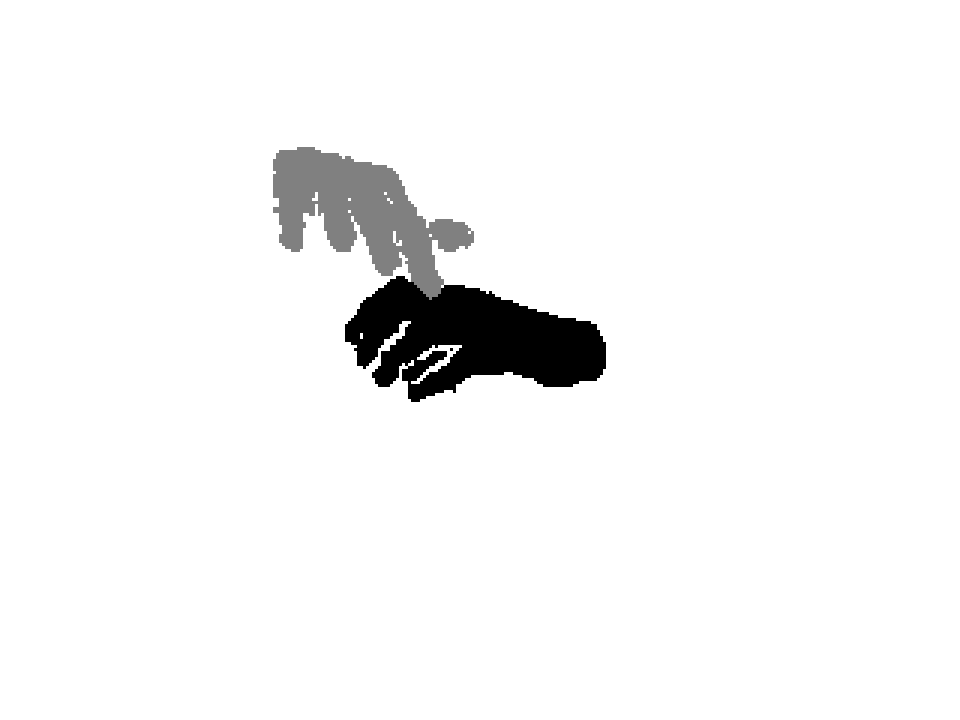}}
  \hfill
  \frame{\includegraphics[width=0.19\linewidth, trim={1.5cm 0cm 1.5cm 5cm}, clip]{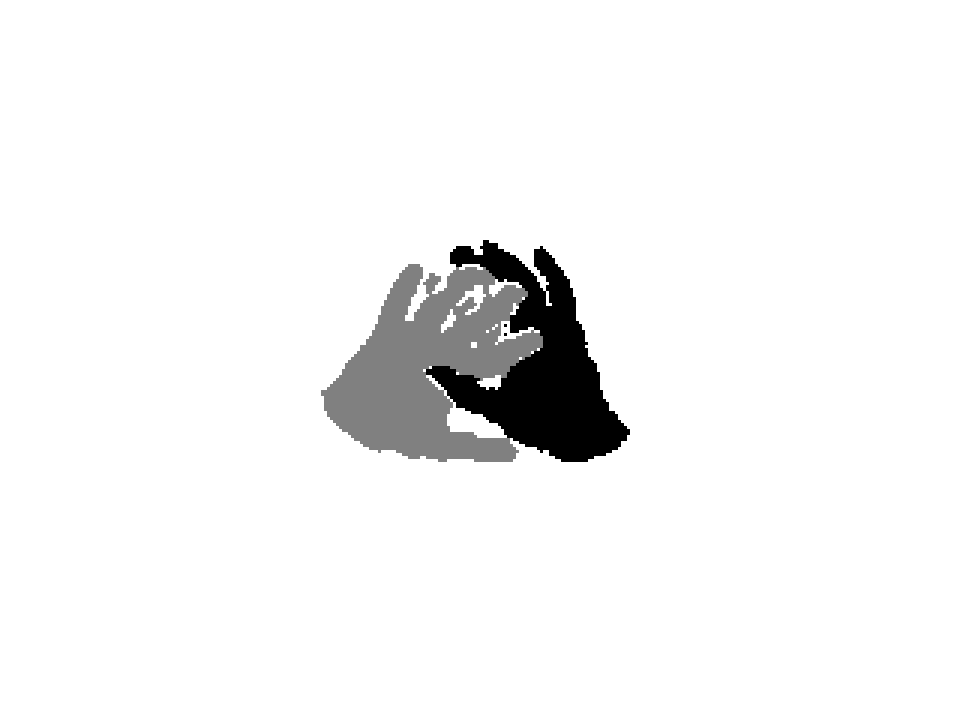}}
  \hfill
  \frame{\includegraphics[width=0.19\linewidth, trim={1.5cm 0.5cm 1.5cm 4.5cm}, clip]{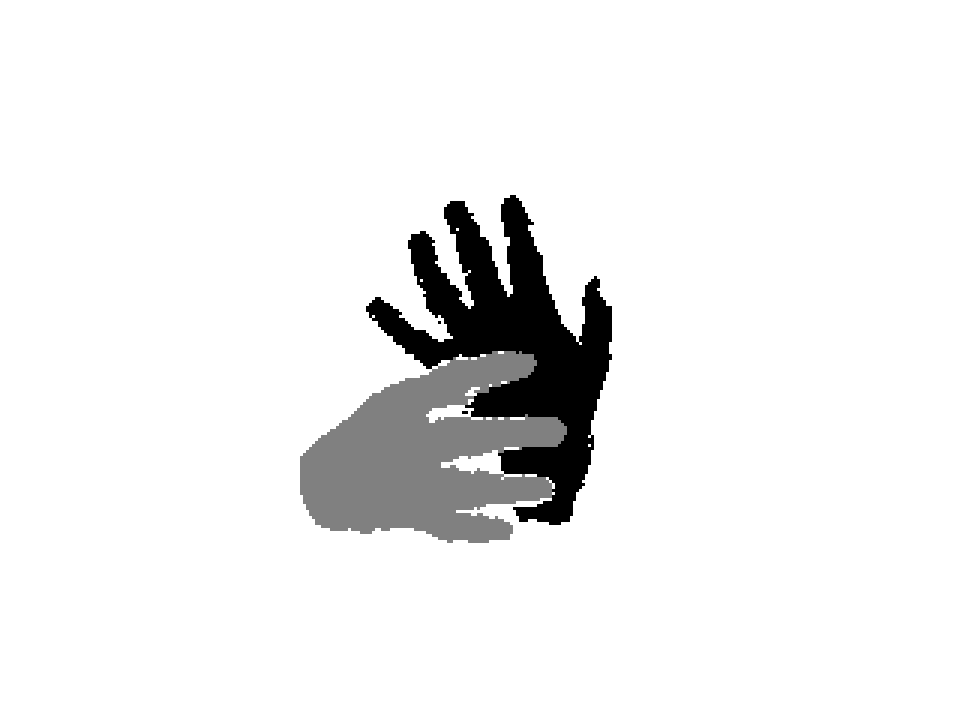}}
  \hfill
  \frame{\includegraphics[width=0.19\linewidth, trim={1.5cm 0cm 1.5cm 5cm}, clip]{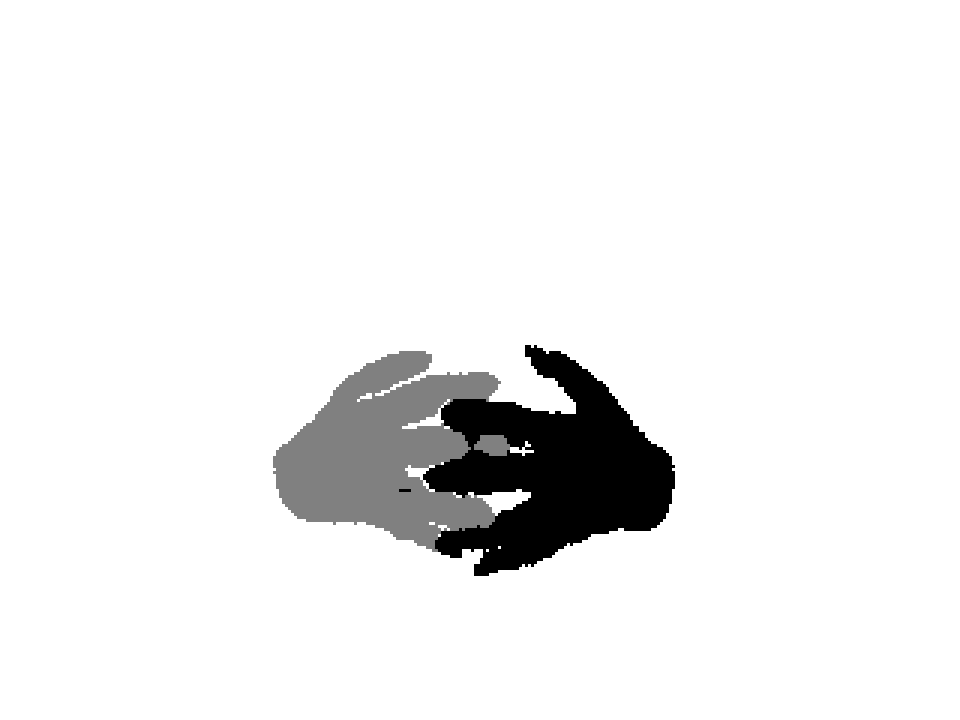}}
  \hfill
  \frame{\includegraphics[width=0.19\linewidth, trim={1.5cm 2.5cm 1.5cm 2.5cm}, clip]{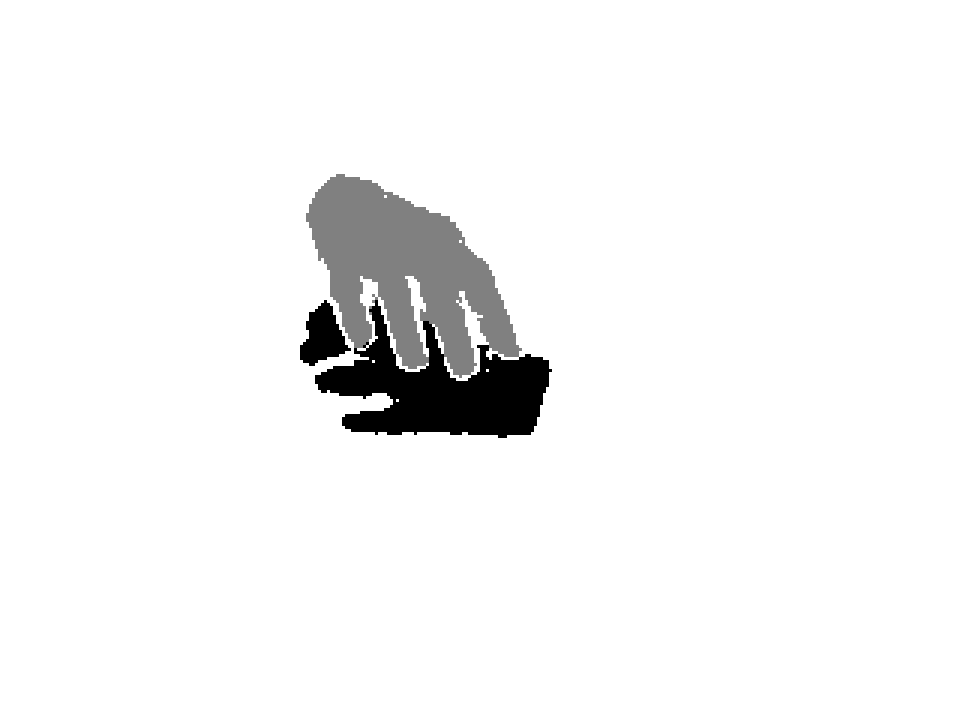}}
}
    \centerline{
    \rotatebox{90}{\hspace{0.55cm}Corresp.}
    \hfill
  \frame{\includegraphics[width=0.19\linewidth, trim={1.5cm 3cm 1.5cm 2cm}, clip]{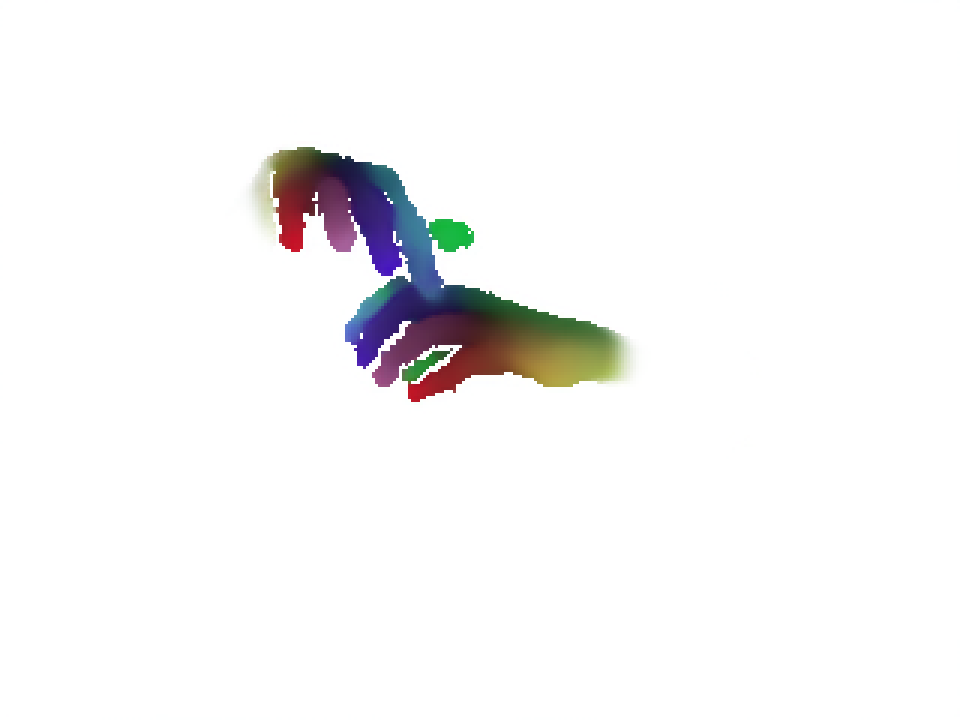}}
  \hfill
  \frame{\includegraphics[width=0.19\linewidth, trim={1.5cm 0cm 1.5cm 5cm}, clip]{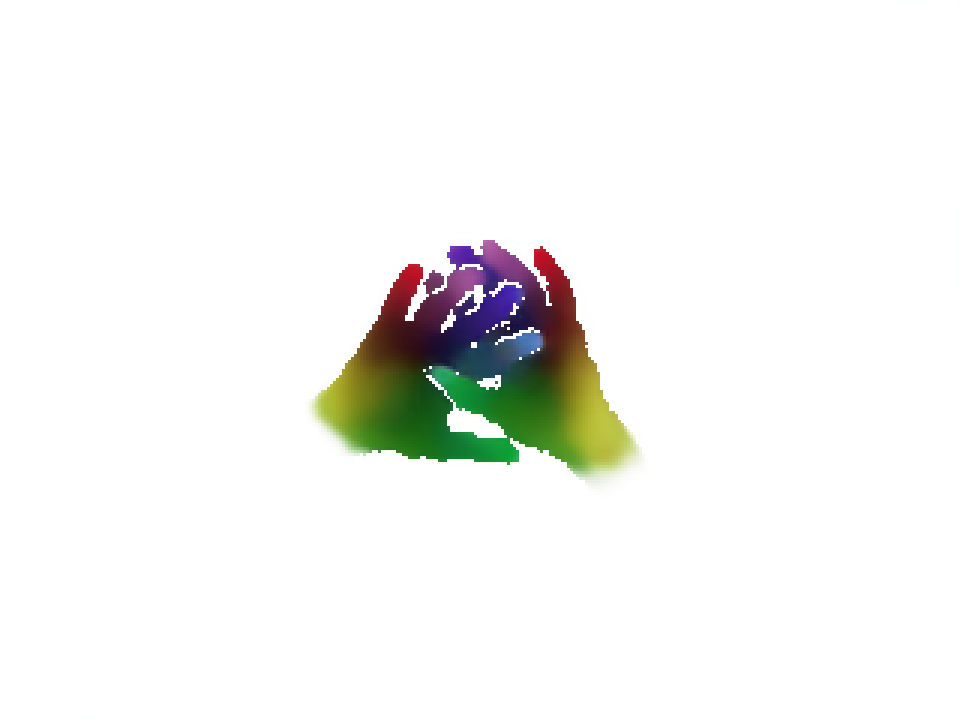}}
  \hfill
  \frame{\includegraphics[width=0.19\linewidth, trim={1.5cm 0.5cm 1.5cm 4.5cm}, clip]{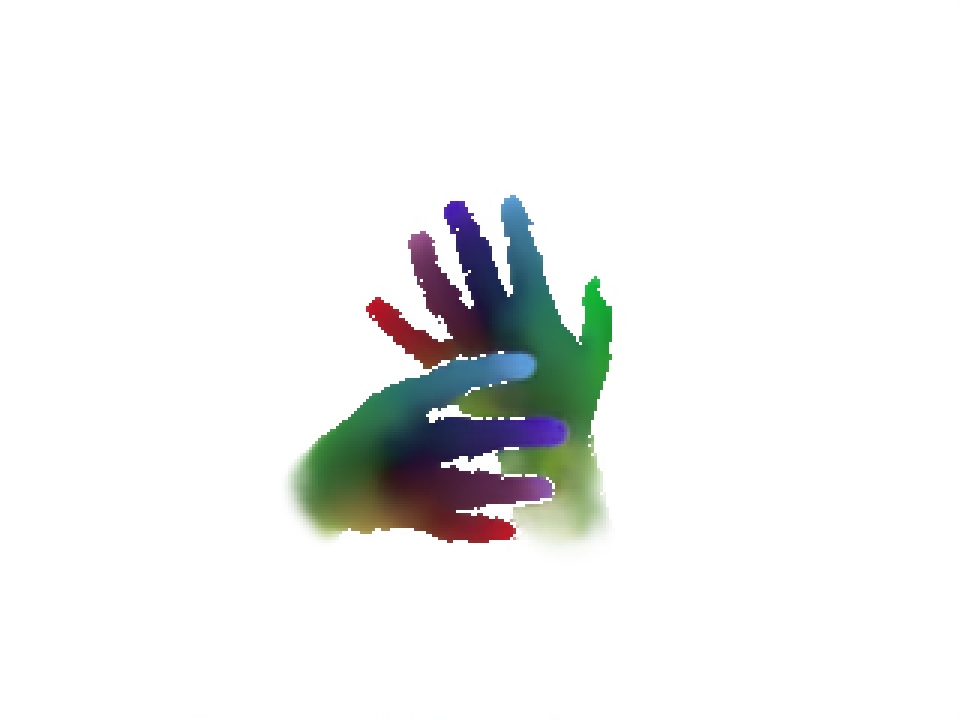}}
  \hfill
  \frame{\includegraphics[width=0.19\linewidth, trim={1.5cm 0cm 1.5cm 5cm}, clip]{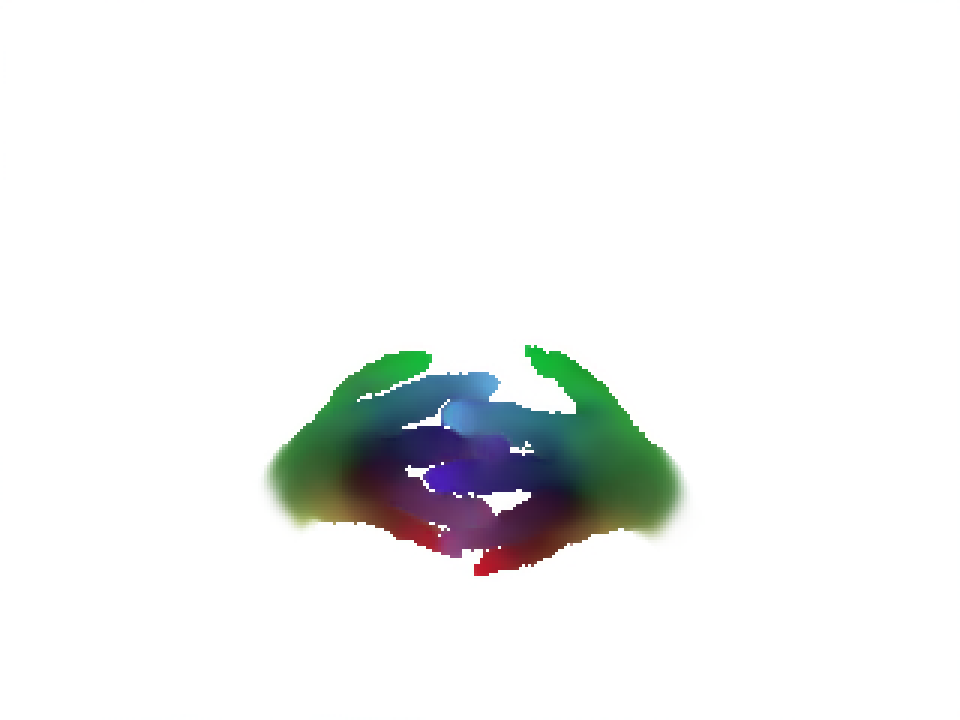}}
  \hfill
  \frame{\includegraphics[width=0.19\linewidth, trim={1.5cm 2.5cm 1.5cm 2.5cm}, clip]{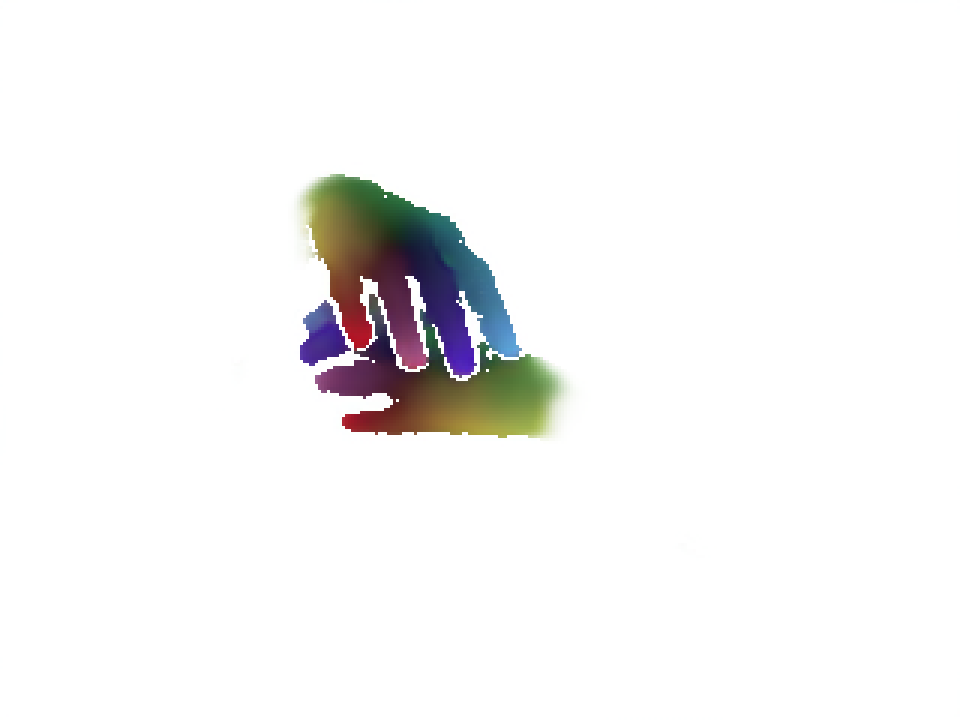}}
}
    \caption{
    Given a depth image (top) as input, our CoRN produces accurate segmentation (middle) and dense correspondences (bottom).
    }
    \label{fig:qualitative_corn}
\end{figure*}

\begin{figure}
    \includegraphics[page=1,width=\columnwidth,trim={0 7.7cm 11.5cm 3cm},clip]{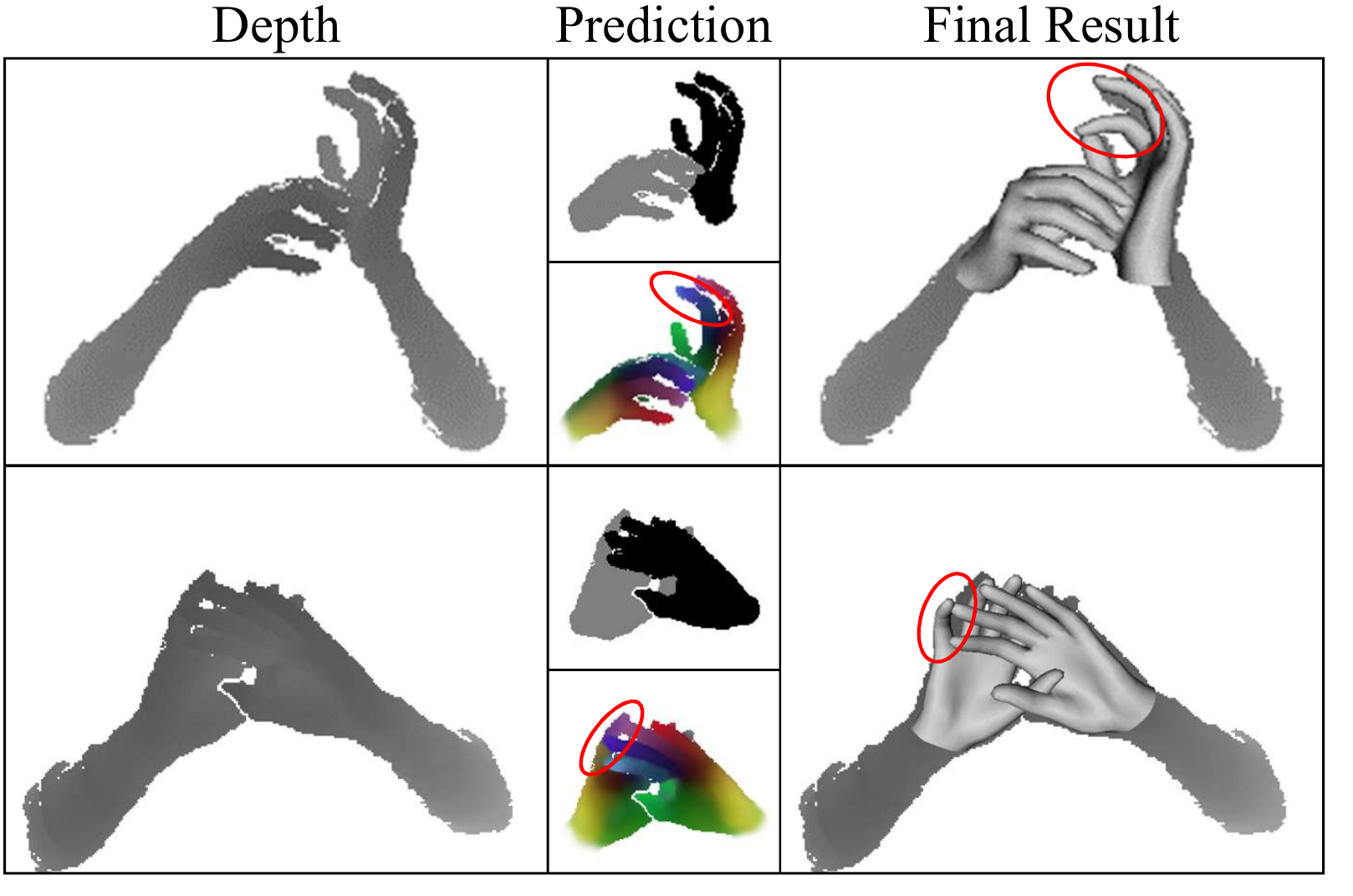}
    \caption{
    Erroneous CoRN predictions, e.g.\ wrongly classified fingers, negatively impact the final tracking result (see Fig.~\ref{fig:model} for the reference coloring).
    }
    \label{fig:corn_fails}
\end{figure}

\paragraph{Qualitative Results}
In Fig.~\ref{fig:qualitative_results} we present qualitative results of our pose and shape estimation method. In the first two rows we show frames for an egocentric view-point, where CoRN was also trained for this setting, whereas the remaining rows show frames for a frontal view-point. It can be seen that in a wide range of complex hand-hand interactions our method robustly estimates the hand pose and shape.
CoRN is an essential part of our method and is able to accurately predict segmentation and dense correspondences for a variety of inputs (see Fig.~\ref{fig:qualitative_corn}).
However, wrong predictions may lead to errors in the final tracking result as demonstrated in Fig.~\ref{fig:corn_fails}. 

%% file: sections/discussion.tex
\section{Limitations and Discussion}
Although in overall we have demonstrated compelling results for the estimation of hand pose and shape in real-time, there are several points that leave room for further improvements.
In terms of computational cost, currently our setup depends on two high-end GPUs, one for the regression network and one for the optimizer. In order to achieve a computationally more light-weight processing pipeline, one could consider lighter neural network architectures, such as CNNs tailored towards mobile platforms (e.g.~\cite{howard2017mobilenets}). 
While our approach better handles complex hand-hand interactions compared to previous real-time approaches, in very challenging situations our method may still struggle. 
For example, this may happen when the user performs extremely fast hand motions that lead to a severely blurred depth image, or when one of the hands is mostly occluded.
In the latter case, temporal jitter may occur due to the insufficient information in the depth image. This could be mitigated by a more involved temporal smoothness term, e.g. stronger smoothing when occlusions are happening, or a temporal network architecture for correspondence prediction.
Also, our current temporal smoothness prior may cause a delay in the tracking for large inter-frame motion.
To further improve the quality of the results, in the future one can use more elaborate strategies for finding correspondences, e.g.\:by using matching methods that are more advanced than nearest-neighbor search, or by incorporating confidence estimates in the correspondence predictor. 
Although our data generation scheme has proven successful for training CoRN, some generated images might not be completely realistic. This is due to the LeapMotion tracker's limitations and the hence mandatory distance between the two real hands. In future work, our proposed method could drive the simulation, and the data could be iteratively refined by bootstrapping.
While our approach is the only real-time approach that can adjust to user-specific hand shapes, the obtained hand shapes are not as detailed as high-quality laser scans. On the one hand, this is because the MANO model~\cite{Romero_siggraphasia2017} is rather coarse with its $778$ vertices per hand, and on the other hand the depth image is generally of lower resolution compared to laser scans. 
One relevant direction for future works is to deal with two hands that manipulate an object. Particular challenges are that one additionally needs to separate the object from the hands, as well as being able to cope with more severe occlusions due to the object. 
Another point that we leave for future work is to also integrate a physics simulation step directly into the tracker, so that at run-time one can immediately take fine-scale collisions into account.
Currently, slight intersections may still happen due to our computationally efficient but coarse collision proxies.

%% file: sections/conclusion.tex
\section{Conclusion}
We have presented a method for real-time pose and shape reconstruction of two interacting hands. The main features that distinguishes our work from previous two-hand tracking approaches is that it combines a wide range of favorable properties, namely it is marker-less, relies on a single depth camera, handles collisions, runs in real time with a commodity camera, and adjusts to user-specific hand shapes. This is achieved by combining a neural network for the prediction of correspondences with an energy minimization framework that optimizes for hand pose and shape parameters. For training the correspondence regression network, we have leveraged a physics-based simulation for generating (annotated) synthetic training data that contains physically plausible interactions between two hands. Due to a highly efficient GPU-based implementation of the energy minimization based on a Gauss-Newton optimizer, the approach is real-time capable. We have experimentally shown that our approach achieves results that are qualitatively similar and quantitatively close to the two-hand tracking solution by Tzionas~et~al.~\shortcite{tzionas_ijcv2016}, while at the same time being two orders of magnitude faster. 
Moreover, we demonstrated that qualitatively our method can handle more complex hand-hand interactions compared to recent state-of-the-art hand trackers. 

%% file: sections/appendix.tex
\appendix

\section{Neural Network Training Details}
All our networks were trained in Tensorflow using the Adam \cite{kingma2014adam} optimizer with the default parameter settings.
We trained for $450{,}000$ iterations using a batch size of 8.
Synthetic and real images were sampled with $50\%$ probability each.
With a training time of approximately $20$ seconds for 100 iterations, the total training process took $25$ hours on an Nvidia Tesla V100 GPU. 

We scale the depth values to meters and subtract the mean value of all valid depth pixels.
Furthermore, we apply the following image augmentations to our training data, where all augmentation parameters are sampled from a uniform random distribution:
\begin{itemize}
	\item rotation augmentation with rotation angle $\in [-90,90]$ degrees,
	\item translation augmentation in the image plane with offset $\in [-0.25, 0.25] \cdot \text{image size}$, as well as
	\item scale augmentation with possibly changing aspect ratio in the range of $[1.0, 2.0]$.
\end{itemize}
Note that all these augmentations are applied on-the-fly while training, i.e.\:the sampled augmentations for a training sample differ for each epoch, effectively increasing the training set size. In addition to these on-the-fly augmentations, we also mirror all images (and apply the respective procedure to the annotations), which however is performed offline.

\section{GPU Implementation Details}
For our Gauss Newton optimization steps, we compute the non-constant entries of the Jacobian matrix $J \in \R^{8871 \times 122}$ and the residuals $f \in \R^{8871}$ using CUDA kernels on the GPU.
We make sure that all threads in the same block compute derivatives for the same energy term.
Subsequently, we compute the matrix-matrix and matrix-vector products $J^\top J$ and $J^\top f$ using an efficient implementation in shared memory.
For solving the linear system $J^\top J \cdot \delta = J^\top f$, we copy $J^\top J \in \R^{122 \times 122}$ and $J^\top f \in \R^{122}$ to the CPU and employ the preconditioned conjugate gradient (PCG) solver of the Eigen library to obtain the parameter update $\delta$.

\section{Collision Energy}
The 3D Gaussian collision proxies are coupled with the hand model s.t.\:the mean $\vecmu$ depends on the pose and shape parameters $\beta, \theta$, whereas the standard deviation $\sigma$ only depends on the shape $\beta$.
As described by \cite{ellipsoidtracker_3dv2014}, an integral over a product of two isotropic Gaussians $G_p(x; \mu_p, \sigma_p)$ and $G_q(x; \mu_q,\sigma_q)$ of dimension $d$ is given as: 
\begin{equation}
\int_{\R^3} G_p(x) \cdot G_q(x) \, dx =
    \frac{(2\pi)^{\frac{3}{2}} (\sigma_p^2\sigma_q^2)^{\frac{3}{2}}}{(\sigma_p^2 + \sigma_q^2)^{\frac{3}{2}}} \text{exp} \Bigg( -\frac{||\vecmu_p - \vecmu_q||^2_2}{2(\sigma_p^2 + \sigma_q^2)} \Bigg)
\end{equation}
This term is differentiable with respect to $\vecmu$ and $\sigma$.
Furthermore, the derivatives $\frac{\partial \vecmu}{\partial \beta}, \frac{\partial \vecmu}{\partial \theta}, \frac{\partial \sigma}{\partial \beta}$ can be derived from the hand model.
Please note that we do not use the derivative $\frac{\partial E_{\coll}(\beta,\theta)}{\partial \beta}$ in the optimization since this encourages shrinking of the hand models when they are interacting.
Instead, the shape $\beta$ is optimized using all other energy terms and the Gaussian parameters are updated according to $\beta$ in every optimizer iteration. 

%% file: TwoHands.bbl

\begin{thebibliography}{61}


\ifx \showCODEN    \undefined \def \showCODEN     #1{\unskip}     \fi
\ifx \showDOI      \undefined \def \showDOI       #1{#1}\fi
\ifx \showISBNx    \undefined \def \showISBNx     #1{\unskip}     \fi
\ifx \showISBNxiii \undefined \def \showISBNxiii  #1{\unskip}     \fi
\ifx \showISSN     \undefined \def \showISSN      #1{\unskip}     \fi
\ifx \showLCCN     \undefined \def \showLCCN      #1{\unskip}     \fi
\ifx \shownote     \undefined \def \shownote      #1{#1}          \fi
\ifx \showarticletitle \undefined \def \showarticletitle #1{#1}   \fi
\ifx \showURL      \undefined \def \showURL       {\relax}        \fi
\providecommand\bibfield[2]{#2}
\providecommand\bibinfo[2]{#2}
\providecommand\natexlab[1]{#1}
\providecommand\showeprint[2][]{arXiv:#2}

\bibitem[\protect\citeauthoryear{Alp~G{\"u}ler, Neverova, and
  Kokkinos}{Alp~G{\"u}ler et~al\mbox{.}}{2018}]%
        {Guler_2018_CVPR}
\bibfield{author}{\bibinfo{person}{Riza Alp~G{\"u}ler},
  \bibinfo{person}{Natalia Neverova}, {and} \bibinfo{person}{Iasonas
  Kokkinos}.} \bibinfo{year}{2018}\natexlab{}.
\newblock \showarticletitle{DensePose: Dense Human Pose Estimation in the
  Wild}. In \bibinfo{booktitle}{\emph{The IEEE Conference on Computer Vision
  and Pattern Recognition (CVPR)}}.
\newblock


\bibitem[\protect\citeauthoryear{Alp~Guler, Trigeorgis, Antonakos, Snape,
  Zafeiriou, and Kokkinos}{Alp~Guler et~al\mbox{.}}{2017}]%
        {Guler_2017_CVPR}
\bibfield{author}{\bibinfo{person}{Riza Alp~Guler}, \bibinfo{person}{George
  Trigeorgis}, \bibinfo{person}{Epameinondas Antonakos},
  \bibinfo{person}{Patrick Snape}, \bibinfo{person}{Stefanos Zafeiriou}, {and}
  \bibinfo{person}{Iasonas Kokkinos}.} \bibinfo{year}{2017}\natexlab{}.
\newblock \showarticletitle{DenseReg: Fully Convolutional Dense Shape
  Regression In-The-Wild}. In \bibinfo{booktitle}{\emph{The IEEE Conference on
  Computer Vision and Pattern Recognition (CVPR)}}.
\newblock


\bibitem[\protect\citeauthoryear{Badrinarayanan, Kendall, and
  Cipolla}{Badrinarayanan et~al\mbox{.}}{2015}]%
        {badrinarayanan2015segnet}
\bibfield{author}{\bibinfo{person}{Vijay Badrinarayanan}, \bibinfo{person}{Alex
  Kendall}, {and} \bibinfo{person}{Roberto Cipolla}.}
  \bibinfo{year}{2015}\natexlab{}.
\newblock \showarticletitle{Segnet: A deep convolutional encoder-decoder
  architecture for image segmentation}.
\newblock \bibinfo{journal}{\emph{arXiv preprint arXiv:1511.00561}}
  (\bibinfo{year}{2015}).
\newblock


\bibitem[\protect\citeauthoryear{Baek, In~Kim, and Kim}{Baek
  et~al\mbox{.}}{2018}]%
        {Baek_2018_CVPR}
\bibfield{author}{\bibinfo{person}{Seungryul Baek}, \bibinfo{person}{Kwang
  In~Kim}, {and} \bibinfo{person}{Tae-Kyun Kim}.}
  \bibinfo{year}{2018}\natexlab{}.
\newblock \showarticletitle{Augmented Skeleton Space Transfer for Depth-Based
  Hand Pose Estimation}. In \bibinfo{booktitle}{\emph{The IEEE Conference on
  Computer Vision and Pattern Recognition (CVPR)}}.
\newblock


\bibitem[\protect\citeauthoryear{Ballan, Taneja, Gall, Gool, and
  Pollefeys}{Ballan et~al\mbox{.}}{2012}]%
        {ballan_eccv2012}
\bibfield{author}{\bibinfo{person}{Luca Ballan}, \bibinfo{person}{Aparna
  Taneja}, \bibinfo{person}{Juergen Gall}, \bibinfo{person}{Luc~Van Gool},
  {and} \bibinfo{person}{Marc Pollefeys}.} \bibinfo{year}{2012}\natexlab{}.
\newblock \showarticletitle{{Motion Capture of Hands in Action using
  Discriminative Salient Points}}. In \bibinfo{booktitle}{\emph{European
  Conference on Computer Vision (ECCV)}}.
\newblock


\bibitem[\protect\citeauthoryear{Bronstein, Bronstein, Kimmel, and
  Yavneh}{Bronstein et~al\mbox{.}}{2006}]%
        {bronstein2006multigrid}
\bibfield{author}{\bibinfo{person}{Michael~M Bronstein},
  \bibinfo{person}{Alexander~M Bronstein}, \bibinfo{person}{Ron Kimmel}, {and}
  \bibinfo{person}{Irad Yavneh}.} \bibinfo{year}{2006}\natexlab{}.
\newblock \showarticletitle{Multigrid multidimensional scaling}.
\newblock \bibinfo{journal}{\emph{Numerical linear algebra with applications}}
  \bibinfo{volume}{13}, \bibinfo{number}{2-3} (\bibinfo{year}{2006}),
  \bibinfo{pages}{149--171}.
\newblock


\bibitem[\protect\citeauthoryear{Cai, Ge, Cai, and Yuan}{Cai
  et~al\mbox{.}}{2018}]%
        {cai2018weakly}
\bibfield{author}{\bibinfo{person}{Yujun Cai}, \bibinfo{person}{Liuhao Ge},
  \bibinfo{person}{Jianfei Cai}, {and} \bibinfo{person}{Junsong Yuan}.}
  \bibinfo{year}{2018}\natexlab{}.
\newblock \showarticletitle{Weakly-supervised 3d hand pose estimation from
  monocular rgb images}. In \bibinfo{booktitle}{\emph{European Conference on
  Computer Vision}}. Springer, Cham, \bibinfo{pages}{1--17}.
\newblock


\bibitem[\protect\citeauthoryear{Choi, Sinha, Hee~Choi, Jang, and Ramani}{Choi
  et~al\mbox{.}}{2015}]%
        {choi2015collaborative}
\bibfield{author}{\bibinfo{person}{Chiho Choi}, \bibinfo{person}{Ayan Sinha},
  \bibinfo{person}{Joon Hee~Choi}, \bibinfo{person}{Sujin Jang}, {and}
  \bibinfo{person}{Karthik Ramani}.} \bibinfo{year}{2015}\natexlab{}.
\newblock \showarticletitle{A collaborative filtering approach to real-time
  hand pose estimation}. In \bibinfo{booktitle}{\emph{Proceedings of the IEEE
  international conference on computer vision}}. \bibinfo{pages}{2336--2344}.
\newblock


\bibitem[\protect\citeauthoryear{Ge, Cai, Weng, and Yuan}{Ge
  et~al\mbox{.}}{2018}]%
        {Ge_2018_CVPR}
\bibfield{author}{\bibinfo{person}{Liuhao Ge}, \bibinfo{person}{Yujun Cai},
  \bibinfo{person}{Junwu Weng}, {and} \bibinfo{person}{Junsong Yuan}.}
  \bibinfo{year}{2018}\natexlab{}.
\newblock \showarticletitle{Hand PointNet: 3D Hand Pose Estimation Using Point
  Sets}. In \bibinfo{booktitle}{\emph{The IEEE Conference on Computer Vision
  and Pattern Recognition (CVPR)}}.
\newblock


\bibitem[\protect\citeauthoryear{Han, Liu, Wang, Ye, Twigg, and Kin}{Han
  et~al\mbox{.}}{2018}]%
        {han2018online}
\bibfield{author}{\bibinfo{person}{Shangchen Han}, \bibinfo{person}{Beibei
  Liu}, \bibinfo{person}{Robert Wang}, \bibinfo{person}{Yuting Ye},
  \bibinfo{person}{Christopher~D Twigg}, {and} \bibinfo{person}{Kenrick Kin}.}
  \bibinfo{year}{2018}\natexlab{}.
\newblock \showarticletitle{Online optical marker-based hand tracking with deep
  labels}.
\newblock \bibinfo{journal}{\emph{ACM Transactions on Graphics (TOG)}}
  \bibinfo{volume}{37}, \bibinfo{number}{4} (\bibinfo{year}{2018}),
  \bibinfo{pages}{166}.
\newblock


\bibitem[\protect\citeauthoryear{H{\"o}ll, Oberweger, Arth, and
  Lepetit}{H{\"o}ll et~al\mbox{.}}{2018}]%
        {holl2018efficient}
\bibfield{author}{\bibinfo{person}{Markus H{\"o}ll}, \bibinfo{person}{Markus
  Oberweger}, \bibinfo{person}{Clemens Arth}, {and} \bibinfo{person}{Vincent
  Lepetit}.} \bibinfo{year}{2018}\natexlab{}.
\newblock \showarticletitle{Efficient Physics-Based Implementation for
  Realistic Hand-Object Interaction in Virtual Reality}. In
  \bibinfo{booktitle}{\emph{2018 IEEE Conference on Virtual Reality and 3D User
  Interfaces}}.
\newblock


\bibitem[\protect\citeauthoryear{Howard, Zhu, Chen, Kalenichenko, Wang, Weyand,
  Andreetto, and Adam}{Howard et~al\mbox{.}}{2017}]%
        {howard2017mobilenets}
\bibfield{author}{\bibinfo{person}{Andrew~G Howard}, \bibinfo{person}{Menglong
  Zhu}, \bibinfo{person}{Bo Chen}, \bibinfo{person}{Dmitry Kalenichenko},
  \bibinfo{person}{Weijun Wang}, \bibinfo{person}{Tobias Weyand},
  \bibinfo{person}{Marco Andreetto}, {and} \bibinfo{person}{Hartwig Adam}.}
  \bibinfo{year}{2017}\natexlab{}.
\newblock \showarticletitle{Mobilenets: Efficient convolutional neural networks
  for mobile vision applications}.
\newblock \bibinfo{journal}{\emph{arXiv:1704.04861}} (\bibinfo{year}{2017}).
\newblock


\bibitem[\protect\citeauthoryear{Huang, Allain, Franco, Navab, Ilic, and
  Boyer}{Huang et~al\mbox{.}}{2016}]%
        {huang2016volumetric}
\bibfield{author}{\bibinfo{person}{Chun-Hao Huang}, \bibinfo{person}{Benjamin
  Allain}, \bibinfo{person}{Jean-S{\'e}bastien Franco}, \bibinfo{person}{Nassir
  Navab}, \bibinfo{person}{Slobodan Ilic}, {and} \bibinfo{person}{Edmond
  Boyer}.} \bibinfo{year}{2016}\natexlab{}.
\newblock \showarticletitle{Volumetric 3d tracking by detection}. In
  \bibinfo{booktitle}{\emph{Proceedings of the IEEE Conference on Computer
  Vision and Pattern Recognition}}. \bibinfo{pages}{3862--3870}.
\newblock


\bibitem[\protect\citeauthoryear{Khamis, Taylor, Shotton, Keskin, Izadi, and
  Fitzgibbon}{Khamis et~al\mbox{.}}{2015}]%
        {khamis2015learning}
\bibfield{author}{\bibinfo{person}{Sameh Khamis}, \bibinfo{person}{Jonathan
  Taylor}, \bibinfo{person}{Jamie Shotton}, \bibinfo{person}{Cem Keskin},
  \bibinfo{person}{Shahram Izadi}, {and} \bibinfo{person}{Andrew Fitzgibbon}.}
  \bibinfo{year}{2015}\natexlab{}.
\newblock \showarticletitle{Learning an efficient model of hand shape variation
  from depth images}. In \bibinfo{booktitle}{\emph{Proceedings of the IEEE
  conference on computer vision and pattern recognition}}.
  \bibinfo{pages}{2540--2548}.
\newblock


\bibitem[\protect\citeauthoryear{Kim, Hilliges, Izadi, Butler, Chen,
  Oikonomidis, and Olivier}{Kim et~al\mbox{.}}{2012}]%
        {kim2012digits}
\bibfield{author}{\bibinfo{person}{David Kim}, \bibinfo{person}{Otmar
  Hilliges}, \bibinfo{person}{Shahram Izadi}, \bibinfo{person}{Alex~D Butler},
  \bibinfo{person}{Jiawen Chen}, \bibinfo{person}{Iason Oikonomidis}, {and}
  \bibinfo{person}{Patrick Olivier}.} \bibinfo{year}{2012}\natexlab{}.
\newblock \showarticletitle{Digits: freehand 3D interactions anywhere using a
  wrist-worn gloveless sensor}. In \bibinfo{booktitle}{\emph{Proceedings of the
  25th annual ACM symposium on User interface software and technology}}. ACM,
  \bibinfo{pages}{167--176}.
\newblock


\bibitem[\protect\citeauthoryear{Kingma and Ba}{Kingma and Ba}{2014}]%
        {kingma2014adam}
\bibfield{author}{\bibinfo{person}{Diederik~P Kingma} {and}
  \bibinfo{person}{Jimmy Ba}.} \bibinfo{year}{2014}\natexlab{}.
\newblock \showarticletitle{Adam: A method for stochastic optimization}.
\newblock \bibinfo{journal}{\emph{arXiv preprint arXiv:1412.6980}}
  (\bibinfo{year}{2014}).
\newblock


\bibitem[\protect\citeauthoryear{Koller, Zargaran, Ney, and Bowden}{Koller
  et~al\mbox{.}}{2016}]%
        {koller2016deep}
\bibfield{author}{\bibinfo{person}{Oscar Koller}, \bibinfo{person}{O Zargaran},
  \bibinfo{person}{Hermann Ney}, {and} \bibinfo{person}{Richard Bowden}.}
  \bibinfo{year}{2016}\natexlab{}.
\newblock \showarticletitle{Deep sign: hybrid CNN-HMM for continuous sign
  language recognition}. In \bibinfo{booktitle}{\emph{Proceedings of the
  British Machine Vision Conference 2016}}.
\newblock


\bibitem[\protect\citeauthoryear{Kyriazis and Argyros}{Kyriazis and
  Argyros}{2014}]%
        {kyriazis_cvpr2014}
\bibfield{author}{\bibinfo{person}{Nikolaos Kyriazis} {and}
  \bibinfo{person}{Antonis Argyros}.} \bibinfo{year}{2014}\natexlab{}.
\newblock \showarticletitle{Scalable 3d tracking of multiple interacting
  objects}. In \bibinfo{booktitle}{\emph{IEEE Conference on Computer Vision and
  Pattern Recognition (CVPR)}}. \bibinfo{pages}{3430--3437}.
\newblock


\bibitem[\protect\citeauthoryear{LeapMotion}{LeapMotion}{2016}]%
        {leap_motion}
\bibfield{author}{\bibinfo{person}{LeapMotion}.}
  \bibinfo{year}{2016}\natexlab{}.
\newblock \bibinfo{howpublished}{\url{https://developer.leapmotion.com/orion}}.
\newblock


\bibitem[\protect\citeauthoryear{Melax, Keselman, and Orsten}{Melax
  et~al\mbox{.}}{2013}]%
        {melax2013dynamics}
\bibfield{author}{\bibinfo{person}{Stan Melax}, \bibinfo{person}{Leonid
  Keselman}, {and} \bibinfo{person}{Sterling Orsten}.}
  \bibinfo{year}{2013}\natexlab{}.
\newblock \showarticletitle{Dynamics based 3D skeletal hand tracking}. In
  \bibinfo{booktitle}{\emph{Proceedings of Graphics Interface 2013}}. Canadian
  Information Processing Society, \bibinfo{pages}{63--70}.
\newblock


\bibitem[\protect\citeauthoryear{Mueller, Bernard, Sotnychenko, Mehta, Sridhar,
  Casas, and Theobalt}{Mueller et~al\mbox{.}}{2018}]%
        {mueller_cvpr2018}
\bibfield{author}{\bibinfo{person}{Franziska Mueller}, \bibinfo{person}{Florian
  Bernard}, \bibinfo{person}{Oleksandr Sotnychenko}, \bibinfo{person}{Dushyant
  Mehta}, \bibinfo{person}{Srinath Sridhar}, \bibinfo{person}{Dan Casas}, {and}
  \bibinfo{person}{Christian Theobalt}.} \bibinfo{year}{2018}\natexlab{}.
\newblock \showarticletitle{GANerated Hands for Real-Time 3D Hand Tracking from
  Monocular RGB}. In \bibinfo{booktitle}{\emph{Proceedings of Computer Vision
  and Pattern Recognition ({CVPR})}}. 11.
\newblock
\urldef\tempurl%
\url{http://handtracker.mpi-inf.mpg.de/projects/GANeratedHands/}
\showURL{%
\tempurl}


\bibitem[\protect\citeauthoryear{Mueller, Mehta, Sotnychenko, Sridhar, Casas,
  and Theobalt}{Mueller et~al\mbox{.}}{2017}]%
        {mueller_iccv2017}
\bibfield{author}{\bibinfo{person}{Franziska Mueller},
  \bibinfo{person}{Dushyant Mehta}, \bibinfo{person}{Oleksandr Sotnychenko},
  \bibinfo{person}{Srinath Sridhar}, \bibinfo{person}{Dan Casas}, {and}
  \bibinfo{person}{Christian Theobalt}.} \bibinfo{year}{2017}\natexlab{}.
\newblock \showarticletitle{Real-time Hand Tracking under Occlusion from an
  Egocentric RGB-D Sensor}. In \bibinfo{booktitle}{\emph{International
  Conference on Computer Vision ({ICCV})}}.
\newblock


\bibitem[\protect\citeauthoryear{Newell, Yang, and Deng}{Newell
  et~al\mbox{.}}{2016}]%
        {newell2016stacked}
\bibfield{author}{\bibinfo{person}{Alejandro Newell}, \bibinfo{person}{Kaiyu
  Yang}, {and} \bibinfo{person}{Jia Deng}.} \bibinfo{year}{2016}\natexlab{}.
\newblock \showarticletitle{Stacked hourglass networks for human pose
  estimation}. In \bibinfo{booktitle}{\emph{European Conference on Computer
  Vision}}. Springer, \bibinfo{pages}{483--499}.
\newblock


\bibitem[\protect\citeauthoryear{Oberweger, Wohlhart, and Lepetit}{Oberweger
  et~al\mbox{.}}{2015}]%
        {oberweger_iccv2015}
\bibfield{author}{\bibinfo{person}{Markus Oberweger}, \bibinfo{person}{Paul
  Wohlhart}, {and} \bibinfo{person}{Vincent Lepetit}.}
  \bibinfo{year}{2015}\natexlab{}.
\newblock \showarticletitle{Training a feedback loop for hand pose estimation}.
  In \bibinfo{booktitle}{\emph{IEEE International Conference on Computer Vision
  (ICCV)}}. \bibinfo{pages}{3316--3324}.
\newblock


\bibitem[\protect\citeauthoryear{Oikonomidis, Kyriazis, and
  Argyros}{Oikonomidis et~al\mbox{.}}{2011a}]%
        {oikonomidis2011efficient}
\bibfield{author}{\bibinfo{person}{Iason Oikonomidis},
  \bibinfo{person}{Nikolaos Kyriazis}, {and} \bibinfo{person}{Antonis~A
  Argyros}.} \bibinfo{year}{2011}\natexlab{a}.
\newblock \showarticletitle{Efficient model-based 3D tracking of hand
  articulations using Kinect.}. In \bibinfo{booktitle}{\emph{BMVC}},
  Vol.~\bibinfo{volume}{1}. \bibinfo{pages}{3}.
\newblock


\bibitem[\protect\citeauthoryear{Oikonomidis, Kyriazis, and
  Argyros}{Oikonomidis et~al\mbox{.}}{2011b}]%
        {oikonomidis2011full}
\bibfield{author}{\bibinfo{person}{Iason Oikonomidis},
  \bibinfo{person}{Nikolaos Kyriazis}, {and} \bibinfo{person}{Antonis~A
  Argyros}.} \bibinfo{year}{2011}\natexlab{b}.
\newblock \showarticletitle{Full dof tracking of a hand interacting with an
  object by modeling occlusions and physical constraints}. In
  \bibinfo{booktitle}{\emph{IEEE International Conference on Computer Vision
  (ICCV)}}. IEEE, \bibinfo{pages}{2088--2095}.
\newblock


\bibitem[\protect\citeauthoryear{Oikonomidis, Kyriazis, and
  Argyros}{Oikonomidis et~al\mbox{.}}{2012}]%
        {oikonomidis2012tracking}
\bibfield{author}{\bibinfo{person}{Iasonas Oikonomidis},
  \bibinfo{person}{Nikolaos Kyriazis}, {and} \bibinfo{person}{Antonis~A
  Argyros}.} \bibinfo{year}{2012}\natexlab{}.
\newblock \showarticletitle{Tracking the articulated motion of two strongly
  interacting hands}. In \bibinfo{booktitle}{\emph{Computer Vision and Pattern
  Recognition (CVPR), 2012 IEEE Conference on}}. IEEE,
  \bibinfo{pages}{1862--1869}.
\newblock


\bibitem[\protect\citeauthoryear{Qian, Sun, Wei, Tang, and Sun}{Qian
  et~al\mbox{.}}{2014}]%
        {qian_cvpr2014}
\bibfield{author}{\bibinfo{person}{Chen Qian}, \bibinfo{person}{Xiao Sun},
  \bibinfo{person}{Yichen Wei}, \bibinfo{person}{Xiaoou Tang}, {and}
  \bibinfo{person}{Jian Sun}.} \bibinfo{year}{2014}\natexlab{}.
\newblock \showarticletitle{{Realtime and Robust Hand Tracking from Depth}}. In
  \bibinfo{booktitle}{\emph{IEEE Conference on Computer Vision and Pattern
  Recognition (CVPR)}}. \bibinfo{pages}{1106--1113}.
\newblock


\bibitem[\protect\citeauthoryear{Remelli, Tkach, Tagliasacchi, and
  Pauly}{Remelli et~al\mbox{.}}{2017}]%
        {Remelli_2017_ICCV}
\bibfield{author}{\bibinfo{person}{Edoardo Remelli}, \bibinfo{person}{Anastasia
  Tkach}, \bibinfo{person}{Andrea Tagliasacchi}, {and} \bibinfo{person}{Mark
  Pauly}.} \bibinfo{year}{2017}\natexlab{}.
\newblock \showarticletitle{Low-Dimensionality Calibration Through Local
  Anisotropic Scaling for Robust Hand Model Personalization}. In
  \bibinfo{booktitle}{\emph{The IEEE International Conference on Computer
  Vision (ICCV)}}.
\newblock


\bibitem[\protect\citeauthoryear{Rogez, Khademi, Supan{\v{c}}i{\v{c}}~III,
  Montiel, and Ramanan}{Rogez et~al\mbox{.}}{2014}]%
        {rogez_eccv2014workshop}
\bibfield{author}{\bibinfo{person}{Gr{\'e}gory Rogez}, \bibinfo{person}{Maryam
  Khademi}, \bibinfo{person}{JS Supan{\v{c}}i{\v{c}}~III},
  \bibinfo{person}{Jose Maria~Martinez Montiel}, {and} \bibinfo{person}{Deva
  Ramanan}.} \bibinfo{year}{2014}\natexlab{}.
\newblock \showarticletitle{{3D hand pose detection in egocentric RGB-D
  images}}. In \bibinfo{booktitle}{\emph{Workshop at the European Conference on
  Computer Vision}}. Springer, \bibinfo{pages}{356--371}.
\newblock


\bibitem[\protect\citeauthoryear{Romero, Tzionas, and Black}{Romero
  et~al\mbox{.}}{2017}]%
        {Romero_siggraphasia2017}
\bibfield{author}{\bibinfo{person}{Javier Romero}, \bibinfo{person}{Dimitrios
  Tzionas}, {and} \bibinfo{person}{Michael~J. Black}.}
  \bibinfo{year}{2017}\natexlab{}.
\newblock \showarticletitle{Embodied Hands: Modeling and Capturing Hands and
  Bodies Together}.
\newblock \bibinfo{journal}{\emph{ACM Trans. Graph.}} \bibinfo{volume}{36},
  \bibinfo{number}{6}, Article \bibinfo{articleno}{245} (\bibinfo{date}{Nov.}
  \bibinfo{year}{2017}), \bibinfo{numpages}{17}~pages.
\newblock
\showISSN{0730-0301}
\urldef\tempurl%
\url{https://doi.org/10.1145/3130800.3130883}
\showDOI{\tempurl}


\bibitem[\protect\citeauthoryear{Ronneberger, Fischer, and Brox}{Ronneberger
  et~al\mbox{.}}{2015}]%
        {unet2015}
\bibfield{author}{\bibinfo{person}{Olaf Ronneberger}, \bibinfo{person}{Philipp
  Fischer}, {and} \bibinfo{person}{Thomas Brox}.}
  \bibinfo{year}{2015}\natexlab{}.
\newblock \showarticletitle{U-net: Convolutional networks for biomedical image
  segmentation}. In \bibinfo{booktitle}{\emph{International Conference on
  Medical image computing and computer-assisted intervention}}. Springer,
  \bibinfo{pages}{234--241}.
\newblock


\bibitem[\protect\citeauthoryear{Sharp, Keskin, Robertson, Taylor, Shotton,
  Kim, Rhemann, Leichter, Vinnikov, Wei, et~al\mbox{.}}{Sharp
  et~al\mbox{.}}{2015}]%
        {sharp2015accurate}
\bibfield{author}{\bibinfo{person}{Toby Sharp}, \bibinfo{person}{Cem Keskin},
  \bibinfo{person}{Duncan Robertson}, \bibinfo{person}{Jonathan Taylor},
  \bibinfo{person}{Jamie Shotton}, \bibinfo{person}{David Kim},
  \bibinfo{person}{Christoph Rhemann}, \bibinfo{person}{Ido Leichter},
  \bibinfo{person}{Alon Vinnikov}, \bibinfo{person}{Yichen Wei},
  {et~al\mbox{.}}} \bibinfo{year}{2015}\natexlab{}.
\newblock \showarticletitle{Accurate, robust, and flexible real-time hand
  tracking}. In \bibinfo{booktitle}{\emph{Proceedings of ACM Conference on
  Human Factors in Computing Systems (CHI)}}. ACM, \bibinfo{pages}{3633--3642}.
\newblock


\bibitem[\protect\citeauthoryear{Simon, Joo, Matthews, and Sheikh}{Simon
  et~al\mbox{.}}{2017}]%
        {simon2017hand}
\bibfield{author}{\bibinfo{person}{Tomas Simon}, \bibinfo{person}{Hanbyul Joo},
  \bibinfo{person}{Iain Matthews}, {and} \bibinfo{person}{Yaser Sheikh}.}
  \bibinfo{year}{2017}\natexlab{}.
\newblock \showarticletitle{Hand Keypoint Detection in Single Images using
  Multiview Bootstrapping}. In \bibinfo{booktitle}{\emph{IEEE Conference on
  Computer Vision and Pattern Recognition (CVPR)}}.
\newblock


\bibitem[\protect\citeauthoryear{Soliman, Mueller, Hegemann, Roo, Theobalt, and
  Steimle}{Soliman et~al\mbox{.}}{2018}]%
        {soliman2018fingerinput}
\bibfield{author}{\bibinfo{person}{Mohamed Soliman}, \bibinfo{person}{Franziska
  Mueller}, \bibinfo{person}{Lena Hegemann}, \bibinfo{person}{Joan~Sol Roo},
  \bibinfo{person}{Christian Theobalt}, {and} \bibinfo{person}{J{\"u}rgen
  Steimle}.} \bibinfo{year}{2018}\natexlab{}.
\newblock \showarticletitle{FingerInput: Capturing Expressive Single-Hand
  Thumb-to-Finger Microgestures}. In \bibinfo{booktitle}{\emph{Proceedings of
  the 2018 ACM International Conference on Interactive Surfaces and Spaces}}.
  ACM, \bibinfo{pages}{177--187}.
\newblock


\bibitem[\protect\citeauthoryear{Spurr, Song, Park, and Hilliges}{Spurr
  et~al\mbox{.}}{2018}]%
        {Spurr_2018_CVPR}
\bibfield{author}{\bibinfo{person}{Adrian Spurr}, \bibinfo{person}{Jie Song},
  \bibinfo{person}{Seonwook Park}, {and} \bibinfo{person}{Otmar Hilliges}.}
  \bibinfo{year}{2018}\natexlab{}.
\newblock \showarticletitle{Cross-Modal Deep Variational Hand Pose Estimation}.
  In \bibinfo{booktitle}{\emph{The IEEE Conference on Computer Vision and
  Pattern Recognition (CVPR)}}.
\newblock


\bibitem[\protect\citeauthoryear{Sridhar, Mueller, Oulasvirta, and
  Theobalt}{Sridhar et~al\mbox{.}}{2015}]%
        {sridhar_cvpr2015}
\bibfield{author}{\bibinfo{person}{Srinath Sridhar}, \bibinfo{person}{Franziska
  Mueller}, \bibinfo{person}{Antti Oulasvirta}, {and}
  \bibinfo{person}{Christian Theobalt}.} \bibinfo{year}{2015}\natexlab{}.
\newblock \showarticletitle{{Fast and Robust Hand Tracking Using
  Detection-Guided Optimization}}. In \bibinfo{booktitle}{\emph{IEEE Conference
  on Computer Vision and Pattern Recognition (CVPR)}}. 9.
\newblock
\urldef\tempurl%
\url{http://handtracker.mpi-inf.mpg.de/projects/FastHandTracker/}
\showURL{%
\tempurl}


\bibitem[\protect\citeauthoryear{Sridhar, Mueller, Zollh{\"o}efer, Casas,
  Oulasvirta, and Theobalt}{Sridhar et~al\mbox{.}}{2016}]%
        {sridhar_eccv2016}
\bibfield{author}{\bibinfo{person}{Srinath Sridhar}, \bibinfo{person}{Franziska
  Mueller}, \bibinfo{person}{Michael Zollh{\"o}efer}, \bibinfo{person}{Dan
  Casas}, \bibinfo{person}{Antti Oulasvirta}, {and} \bibinfo{person}{Christian
  Theobalt}.} \bibinfo{year}{2016}\natexlab{}.
\newblock \showarticletitle{{Real-time Joint Tracking of a Hand Manipulating an
  Object from RGB-D Input}}. In \bibinfo{booktitle}{\emph{European Conference
  on Computer Vision ({ECCV})}}. 17.
\newblock
\urldef\tempurl%
\url{http://handtracker.mpi-inf.mpg.de/projects/RealtimeHO/}
\showURL{%
\tempurl}


\bibitem[\protect\citeauthoryear{Sridhar, Oulasvirta, and Theobalt}{Sridhar
  et~al\mbox{.}}{2013}]%
        {sridhar_iccv2013}
\bibfield{author}{\bibinfo{person}{Srinath Sridhar}, \bibinfo{person}{Antti
  Oulasvirta}, {and} \bibinfo{person}{Christian Theobalt}.}
  \bibinfo{year}{2013}\natexlab{}.
\newblock \showarticletitle{{Interactive markerless articulated hand motion
  tracking using RGB and depth data}}. In \bibinfo{booktitle}{\emph{IEEE
  Conference on Computer Vision and Pattern Recognition (CVPR)}}.
  \bibinfo{pages}{2456--2463}.
\newblock


\bibitem[\protect\citeauthoryear{Sridhar, Rhodin, Seidel, Oulasvirta, and
  Theobalt}{Sridhar et~al\mbox{.}}{2014}]%
        {ellipsoidtracker_3dv2014}
\bibfield{author}{\bibinfo{person}{Srinath Sridhar}, \bibinfo{person}{Helge
  Rhodin}, \bibinfo{person}{Hans-Peter Seidel}, \bibinfo{person}{Antti
  Oulasvirta}, {and} \bibinfo{person}{Christian Theobalt}.}
  \bibinfo{year}{2014}\natexlab{}.
\newblock \showarticletitle{Real-time Hand Tracking Using a Sum of Anisotropic
  Gaussians Model}. In \bibinfo{booktitle}{\emph{Proceedings of the
  International Conference on 3D Vision (3DV)}}.
\newblock


\bibitem[\protect\citeauthoryear{Supan{\v{c}}i{\v{c}}, Rogez, Yang, Shotton,
  and Ramanan}{Supan{\v{c}}i{\v{c}} et~al\mbox{.}}{2018}]%
        {Supan2018}
\bibfield{author}{\bibinfo{person}{James~Steven Supan{\v{c}}i{\v{c}}},
  \bibinfo{person}{Gr{\'e}gory Rogez}, \bibinfo{person}{Yi Yang},
  \bibinfo{person}{Jamie Shotton}, {and} \bibinfo{person}{Deva Ramanan}.}
  \bibinfo{year}{2018}\natexlab{}.
\newblock \showarticletitle{Depth-Based Hand Pose Estimation: Methods, Data,
  and Challenges}.
\newblock \bibinfo{journal}{\emph{International Journal of Computer Vision}}
  \bibinfo{volume}{126}, \bibinfo{number}{11} (\bibinfo{date}{01 Nov}
  \bibinfo{year}{2018}), \bibinfo{pages}{1180--1198}.
\newblock
\showISSN{1573-1405}
\urldef\tempurl%
\url{https://doi.org/10.1007/s11263-018-1081-7}
\showDOI{\tempurl}


\bibitem[\protect\citeauthoryear{Tagliasacchi, Schroeder, Tkach, Bouaziz,
  Botsch, and Pauly}{Tagliasacchi et~al\mbox{.}}{2015}]%
        {tagliasacchi_sgp2015}
\bibfield{author}{\bibinfo{person}{Andrea Tagliasacchi},
  \bibinfo{person}{Matthias Schroeder}, \bibinfo{person}{Anastasia Tkach},
  \bibinfo{person}{Sofien Bouaziz}, \bibinfo{person}{Mario Botsch}, {and}
  \bibinfo{person}{Mark Pauly}.} \bibinfo{year}{2015}\natexlab{}.
\newblock \showarticletitle{{Robust Articulated-ICP for Real-Time Hand
  Tracking}}.
\newblock \bibinfo{journal}{\emph{Computer Graphics Forum (Symposium on
  Geometry Processing)}} \bibinfo{volume}{34}, \bibinfo{number}{5}
  (\bibinfo{year}{2015}).
\newblock


\bibitem[\protect\citeauthoryear{Tan, Cashman, Taylor, Fitzgibbon, Tarlow,
  Khamis, Izadi, and Shotton}{Tan et~al\mbox{.}}{2016}]%
        {tan_cvpr2016}
\bibfield{author}{\bibinfo{person}{David~Joseph Tan}, \bibinfo{person}{Thomas
  Cashman}, \bibinfo{person}{Jonathan Taylor}, \bibinfo{person}{Andrew
  Fitzgibbon}, \bibinfo{person}{Daniel Tarlow}, \bibinfo{person}{Sameh Khamis},
  \bibinfo{person}{Shahram Izadi}, {and} \bibinfo{person}{Jamie Shotton}.}
  \bibinfo{year}{2016}\natexlab{}.
\newblock \showarticletitle{Fits Like a Glove: Rapid and Reliable Hand Shape
  Personalization}. In \bibinfo{booktitle}{\emph{IEEE Conference on Computer
  Vision and Pattern Recognition (CVPR)}}. \bibinfo{pages}{5610--5619}.
\newblock


\bibitem[\protect\citeauthoryear{Tang, Jin~Chang, Tejani, and Kim}{Tang
  et~al\mbox{.}}{2014}]%
        {tang2014latent}
\bibfield{author}{\bibinfo{person}{Danhang Tang}, \bibinfo{person}{Hyung
  Jin~Chang}, \bibinfo{person}{Alykhan Tejani}, {and} \bibinfo{person}{Tae-Kyun
  Kim}.} \bibinfo{year}{2014}\natexlab{}.
\newblock \showarticletitle{Latent regression forest: Structured estimation of
  3d articulated hand posture}. In \bibinfo{booktitle}{\emph{Proceedings of the
  IEEE Conference on Computer Vision and Pattern Recognition}}.
  \bibinfo{pages}{3786--3793}.
\newblock


\bibitem[\protect\citeauthoryear{Tang, Taylor, Kohli, Keskin, Kim, and
  Shotton}{Tang et~al\mbox{.}}{2015}]%
        {tang_iccv2015}
\bibfield{author}{\bibinfo{person}{Danhang Tang}, \bibinfo{person}{Jonathan
  Taylor}, \bibinfo{person}{Pushmeet Kohli}, \bibinfo{person}{Cem Keskin},
  \bibinfo{person}{Tae-Kyun Kim}, {and} \bibinfo{person}{Jamie Shotton}.}
  \bibinfo{year}{2015}\natexlab{}.
\newblock \showarticletitle{Opening the Black Box: Hierarchical Sampling
  Optimization for Estimating Human Hand Pose}. In
  \bibinfo{booktitle}{\emph{Proc. ICCV}}.
\newblock


\bibitem[\protect\citeauthoryear{Taylor, Bordeaux, Cashman, Corish, Keskin,
  Sharp, Soto, Sweeney, Valentin, Luff, et~al\mbox{.}}{Taylor
  et~al\mbox{.}}{2016}]%
        {taylor_siggraph2016}
\bibfield{author}{\bibinfo{person}{Jonathan Taylor}, \bibinfo{person}{Lucas
  Bordeaux}, \bibinfo{person}{Thomas Cashman}, \bibinfo{person}{Bob Corish},
  \bibinfo{person}{Cem Keskin}, \bibinfo{person}{Toby Sharp},
  \bibinfo{person}{Eduardo Soto}, \bibinfo{person}{David Sweeney},
  \bibinfo{person}{Julien Valentin}, \bibinfo{person}{Benjamin Luff},
  {et~al\mbox{.}}} \bibinfo{year}{2016}\natexlab{}.
\newblock \showarticletitle{Efficient and precise interactive hand tracking
  through joint, continuous optimization of pose and correspondences}.
\newblock \bibinfo{journal}{\emph{ACM Transactions on Graphics (TOG)}}
  \bibinfo{volume}{35}, \bibinfo{number}{4} (\bibinfo{year}{2016}),
  \bibinfo{pages}{143}.
\newblock


\bibitem[\protect\citeauthoryear{Taylor, Shotton, Sharp, and Fitzgibbon}{Taylor
  et~al\mbox{.}}{2012}]%
        {taylor2012vitruvian}
\bibfield{author}{\bibinfo{person}{Jonathan Taylor}, \bibinfo{person}{Jamie
  Shotton}, \bibinfo{person}{Toby Sharp}, {and} \bibinfo{person}{Andrew
  Fitzgibbon}.} \bibinfo{year}{2012}\natexlab{}.
\newblock \showarticletitle{The vitruvian manifold: Inferring dense
  correspondences for one-shot human pose estimation}. In
  \bibinfo{booktitle}{\emph{Computer Vision and Pattern Recognition (CVPR),
  2012 IEEE Conference on}}. IEEE, \bibinfo{pages}{103--110}.
\newblock


\bibitem[\protect\citeauthoryear{Taylor, Tankovich, Tang, Keskin, Kim,
  Davidson, Kowdle, and Izadi}{Taylor et~al\mbox{.}}{2017}]%
        {Taylor_siggraphasia2017}
\bibfield{author}{\bibinfo{person}{Jonathan Taylor}, \bibinfo{person}{Vladimir
  Tankovich}, \bibinfo{person}{Danhang Tang}, \bibinfo{person}{Cem Keskin},
  \bibinfo{person}{David Kim}, \bibinfo{person}{Philip Davidson},
  \bibinfo{person}{Adarsh Kowdle}, {and} \bibinfo{person}{Shahram Izadi}.}
  \bibinfo{year}{2017}\natexlab{}.
\newblock \showarticletitle{Articulated Distance Fields for Ultra-fast Tracking
  of Hands Interacting}.
\newblock \bibinfo{journal}{\emph{ACM Trans. Graph.}} \bibinfo{volume}{36},
  \bibinfo{number}{6}, Article \bibinfo{articleno}{244} (\bibinfo{date}{Nov.}
  \bibinfo{year}{2017}), \bibinfo{numpages}{12}~pages.
\newblock
\showISSN{0730-0301}
\urldef\tempurl%
\url{https://doi.org/10.1145/3130800.3130853}
\showDOI{\tempurl}


\bibitem[\protect\citeauthoryear{Tkach, Pauly, and Tagliasacchi}{Tkach
  et~al\mbox{.}}{2016}]%
        {tkach2016sphere}
\bibfield{author}{\bibinfo{person}{Anastasia Tkach}, \bibinfo{person}{Mark
  Pauly}, {and} \bibinfo{person}{Andrea Tagliasacchi}.}
  \bibinfo{year}{2016}\natexlab{}.
\newblock \showarticletitle{Sphere-meshes for real-time hand modeling and
  tracking}.
\newblock \bibinfo{journal}{\emph{ACM Transactions on Graphics (TOG)}}
  \bibinfo{volume}{35}, \bibinfo{number}{6} (\bibinfo{year}{2016}),
  \bibinfo{pages}{222}.
\newblock


\bibitem[\protect\citeauthoryear{Tkach, Tagliasacchi, Remelli, Pauly, and
  Fitzgibbon}{Tkach et~al\mbox{.}}{2017}]%
        {Tkach_siggraphasia2017}
\bibfield{author}{\bibinfo{person}{Anastasia Tkach}, \bibinfo{person}{Andrea
  Tagliasacchi}, \bibinfo{person}{Edoardo Remelli}, \bibinfo{person}{Mark
  Pauly}, {and} \bibinfo{person}{Andrew Fitzgibbon}.}
  \bibinfo{year}{2017}\natexlab{}.
\newblock \showarticletitle{Online Generative Model Personalization for Hand
  Tracking}.
\newblock \bibinfo{journal}{\emph{ACM Trans. Graph.}} \bibinfo{volume}{36},
  \bibinfo{number}{6}, Article \bibinfo{articleno}{243} (\bibinfo{date}{Nov.}
  \bibinfo{year}{2017}), \bibinfo{numpages}{11}~pages.
\newblock
\showISSN{0730-0301}
\urldef\tempurl%
\url{https://doi.org/10.1145/3130800.3130830}
\showDOI{\tempurl}


\bibitem[\protect\citeauthoryear{Tompson, Stein, Lecun, and Perlin}{Tompson
  et~al\mbox{.}}{2014}]%
        {tompson_tog2014}
\bibfield{author}{\bibinfo{person}{Jonathan Tompson}, \bibinfo{person}{Murphy
  Stein}, \bibinfo{person}{Yann Lecun}, {and} \bibinfo{person}{Ken Perlin}.}
  \bibinfo{year}{2014}\natexlab{}.
\newblock \showarticletitle{Real-Time Continuous Pose Recovery of Human Hands
  Using Convolutional Networks}.
\newblock \bibinfo{journal}{\emph{ACM Transactions on Graphics}}
  \bibinfo{volume}{33} (\bibinfo{date}{August} \bibinfo{year}{2014}).
\newblock


\bibitem[\protect\citeauthoryear{Tzionas, Ballan, Srikantha, Aponte, Pollefeys,
  and Gall}{Tzionas et~al\mbox{.}}{2016}]%
        {tzionas_ijcv2016}
\bibfield{author}{\bibinfo{person}{Dimitrios Tzionas}, \bibinfo{person}{Luca
  Ballan}, \bibinfo{person}{Abhilash Srikantha}, \bibinfo{person}{Pablo
  Aponte}, \bibinfo{person}{Marc Pollefeys}, {and} \bibinfo{person}{Juergen
  Gall}.} \bibinfo{year}{2016}\natexlab{}.
\newblock \showarticletitle{Capturing Hands in Action using Discriminative
  Salient Points and Physics Simulation}.
\newblock \bibinfo{journal}{\emph{International Journal of Computer Vision
  (IJCV)}} (\bibinfo{year}{2016}).
\newblock
\urldef\tempurl%
\url{http://files.is.tue.mpg.de/dtzionas/Hand-Object-Capture}
\showURL{%
\tempurl}


\bibitem[\protect\citeauthoryear{Verschoor, Lobo, and Otaduy}{Verschoor
  et~al\mbox{.}}{2018}]%
        {verschoor2018soft}
\bibfield{author}{\bibinfo{person}{Mickeal Verschoor}, \bibinfo{person}{Daniel
  Lobo}, {and} \bibinfo{person}{Miguel~A Otaduy}.}
  \bibinfo{year}{2018}\natexlab{}.
\newblock \showarticletitle{Soft Hand Simulation for Smooth and Robust Natural
  Interaction}. In \bibinfo{booktitle}{\emph{IEEE Conference on Virtual Reality
  and 3D User Interfaces (VR)}}. IEEE, \bibinfo{pages}{183--190}.
\newblock


\bibitem[\protect\citeauthoryear{Wan, Probst, Van~Gool, and Yao}{Wan
  et~al\mbox{.}}{2017}]%
        {wan2017crossing}
\bibfield{author}{\bibinfo{person}{Chengde Wan}, \bibinfo{person}{Thomas
  Probst}, \bibinfo{person}{Luc Van~Gool}, {and} \bibinfo{person}{Angela Yao}.}
  \bibinfo{year}{2017}\natexlab{}.
\newblock \showarticletitle{Crossing Nets: Combining GANs and VAEs with a
  Shared Latent Space for Hand Pose Estimation}. In
  \bibinfo{booktitle}{\emph{Proceedings of the IEEE Conference on Computer
  Vision and Pattern Recognition}}. \bibinfo{pages}{680--689}.
\newblock


\bibitem[\protect\citeauthoryear{Wan, Yao, and Van~Gool}{Wan
  et~al\mbox{.}}{2016}]%
        {wan2016hand}
\bibfield{author}{\bibinfo{person}{Chengde Wan}, \bibinfo{person}{Angela Yao},
  {and} \bibinfo{person}{Luc Van~Gool}.} \bibinfo{year}{2016}\natexlab{}.
\newblock \showarticletitle{Hand pose estimation from local surface normals}.
  In \bibinfo{booktitle}{\emph{European conference on computer vision}}.
  Springer, \bibinfo{pages}{554--569}.
\newblock


\bibitem[\protect\citeauthoryear{Wei, Huang, Ceylan, Vouga, and Li}{Wei
  et~al\mbox{.}}{2016}]%
        {wei2016dense}
\bibfield{author}{\bibinfo{person}{Lingyu Wei}, \bibinfo{person}{Qixing Huang},
  \bibinfo{person}{Duygu Ceylan}, \bibinfo{person}{Etienne Vouga}, {and}
  \bibinfo{person}{Hao Li}.} \bibinfo{year}{2016}\natexlab{}.
\newblock \showarticletitle{Dense Human Body Correspondences Using
  Convolutional Networks}. In \bibinfo{booktitle}{\emph{Computer Vision and
  Pattern Recognition (CVPR)}}.
\newblock


\bibitem[\protect\citeauthoryear{Ye and Kim}{Ye and Kim}{2018}]%
        {Ye_2018_ECCV}
\bibfield{author}{\bibinfo{person}{Qi Ye} {and} \bibinfo{person}{Tae-Kyun
  Kim}.} \bibinfo{year}{2018}\natexlab{}.
\newblock \showarticletitle{Occlusion-aware Hand Pose Estimation Using
  Hierarchical Mixture Density Network}. In \bibinfo{booktitle}{\emph{The
  European Conference on Computer Vision (ECCV)}}.
\newblock


\bibitem[\protect\citeauthoryear{Yuan, Garcia-Hernando, Stenger, Moon,
  Yong~Chang, Mu~Lee, Molchanov, Kautz, Honari, Ge, Yuan, Chen, Wang, Yang,
  Akiyama, Wu, Wan, Madadi, Escalera, Li, Lee, Oikonomidis, Argyros, and
  Kim}{Yuan et~al\mbox{.}}{2018}]%
        {Yuan_2018_CVPR}
\bibfield{author}{\bibinfo{person}{Shanxin Yuan}, \bibinfo{person}{Guillermo
  Garcia-Hernando}, \bibinfo{person}{Bj{\"o}rn Stenger},
  \bibinfo{person}{Gyeongsik Moon}, \bibinfo{person}{Ju Yong~Chang},
  \bibinfo{person}{Kyoung Mu~Lee}, \bibinfo{person}{Pavlo Molchanov},
  \bibinfo{person}{Jan Kautz}, \bibinfo{person}{Sina Honari},
  \bibinfo{person}{Liuhao Ge}, \bibinfo{person}{Junsong Yuan},
  \bibinfo{person}{Xinghao Chen}, \bibinfo{person}{Guijin Wang},
  \bibinfo{person}{Fan Yang}, \bibinfo{person}{Kai Akiyama},
  \bibinfo{person}{Yang Wu}, \bibinfo{person}{Qingfu Wan},
  \bibinfo{person}{Meysam Madadi}, \bibinfo{person}{Sergio Escalera},
  \bibinfo{person}{Shile Li}, \bibinfo{person}{Dongheui Lee},
  \bibinfo{person}{Iason Oikonomidis}, \bibinfo{person}{Antonis Argyros}, {and}
  \bibinfo{person}{Tae-Kyun Kim}.} \bibinfo{year}{2018}\natexlab{}.
\newblock \showarticletitle{Depth-Based 3D Hand Pose Estimation: From Current
  Achievements to Future Goals}. In \bibinfo{booktitle}{\emph{The IEEE
  Conference on Computer Vision and Pattern Recognition (CVPR)}}.
\newblock


\bibitem[\protect\citeauthoryear{Zhang, Zuo, Chen, Meng, and Zhang}{Zhang
  et~al\mbox{.}}{2017}]%
        {zhang2017beyond}
\bibfield{author}{\bibinfo{person}{Kai Zhang}, \bibinfo{person}{Wangmeng Zuo},
  \bibinfo{person}{Yunjin Chen}, \bibinfo{person}{Deyu Meng}, {and}
  \bibinfo{person}{Lei Zhang}.} \bibinfo{year}{2017}\natexlab{}.
\newblock \showarticletitle{Beyond a gaussian denoiser: Residual learning of
  deep cnn for image denoising}.
\newblock \bibinfo{journal}{\emph{IEEE Transactions on Image Processing}}
  \bibinfo{volume}{26}, \bibinfo{number}{7} (\bibinfo{year}{2017}),
  \bibinfo{pages}{3142--3155}.
\newblock


\bibitem[\protect\citeauthoryear{Zhao, Zhang, Min, and Chai}{Zhao
  et~al\mbox{.}}{2013}]%
        {Zhao_TOG2013}
\bibfield{author}{\bibinfo{person}{Wenping Zhao}, \bibinfo{person}{Jianjie
  Zhang}, \bibinfo{person}{Jianyuan Min}, {and} \bibinfo{person}{Jinxiang
  Chai}.} \bibinfo{year}{2013}\natexlab{}.
\newblock \showarticletitle{Robust Realtime Physics-based Motion Control for
  Human Grasping}.
\newblock \bibinfo{journal}{\emph{ACM Trans. Graph.}} \bibinfo{volume}{32},
  \bibinfo{number}{6}, Article \bibinfo{articleno}{207} (\bibinfo{date}{Nov.}
  \bibinfo{year}{2013}), \bibinfo{numpages}{12}~pages.
\newblock
\showISSN{0730-0301}
\urldef\tempurl%
\url{https://doi.org/10.1145/2508363.2508412}
\showDOI{\tempurl}


\bibitem[\protect\citeauthoryear{Zimmermann and Brox}{Zimmermann and
  Brox}{2017}]%
        {Zimmermann:2017um}
\bibfield{author}{\bibinfo{person}{Christian Zimmermann} {and}
  \bibinfo{person}{Thomas Brox}.} \bibinfo{year}{2017}\natexlab{}.
\newblock \showarticletitle{{Learning to Estimate 3D Hand Pose from Single RGB
  Images.}}. In \bibinfo{booktitle}{\emph{International Conference on Computer
  Vision (ICCV)}}.
\newblock


\end{thebibliography}
